%% file: main.tex
\documentclass[10pt]{article} % For LaTeX2e
\usepackage[preprint]{tmlr}
\input{math_commands.tex}

\usepackage{hyperref}
\usepackage{url}
\usepackage{placeins}
\usepackage{float}
\usepackage{tikz}
\usetikzlibrary{positioning}
\usepackage{booktabs}
\usepackage{graphicx}
\usepackage{multirow}
\definecolor{green}{RGB}{0, 128, 0}
\usepackage[table,xcdraw]{xcolor}
\usepackage{amsmath}
\usepackage{bm}
\usepackage{longtable}
\usepackage{caption}
\usepackage{subcaption}
\usepackage{hhline}
\DeclareCaptionLabelFormat{mysubtable}{Table #2}
\usepackage{makecell}
\makeatletter
\renewcommand{\p@subtable}{}
\makeatother
\usepackage{enumitem}

\usepackage[export]{adjustbox}
\title{DPAL: A Test-Time Adaptation Loss for Financial Time\\ Series Forecasting}

\author{%
\name Wuzhou Lian\footnotemark[2]
\email wlian@umich.edu \\
\addr University of Michigan
\AND
\name Haochun Wang\footnotemark[1]
\email haochun.wang@stonybrook.edu \\
\addr Stony Brook University
\AND
\name Yuefan Deng\footnotemark[1]
\email yuefan.deng@stonybrook.edu \\
\addr Stony Brook University
}

\def\month{MM}  % Insert correct month for camera-ready version
\def\year{YYYY} % Insert correct year for camera-ready version
\def\openreview{\url{https://openreview.net/forum?id=XXXX}} % Insert correct link to OpenReview for camera-ready version

\begin{document}

\maketitle

\begingroup
\renewcommand{\thefootnote}{\fnsymbol{footnote}}
\footnotetext[1]{Co-corresponding authors: Haochun Wang and Yuefan Deng.}
\footnotetext[2]{Wuzhou Lian conducted part of this work while at Stony Brook University.}
\endgroup

\begin{abstract}
Test-time adaptation (TTA) enables time series forecasters to update after deployment as new observations become available. However, adaptation objectives that focus primarily on prediction errors at individual time steps do not fully address the needs of financial time series forecasting. In addition to pointwise return fitting, return direction and the price trajectories reconstructed from sequences of returns are also important. To better account for these properties, we introduce Direction- and Price-Aware Adaptation Loss (DPAL), which combines pointwise return fitting with directional and price-calibration terms. We apply DPAL within existing TSF-TTA frameworks without modifying the source forecaster or the adapter architecture. Across a diverse collection of financial datasets, DPAL reduces reconstructed-price mean squared error in most evaluated settings.
\end{abstract}

\section{Introduction}
Time series forecasting aims to predict future observations from historical temporal patterns. In financial markets, reliable forecasts support a wide range of decisions, including portfolio analysis, risk assessment, and capital allocation. Although modern forecasting methods \citep{PatchTST,iTransformer,DLinear} have achieved good performance on standard forecasting benchmarks, their deployment in financial markets remains challenging. In particular, asset returns evolve over time, leading to distribution shifts that may degrade predictive accuracy after deployment.

Test-time adaptation for time series forecasting (TSF-TTA) \citep{TAFAS, PETSA, DynaTTA, COSA} provides a way to adjust a deployed forecaster as test targets become available. TAFAS \citep{TAFAS} keeps the source forecaster frozen and updates lightweight gated calibration modules using fully and partially observed targets. PETSA \citep{PETSA} improves parameter efficiency through low-rank calibration modules while using a composite loss to preserve forecasting accuracy under limited adaptation capacity. Although these methods were developed for non-stationary time series forecasting, their original objectives do not explicitly account for two properties relevant to financial forecasting: whether a predicted return has the correct sign and whether return errors accumulate into a distorted price trajectory. 

\begin{figure}[t]
\centering
\includegraphics[width=1.0\linewidth]{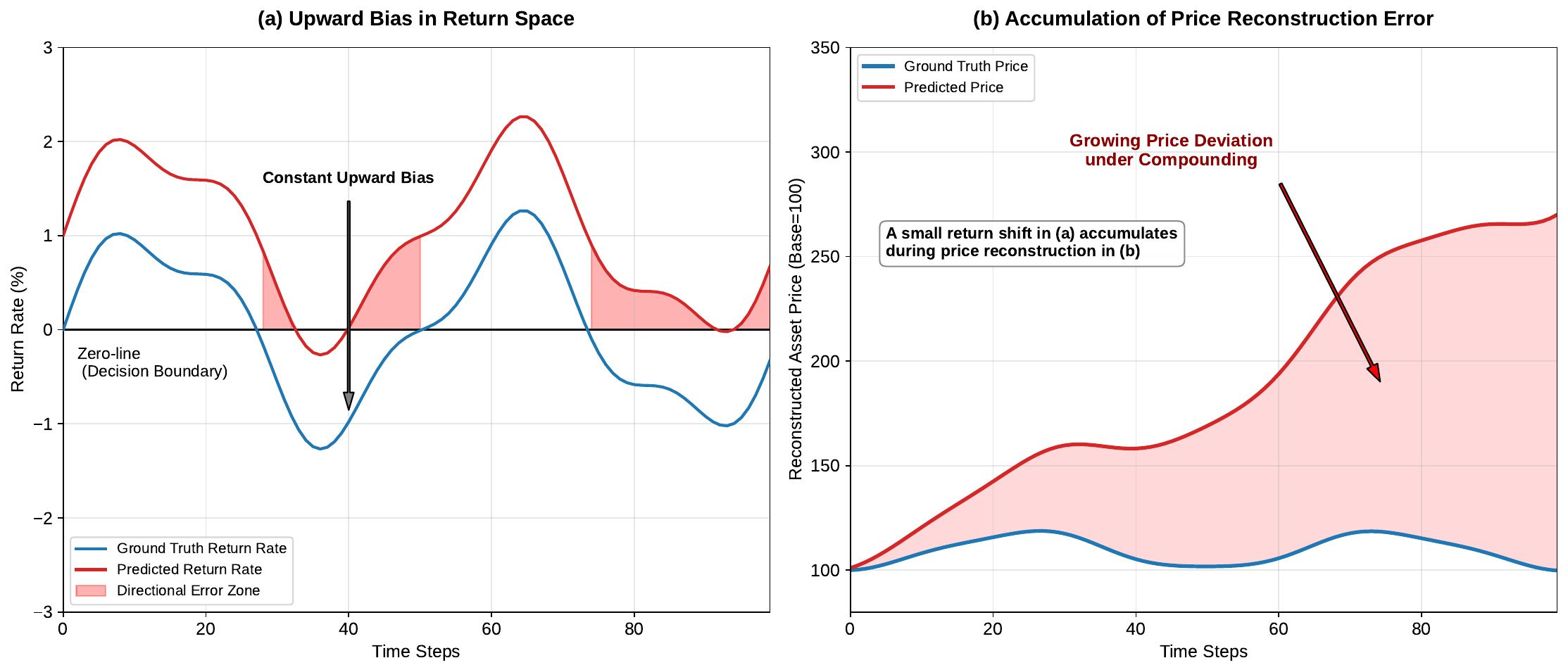}\hfill
\caption{Illustration of constant upward bias and its impact on price reconstruction. (a) A small upward shift in the predicted return rate preserves the original temporal shape but induces directional misjudgments (red areas), which are not distinguished by standard MSE. (b) During price reconstruction,  the effect of this persistent return shift accumulates through compounding, leading to a progressively widening deviation between the predicted and ground-truth price trajectories.}
\label{rate_and_price}
\vspace{-1em}
\end{figure}

Mean squared error (MSE) measures numerical deviation but does not treat a sign reversal as a distinct type of forecasting error. For example, if the realized return is 1\%, predictions of 3\% and -1\% have the same absolute error, although only the latter reverses the direction. This distinction matters whenever a decision uses the predicted sign of the return. In addition, pointwise return fitting does not directly constrain the price trajectory reconstructed by compounding a sequence of predicted returns. Small but persistent return biases can accumulate over the forecasting horizon and produce a growing discrepancy between predicted and realized price paths, as illustrated in Figure \ref{rate_and_price}. 

Together, these considerations motivate an adaptation objective that jointly incorporates pointwise return fitting, penalties for predictions whose signs differ from the observed returns, and calibration of cumulative price trajectories. To this end, we introduce Direction- and Price-Aware Adaptation Loss (DPAL). Following \citet{FAC}, we evaluate DPAL under a matured-ground-truth protocol, in which the adapter is updated only after the complete target window becomes available, without using partially observed ground truth for adaptation. Specifically, a past forecast window is eligible for adaptation only when its entire target span has been observed before the current prediction begins. This temporal separation prevents adaptation supervision from overlapping with the evaluated prediction span and avoids using partially revealed values from the current window as adaptation targets. We apply the same DPAL objective to both TAFAS \citep{TAFAS} and PETSA \citep{PETSA}, while retaining their original adaptation modules. For TAFAS, DPAL augments the original MSE objective with directional and price-calibration terms. For PETSA, DPAL replaces the original Huber, frequency-domain, and patch-wise structural terms.

We evaluate TAFAS and PETSA equipped with DPAL on eight financial datasets assembled for this study, spanning diverse equity and broader market settings. The experiments use five source forecasters: PatchTST \citep{PatchTST}, iTransformer \citep{iTransformer}, DLinear \citep{DLinear}, OLS \citep{OLS}, and FreTS \citep{FreTS}. Each model is evaluated at forecasting horizons of 60, 90, and 120 trading days. Compared with the frozen source forecasters and the TAFAS and PETSA baselines, the DPAL variants reduce reconstructed-price MSE in the large majority of the evaluated settings.

The main contributions of this work are summarized as follows:

\begin{itemize}
\item We highlight that existing TSF-TTA objectives do not explicitly account for two properties relevant to financial time series forecasting: return direction and compounded price trajectories.
\item We introduce DPAL, which combines MSE, a directional loss, and a price-calibration loss, and we instantiate it in TAFAS \citep{TAFAS} and PETSA \citep{PETSA}.
\item We evaluate TAFAS \citep{TAFAS} and PETSA \citep{PETSA} equipped with DPAL across eight financial datasets, five source forecasters, and three forecasting horizons. The results show lower reconstructed-price MSE in the large majority of the evaluated settings.\end{itemize}

\section{Related Work}

\subsection{Deep Learning in Time Series Forecasting}
Recent advances in long-term time series forecasting (LTSF) include Transformer-based models such as PatchTST \citep{PatchTST} and iTransformer \citep{iTransformer}, linear models such as DLinear \citep{DLinear} and OLS \citep{OLS}, and frequency-domain methods such as FreTS \citep{FreTS}. These models are commonly evaluated on datasets from domains such as energy \citep{energy}, traffic \citep{traffic1, traffic2}, and weather \citep{weather1, weather2}. These approaches have demonstrated competitive forecasting performance on standard LTSF benchmarks.

\subsection{Test-Time Adaptation for Distribution Shifts}

As data distributions shift over time, time series forecasters may experience performance degradation after deployment. In this setting, test-time adaptation (TTA) \citep{TAFAS, PETSA, DynaTTA, COSA} has been introduced into time series forecasting, using information that becomes available after deployment to update a pretrained forecaster. TAFAS~\citep{TAFAS} freezes the source forecaster and updates gated calibration modules (GCMs) using fully and partially observed targets, with an adaptation schedule based on temporal periodicity. PETSA~\citep{PETSA} reduces the trainable parameter count through low-rank input and output calibration modules and optimizes a composite objective containing robust, frequency-domain, and patch-wise structural terms. DynaTTA~\citep{DynaTTA} estimates distribution shift from prediction errors and representation drift, then adjusts the adaptation rate and a shift-conditioned gate that controls the influence of its calibration modules. COSA~\citep{COSA} performs gated residual correction in the output space using context from recently observed targets. Despite their differences in architecture and adaptation strategy, their adaptation objectives are not specifically designed for financial time series forecasting. 

\subsection{Objectives and Forecasting in Financial Time Series}
Deep stock forecasting models commonly use mean squared error (MSE) either as the primary objective or together with auxiliary objectives. MASTER \citep{MASTER} uses market-guided feature selection and Transformer-based aggregation to predict normalized stock returns with MSE, whereas FactorGCL \citep{FactorGCL} combines recurrent and hypergraph encoders and jointly optimizes multi-horizon return MSE and a temporal residual contrastive loss. To better align prediction with stock selection, RSR \citep{RSR}, STHAN-SR \citep{STHAN}, and StockMixer \citep{StockMixer} augment pointwise squared-error regression with pairwise ranking losses, thereby preserving return-magnitude supervision while encouraging correct cross-stock ordering. Their evaluations further consider ranking quality and portfolio performance. Directional information has also been incorporated into forecasting objectives. MADL \citep{MICHANKOW2024102375} uses discrete sign agreement between predicted and realized returns, weighted by realized return magnitudes, for trading signal generation.
CosDir \citep{lee2026magnitudeshapedirectionawareloss} augments MSE with
cosine alignment between the first-difference vectors of predictions
and targets. DPAL uses a pointwise rectified-product penalty on
original-scale returns and combines it with pointwise return fitting
and compounded price-path calibration for test-time adaptation.

\section{DPAL: Direction- and Price-Aware Adaptation Loss}

Recall that existing TSF-TTA objectives do not explicitly account for return direction or compounded price trajectories in financial time series forecasting. To incorporate these properties into the adaptation objective, we introduce Direction- and Price-Aware Adaptation Loss (DPAL), which changes only the adaptation objective while leaving the forecasting backbone and the test-time adapter unchanged.

To formulate the loss components, let $b \in \{1,\ldots,B\}$ denote the batch index, $h \in \{1,\ldots,H\}$ the forecast position, and $c \in \{1,\ldots,C\}$ the target-series index. The pointwise mean squared error is computed in the normalized return space. For the directional and price-aware terms, the predicted and ground-truth returns are transformed back to the original return scale using the inverse normalization transform defined for each dataset. We denote the corresponding original-scale returns by $\hat{r}_{b,h,c}$ and $r_{b,h,c}$, respectively.

\subsection{Direction-Aware Loss}

\begin{figure}[!htbp]
\centering
\includegraphics[width=0.75\linewidth]{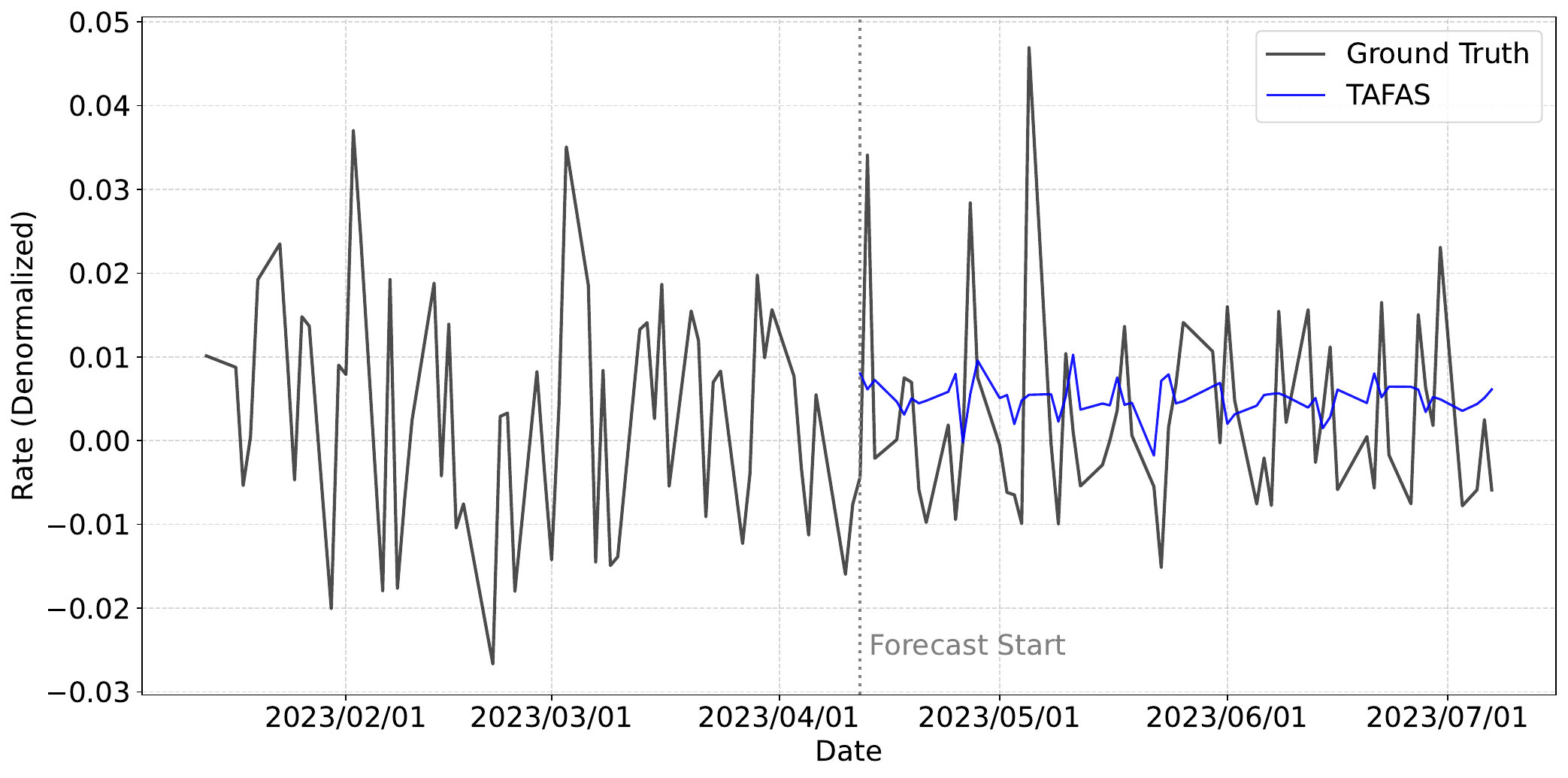}
\caption{Illustrative example of MSE-based test-time adaptation on AAPL. The figure shows return predictions from iTransformer with TAFAS without POGT, using both a look-back window and a forecasting horizon of 60 trading days. The vertical dashed line marks the beginning of the forecasting window. In this example, the adapted predictions remain within a narrow range of small positive values while the observed returns frequently change sign.}
\vspace{-0.5em}
\label{iTransformer_rate}
\end{figure}

When pointwise MSE is used as the test-time adaptation objective, the expected squared error is minimized by the conditional mean of the target return given the available inputs. In financial return forecasting, much of the observed variation may be difficult to predict from the available inputs. In such settings, MSE-based adaptation may produce predictions that remain near the center of the return distribution and vary less than the observed returns. Figure \ref{iTransformer_rate} provides an illustrative example: the adapted predictions are concentrated within a narrow range of small positive values, whereas the observed returns vary more widely and frequently change sign. Such predictions may achieve relatively small pointwise error while failing to reflect changes in return direction.

\begin{figure}[!htbp]
\centering
\includegraphics[width=1.0\linewidth]{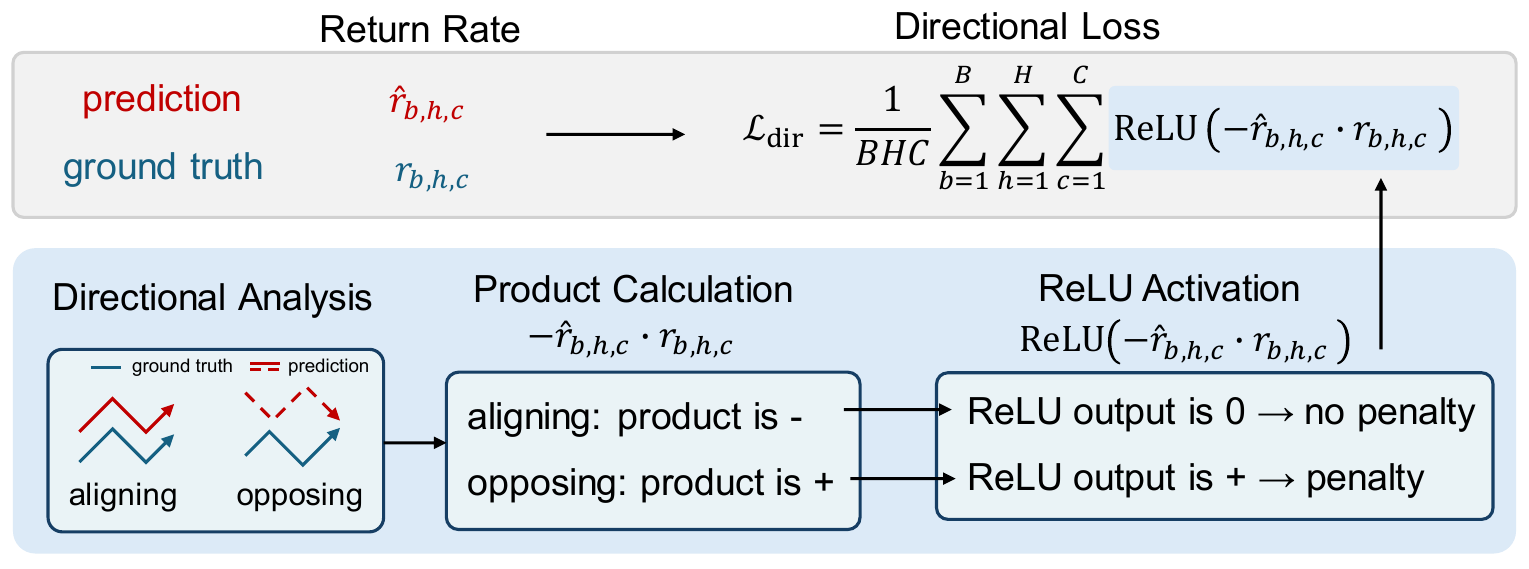}
\vspace{-15pt}
\caption{Overview of Direction-Aware Loss.}
\label{dir_loss}
\end{figure}

To incorporate return direction into the adaptation objective, we add a directional loss that explicitly penalizes predictions whose signs oppose the observed returns. Figure \ref{dir_loss} provides an overview.

We construct the directional loss $\mathcal{L}_\mathrm{dir}$ by taking the negative product of the predicted and observed returns, followed by a ReLU activation function. For batch index $b\in\{1,\ldots,B\}$, forecast position $h\in\{1,\ldots,H\}$, and target series $c\in\{1,\ldots,C\}$, the directional loss is
\begin{equation}
\mathcal{L}_\mathrm{dir} = \frac{1}{BHC} \sum_{b=1}^{B}\sum_{h=1}^{H}\sum_{c=1}^{C} \text{ReLU}(- \hat{r}_{b,h,c} \cdot r_{b,h,c}).
\end{equation}

When predicted and observed returns have matching signs, $\hat{r}_{b,h,c}\cdot r_{b,h,c}>0$. The negative product passed to ReLU is below zero, so the corresponding loss term is zero, yielding no penalty. For returns with opposing signs, $\hat{r}_{b,h,c} \cdot r_{b,h,c}<0$, and ReLU returns the positive penalty $\lvert\hat{r}_{b,h,c} \cdot r_{b,h,c}\rvert$, whose magnitude depends on both returns. 

\subsection{Price Calibration Loss}
\begin{figure}[!htbp]
\centering
\includegraphics[width=1.0\linewidth]{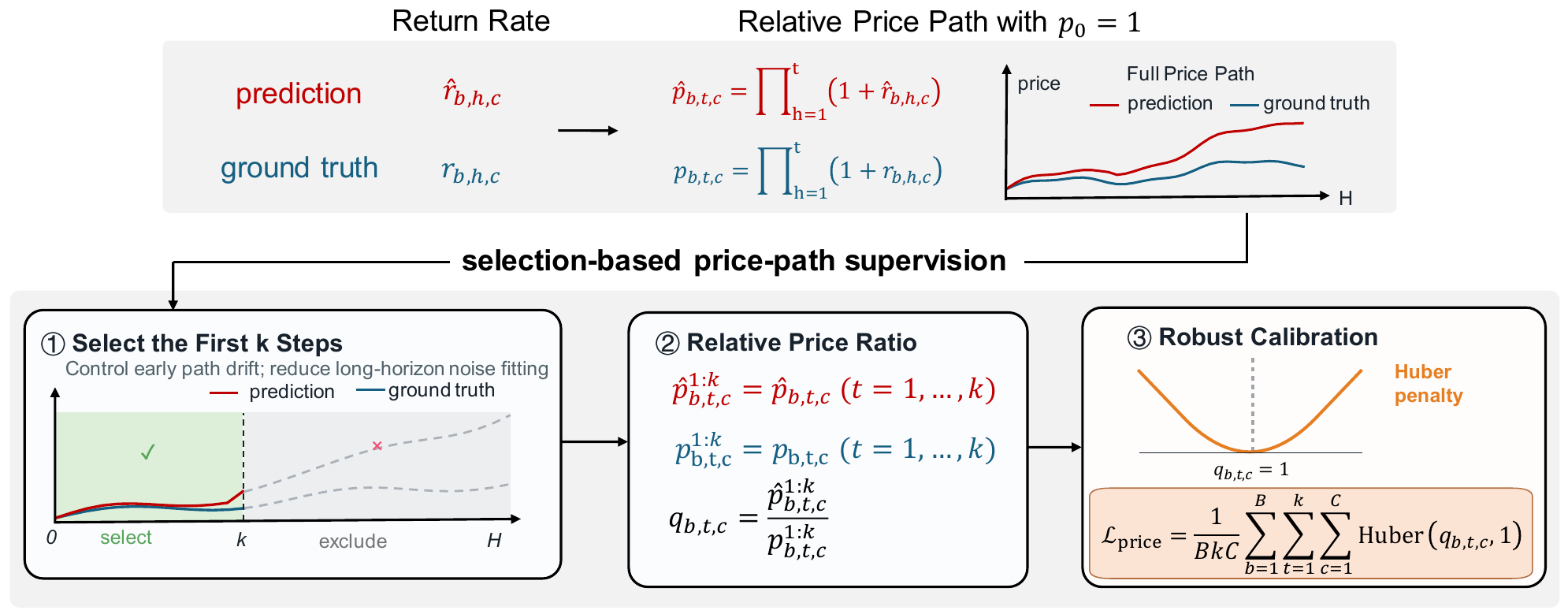}
\caption{Overview of Price Calibration Loss.}
\label{price_loss}
\vspace{-10pt}
\end{figure}

The directional loss penalizes opposite-sign predictions at individual forecast steps, but it does not account for how return errors accumulate over the forecasting horizon. This cumulative effect can be represented through the price trajectory reconstructed by compounding the sequence of returns. To account for this effect, we introduce a price calibration loss to reduce discrepancies between the trajectories reconstructed from predicted and observed returns. Both trajectories are initialized at 1, allowing the loss to measure discrepancies in compounded returns without depending on the nominal price of an asset. Figure \ref{price_loss} provides an overview.

By setting the initial relative price to 1 and compounding the predicted and observed returns in the original scale, we define the predicted and true price trajectories at step $t\in\{1,\ldots, H\}$, respectively, as 
\begin{equation}
\hat{p}_{b,t,c} = \prod_{h=1}^{t} (1 + \hat{r}_{b,h,c}), \quad p_{b,t,c} = \prod_{h=1}^{t} (1 + r_{b,h,c}).    
\end{equation}

\subsubsection{Early Forecast Step Selection}
Because prices are reconstructed by compounding predicted returns, each reconstructed price at a later forecast step depends on a longer sequence of return predictions. As a result, return errors can accumulate over the forecast horizon and lead to trajectory drift. To limit the influence of discrepancies arising after many compounding steps, we apply the price calibration loss only to the first $k$ forecasting steps. We denote the corresponding truncated trajectories by $\hat{p}^{1:k}_{b,t,c}$ and $p^{1:k}_{b,t,c}$. In the evaluated implementation, $k=15$ for $H=60$, and $k=30$ for $H\in\{90,120\}$. The candidate combinations and their comparative results are reported in Appendix~\ref{test_of_k}.

\subsubsection{Relative Price Trajectory Discrepancy}

To compare the predicted and observed reconstructed trajectories without directly depending on their absolute levels, we measure their discrepancy in relative terms. We define the relative price ratio as
\begin{equation}
q_{b,t,c}=\frac{\hat{p}^{1:k}_{b,t,c}}{p^{1:k}_{b,t,c}}.
\end{equation}
A value of $q_{b,t,c}=1$ indicates exact agreement between the two reconstructed prices, while deviations from 1 represent multiplicative discrepancies relative to the observed trajectory.

Compounding can occasionally produce large residuals in the relative price ratio, particularly when return deviations accumulate over multiple forecast steps. We therefore apply Huber loss, with the threshold parameter set to $\delta=1.0$, to the residual $q_{b,t,c}-1$. The quadratic region provides smooth sensitivity near the target, while the linear region limits the influence of large residuals.

For batch index $b\in\{1,\ldots,B\}$, forecast position $t\in\{1,\ldots,k\}$, and target series $c\in\{1,\ldots,C\}$, the price calibration loss is
\begin{equation}
\mathcal{L}_\mathrm{price}
=
\frac{1}{BkC}
\sum_{b=1}^{B}
\sum_{t=1}^{k}
\sum_{c=1}^{C}
\text{Huber}\left(q_{b,t,c},1\right).
\end{equation}

\subsection{Combined Adaptation Objective}
DPAL combines pointwise return fitting in normalized space with directional and price constraints computed from returns in the original scale:
\begin{equation}
\mathcal{L}_\mathrm{DPAL}
=
\mathcal{L}_\mathrm{MSE}
+
\alpha \mathcal{L}_\mathrm{dir}
+
\beta \mathcal{L}_\mathrm{price}.
\end{equation}
Here, $\alpha$ and $\beta$ control the contributions of the directional loss and price calibration loss, respectively. Because the three loss components differ in numerical scale, we use $\alpha=10^5$ and $\beta=10^2$ as a common weighting configuration throughout the main experiments. The same coefficients are used across all datasets, forecasting backbones, prediction horizons, and adaptation frameworks, without selecting coefficients separately for each setting. Section~\ref{Coefficient Robustness} examines the robustness of DPAL to alternative coefficient values.

\section{Experiments}
 
\subsection{Experimental Setup}

\subsubsection{Datasets}

We construct eight financial datasets spanning a range of equity and broader market settings. Seven are manually curated equity panels, while the remaining dataset contains market-level series from several asset classes. Complete ticker lists, preprocessing procedures, trading-date alignment, and chronological splits are provided in Appendix~\ref{dataset_details}.

\begin{enumerate}[label=\Alph*.]
    \item \textbf{Magnificent 7:} Seven widely followed U.S. mega-cap companies with substantial exposure to technology and technology-driven businesses.
    \item \textbf{S\&P 500 Excerpt:} Twenty constituents selected from the S\&P 500, covering a broader cross-section of established large-cap U.S. companies.
    \item \textbf{Russell 2000 Excerpt:} Twenty constituents selected from the Russell 2000 to represent the smaller-cap segment of the U.S. equity market.
    \item \textbf{Technology Sector:} Twenty technology-oriented companies spanning semiconductors, software, hardware, networking infrastructure, and digital services.
    \item \textbf{Consumer Sector:} Twenty companies from consumer-related industries, including consumer staples, retail, apparel, home improvement, and restaurants.
    \item \textbf{Energy Sector:} Twenty companies covering multiple parts of the energy industry, including integrated oil and gas, exploration and production, oilfield services, refining, and midstream operations.
    \item \textbf{Nasdaq Mixed Excerpt:} Twenty manually selected Nasdaq-listed companies from multiple industries, including health care, industrials, technology, communication services, consumer services, and utilities.
    \item \textbf{Macro and Market Series:} Ten market-level series comprising six major U.S. equity indices, one Treasury yield index, one currency index, and two commodity futures series.
\end{enumerate}

For all datasets, each constituent series is treated as an individual prediction target, with no cross-series aggregation or weighting. The targets are daily simple returns computed from adjusted closing values.

 \subsubsection{Forecasting Backbones and Comparison Settings}
 
We use five forecasting backbones: 
PatchTST \citep{PatchTST}, iTransformer \citep{iTransformer}, DLinear \citep{DLinear}, OLS \citep{OLS}, and FreTS \citep{FreTS}. This collection includes Transformer-based, linear, and frequency-domain forecasting models and allows us to examine DPAL  across different forecasting architectures.

For each backbone, we evaluate five settings. \textbf{Base} denotes the frozen pretrained source forecaster without test-time adaptation. \textbf{TAFAS} \citep{TAFAS} and \textbf{PETSA} \citep{PETSA} denote the corresponding test-time adaptation frameworks. In our evaluation, their updates use only completed forecast windows, and partially observed targets are not used for gradient updates. The remaining settings are \textbf{TAFAS with DPAL} and \textbf{PETSA with DPAL}. In both settings, DPAL is used as the adaptation objective while the corresponding adapter and source forecaster are retained.

\subsubsection{Implementation Details}
 
The historical input length is set to $L=60$ trading days, and the forecast horizon is $H\in\{60,90,120\}$ trading days. All five backbones use dataset-level standard scaling. TAFAS \citep{TAFAS} and PETSA \citep{PETSA} retain their original mechanisms for constructing adaptation batches from detected temporal periods, along with their framework-specific configurations.

\subsubsection{Evaluation Protocol}

The forecasting and evaluation pipeline uses two separate transformations. Dataset scale standardization is used during model estimation and history-conditioned rescaling is used during final evaluation. These transformations provide a fixed standardization scale for model estimation and incorporate recent return levels and variability into the mapping of predictions for evaluation, respectively.

Let $w\in\{1,\dots,W\}$ index rolling forecast windows, $i\in\{1,\dots,L\}$ index positions in the historical input, and $c\in\{1,\dots,C\}$ index target series. For each target series $c$, $\mu^{\mathrm{sc}}_{c}$ and $\sigma^{\mathrm{sc}}_{c}$ denote the mean and standard deviation, respectively, estimated exclusively from the training portion of the dataset. The historical input  $r^{\mathrm{hist}}_{w,i,c}$ is transformed by the standard scaler
\begin{equation}
r^{\mathrm{norm}}_{w,i,c}=\frac{r^{\mathrm{hist}}_{w,i,c}-\mu^{\mathrm{sc}}_{c}}{\sigma^{\mathrm{sc}}_{c}}.   
\end{equation}

The standardized history $r^{\mathrm{norm}}_{w,i,c}$ is then passed to the forecasting model to obtain the standardized return forecast $\hat r^{\mathrm{norm}}_{w,h,c}$, where $h\in\{1,\dots, H\}$. During adaptation, the normalized predicted and ground-truth returns are transformed back to the original scale using the scaler statistics $\mu^{\mathrm{sc}}_{c}$ and $\sigma^{\mathrm{sc}}_{c}$ to compute the direction-aware and price calibration losses. This ensures that the return signs and compounding operations in these losses are evaluated in the original return units rather than standardized values.

For final evaluation, we locally rescale the predicted returns with the aim of aligning their central level and variability with recent market conditions. We calculate the mean and standard deviation directly from the raw historical return window associated with each forecast window:
\begin{equation}
\mu^{\mathrm{hist}}_{w,c}=\frac{1}{L}\sum_{i=1}^{L}r^{\mathrm{hist}}_{w,i,c},\quad\sigma^{\mathrm{hist}}_{w,c}=\operatorname{Std}\!\left(r^{\mathrm{hist}}_{w,1:L,c}\right).
\end{equation}

The standardized prediction is locally rescaled for evaluation as
\begin{equation}
\label{eq:history_rescaling}
\hat r^{\mathrm{real}}_{w,h,c}=\hat r^{\mathrm{norm}}_{w,h,c}\sigma^{\mathrm{hist}}_{w,c}+\mu^{\mathrm{hist}}_{w,c}.
\end{equation}

\Eqref{eq:history_rescaling} rescales the standardized predictions using
the mean and standard deviation computed from the observed look-back
window and held fixed across the forecast horizon. The same rescaling
rule is applied to the predictions of all compared methods, while the
ground-truth returns used for evaluation are taken directly from the
original return data.

We evaluate reconstructed relative price trajectories and set their value at each forecast origin to 1 for scale-invariant evaluation across assets and rolling windows. Let $\hat r^{\mathrm{real}}_{w,h,c}$ and  $r^{\mathrm{real}}_{w,h,c}$ denote the predicted and observed future returns on the original return scale, respectively. For a forecast window, the predicted and observed relative prices at step $t$ are
\begin{equation}
\hat{p}_{w,t,c}=\prod_{h=1}^{t}\left(1+\hat r^{\mathrm{real}}_{w,h,c}\right),\qquad p_{w,t,c}=\prod_{h=1}^{t}\left(1+r^{\mathrm{real}}_{w,h,c}\right), \qquad t\in\{1,\dots,H\}.    
\end{equation}
For window index $w\in\{1,\dots,W\}$, forecast position $t\in\{1,\dots,H\}$, and target series $c\in\{1,\dots, C\}$, we compute the price MSE over the complete forecast horizon:
\begin{equation}
\text{MSE}_\mathrm{price} = \frac{1}{WHC} \sum_{w=1}^{W} \sum_{t=1}^{H} \sum_{c=1}^{C} (\hat{p}_{w,t,c} - p_{w,t,c})^2.   
\end{equation}

\subsection{Main Results}
\subsubsection{Forecasting Performance}
Tables \ref{main results tafas} and  \ref{main results petsa} report the MSE of reconstructed relative prices for experiments based on TAFAS \citep{TAFAS} and PETSA \citep{PETSA}, respectively. Lower values indicate better forecasting performance.

Using DPAL as the adaptation objective reduces reconstructed-price MSE in most configurations for both TAFAS \citep{TAFAS} and PETSA \citep{PETSA} across the evaluated datasets and all three forecast horizons. Each of the five forecasting backbones exhibits reductions in most configurations, suggesting that DPAL is not tied to a single adaptation mechanism. An ablation study evaluating the individual contributions of the direction-aware loss and the price calibration loss, as well as all possible combinations of the loss components, is provided in Appendix \ref{appendix:ablation}.

Additional quantitative results are provided in Appendix~\ref{appendix_additional_results}. Specifically, Appendix~\ref{appendix_relative_improvement} reports the relative improvements in reconstructed-price MSE across datasets, backbones, and forecast horizons, while Appendix~\ref{appendix_standardized_return_errors} presents supplementary MSE and MAE results computed directly in the standardized return space before inverse transformation. These results provide complementary views of the magnitude of the improvements in the primary evaluation metric and the pointwise accuracy of the predicted returns. As a robustness check, Appendix~\ref{appendix_scale_consistent_evaluation} additionally evaluates reconstructed-price MSE using the same training-set inverse standardization used to compute DPAL’s directional and price calibration losses during adaptation.

\begin{table}[t]
\centering
\caption{Forecasting performance comparison of TAFAS. Lower MSE is better. The best result within each backbone and horizon is highlighted in green.}
\label{main results tafas}
\renewcommand{\arraystretch}{1.2} 
\resizebox{\textwidth}{!}{
\begin{tabular}{cc|lcr|lcr|lcr|lcr|lcr}
\hline
\multirow{2}{*}{Datasets} & \multirow{2}{*}{$H$} & \multicolumn{3}{c|}{iTransformer} & \multicolumn{3}{c|}{DLinear} & \multicolumn{3}{c|}{FreTS} & \multicolumn{3}{c|}{PatchTST} & \multicolumn{3}{c}{OLS} \\
\cline{3-5} \cline{6-8} \cline{9-11} \cline{12-14} \cline{15-17}
 & & Base & TAFAS & \makecell{with\\DPAL} & Base & TAFAS & \makecell{with\\DPAL} & Base & TAFAS & \makecell{with\\DPAL} & Base & TAFAS & \makecell{with\\DPAL} & Base & TAFAS & \makecell{with\\DPAL} \\
\hline
\multirow{3}{*}{\makecell{Magnificent 7}} 
 & 60 & 1.617 & 1.137 & \textcolor{green}{0.547} & 0.754 & 0.742 & \textcolor{green}{0.477} & 0.530 & 0.528 & \textcolor{green}{0.476} & 1.502 & 0.764 & \textcolor{green}{0.489} & 1.784 & 1.118 & \textcolor{green}{0.515} \\ 
 & 90 & 6.164 & 3.634 & \textcolor{green}{1.482} & 2.062 & 1.993 & \textcolor{green}{0.943} & 1.152 & 1.152 & \textcolor{green}{0.933} & 5.265 & 2.293 & \textcolor{green}{1.139} & 5.761 & 2.864 & \textcolor{green}{1.379} \\ 
 & 120 & 17.717 & 9.087 & \textcolor{green}{2.780} & 3.879 & 3.685 & \textcolor{green}{1.503} & 2.030 & 2.032 & \textcolor{green}{1.616} & 13.147 & 5.506 & \textcolor{green}{2.075} & 15.585 & 6.792 & \textcolor{green}{3.004} \\ 
\hline
\multirow{3}{*}{\makecell{S\&P 500 \\ Excerpt}} 
 & 60 & 0.310 & 0.236 & \textcolor{green}{0.104} & 0.127 & 0.126 & \textcolor{green}{0.086} & 0.080 & 0.079 & \textcolor{green}{0.075} & 0.287 & 0.138 & \textcolor{green}{0.094} & 0.360 & 0.239 & \textcolor{green}{0.115} \\ 
 & 90 & 1.084 & 0.758 & \textcolor{green}{0.286} & 0.307 & 0.293 & \textcolor{green}{0.164} & 0.153 & 0.152 & \textcolor{green}{0.137} & 0.794 & 0.319 & \textcolor{green}{0.197} & 1.187 & 0.717 & \textcolor{green}{0.319} \\ 
 & 120 & 3.662 & 2.501 & \textcolor{green}{0.889} & 0.613 & 0.594 & \textcolor{green}{0.289} & 0.275 & 0.273 & \textcolor{green}{0.231} & 2.415 & 0.735 & \textcolor{green}{0.499} & 4.108 & 2.485 & \textcolor{green}{0.979} \\ 
\hline
\multirow{3}{*}{\makecell{Russell 2000 \\ Excerpt}} 
 & 60 & 0.305 & 0.223 & \textcolor{green}{0.105} & 0.123 & 0.126 & \textcolor{green}{0.073} & 0.068 & 0.069 & \textcolor{green}{0.066} & 0.237 & 0.100 & \textcolor{green}{0.071} & 0.310 & 0.149 & \textcolor{green}{0.081} \\ 
 & 90 & 0.699 & 0.408 & \textcolor{green}{0.153} & 0.296 & 0.287 & \textcolor{green}{0.143} & 0.143 & 0.143 & \textcolor{green}{0.127} & 0.622 & 0.231 & \textcolor{green}{0.135} & 0.795 & 0.352 & \textcolor{green}{0.150} \\ 
 & 120 & 1.454 & 0.919 & \textcolor{green}{0.339} & 0.503 & 0.493 & \textcolor{green}{0.262} & 0.246 & 0.246 & \textcolor{green}{0.226} & 1.198 & 0.535 & \textcolor{green}{0.279} & 1.586 & 0.834 & \textcolor{green}{0.331} \\ 
\hline
\multirow{3}{*}{\makecell{Technology \\ Sector}} 
 & 60 & 0.422 & 0.214 & \textcolor{green}{0.102} & 0.162 & 0.153 & \textcolor{green}{0.097} & 0.097 & 0.098 & \textcolor{green}{0.089} & 0.307 & 0.115 & \textcolor{green}{0.095} & 0.466 & 0.196 & \textcolor{green}{0.102} \\ 
 & 90 & 1.263 & 0.515 & \textcolor{green}{0.187} & 0.352 & 0.338 & \textcolor{green}{0.172} & 0.184 & 0.186 & \textcolor{green}{0.159} & 0.838 & 0.255 & \textcolor{green}{0.172} & 1.265 & 0.397 & \textcolor{green}{0.197} \\ 
 & 120 & 2.852 & 1.316 & \textcolor{green}{0.313} & 0.598 & 0.584 & \textcolor{green}{0.277} & 0.304 & 0.304 & \textcolor{green}{0.255} & 1.779 & 0.547 & \textcolor{green}{0.277} & 2.857 & 1.022 & \textcolor{green}{0.324} \\ 
\hline
\multirow{3}{*}{\makecell{Consumer \\ Sector}} 
 & 60 & 0.026 & 0.023 & \textcolor{green}{0.013} & 0.014 & 0.014 & \textcolor{green}{0.011} & 0.010 & 0.010 & \textcolor{green}{0.010} & 0.021 & 0.016 & \textcolor{green}{0.012} & 0.026 & 0.022 & \textcolor{green}{0.013} \\ 
 & 90 & 0.057 & 0.051 & \textcolor{green}{0.026} & 0.028 & 0.028 & \textcolor{green}{0.021} & 0.018 & 0.018 & \textcolor{green}{0.017} & 0.044 & 0.032 & \textcolor{green}{0.024} & 0.059 & 0.049 & \textcolor{green}{0.027} \\ 
 & 120 & 0.113 & 0.102 & \textcolor{green}{0.049} & 0.046 & 0.046 & \textcolor{green}{0.034} & 0.029 & 0.030 & \textcolor{green}{0.028} & 0.081 & 0.058 & \textcolor{green}{0.043} & 0.112 & 0.094 & \textcolor{green}{0.051} \\ 
\hline
\multirow{3}{*}{\makecell{Energy \\ Sector}} 
 & 60 & 0.037 & 0.036 & \textcolor{green}{0.024} & 0.027 & 0.027 & \textcolor{green}{0.023} & \textcolor{green}{0.019} & 0.019 & 0.019 & 0.036 & 0.030 & \textcolor{green}{0.023} & 0.039 & 0.035 & \textcolor{green}{0.024} \\ 
 & 90 & 0.082 & 0.079 & \textcolor{green}{0.048} & 0.055 & 0.054 & \textcolor{green}{0.044} & \textcolor{green}{0.035} & 0.035 & 0.036 & 0.075 & 0.060 & \textcolor{green}{0.046} & 0.085 & 0.076 & \textcolor{green}{0.049} \\ 
 & 120 & 0.150 & 0.146 & \textcolor{green}{0.086} & 0.093 & 0.093 & \textcolor{green}{0.073} & \textcolor{green}{0.059} & 0.059 & 0.060 & 0.143 & 0.110 & \textcolor{green}{0.085} & 0.158 & 0.143 & \textcolor{green}{0.091} \\ 
\hline
\multirow{3}{*}{\makecell{Nasdaq Mixed \\ Excerpt}} 
 & 60 & 0.058 & 0.051 & \textcolor{green}{0.031} & 0.035 & 0.035 & \textcolor{green}{0.028} & 0.026 & 0.026 & \textcolor{green}{0.026} & 0.052 & 0.036 & \textcolor{green}{0.029} & 0.059 & 0.045 & \textcolor{green}{0.032} \\ 
 & 90 & 0.125 & 0.110 & \textcolor{green}{0.059} & 0.071 & 0.070 & \textcolor{green}{0.051} & 0.049 & 0.049 & \textcolor{green}{0.044} & 0.109 & 0.071 & \textcolor{green}{0.053} & 0.128 & 0.094 & \textcolor{green}{0.063} \\ 
 & 120 & 0.223 & 0.197 & \textcolor{green}{0.104} & 0.113 & 0.112 & \textcolor{green}{0.079} & 0.077 & 0.077 & \textcolor{green}{0.068} & 0.192 & 0.124 & \textcolor{green}{0.090} & 0.228 & 0.169 & \textcolor{green}{0.107} \\ 
\hline
\multirow{3}{*}{\makecell{Macro and \\ Market Series}} 
 & 60 & 0.019 & 0.016 & \textcolor{green}{0.011} & 0.012 & 0.012 & \textcolor{green}{0.011} & \textcolor{green}{0.009} & 0.009 & 0.009 & 0.018 & 0.012 & \textcolor{green}{0.010} & 0.020 & 0.015 & \textcolor{green}{0.011} \\ 
 & 90 & 0.040 & 0.035 & \textcolor{green}{0.021} & 0.026 & 0.025 & \textcolor{green}{0.020} & \textcolor{green}{0.017} & 0.017 & 0.017 & 0.037 & 0.024 & \textcolor{green}{0.019} & 0.045 & 0.033 & \textcolor{green}{0.022} \\ 
 & 120 & 0.073 & 0.061 & \textcolor{green}{0.036} & 0.040 & 0.040 & \textcolor{green}{0.031} & \textcolor{green}{0.026} & 0.026 & 0.026 & 0.061 & 0.040 & \textcolor{green}{0.031} & 0.079 & 0.057 & \textcolor{green}{0.035} \\ 
\hline
\end{tabular}
}
\end{table}

\begin{table}[!htbp]
\centering
\caption{Forecasting performance comparison of PETSA. Lower MSE is better. The best result within each backbone and horizon is highlighted in green.}
\label{main results petsa}
\renewcommand{\arraystretch}{1.2} 
\resizebox{\textwidth}{!}{
\begin{tabular}{cc|lcr|lcr|lcr|lcr|lcr}
\hline
\multirow{2}{*}{Datasets} & \multirow{2}{*}{$H$} & \multicolumn{3}{c|}{iTransformer} & \multicolumn{3}{c|}{DLinear} & \multicolumn{3}{c|}{FreTS} & \multicolumn{3}{c|}{PatchTST} & \multicolumn{3}{c}{OLS} \\
\cline{3-5} \cline{6-8} \cline{9-11} \cline{12-14} \cline{15-17}
 & & Base & PETSA & \makecell{with\\DPAL} & Base & PETSA & \makecell{with\\DPAL} & Base & PETSA & \makecell{with\\DPAL} & Base & PETSA & \makecell{with\\DPAL} & Base & PETSA & \makecell{with\\DPAL} \\
\hline
\multirow{3}{*}{\makecell{Magnificent 7}} 
 & 60 & 1.617 & 1.079 & \textcolor{green}{0.782} & 0.754 & 0.765 & \textcolor{green}{0.502} & 0.530 & 0.578 & \textcolor{green}{0.469} & 1.502 & 0.709 & \textcolor{green}{0.489} & 1.784 & 1.107 & \textcolor{green}{0.646} \\ 
 & 90 & 6.164 & 3.459 & \textcolor{green}{1.847} & 2.062 & 1.826 & \textcolor{green}{1.013} & 1.152 & 1.199 & \textcolor{green}{0.888} & 5.265 & 1.968 & \textcolor{green}{1.106} & 5.761 & 2.454 & \textcolor{green}{1.299} \\ 
 & 120 & 17.717 & 7.588 & \textcolor{green}{3.900} & 3.879 & 3.603 & \textcolor{green}{1.552} & 2.030 & 2.142 & \textcolor{green}{1.463} & 13.147 & 5.283 & \textcolor{green}{2.654} & 15.585 & 5.352 & \textcolor{green}{2.783} \\ 
\hline
\multirow{3}{*}{\makecell{S\&P 500 \\ Excerpt}} 
 & 60 & 0.310 & 0.172 & \textcolor{green}{0.122} & 0.127 & 0.144 & \textcolor{green}{0.084} & 0.080 & 0.087 & \textcolor{green}{0.075} & 0.287 & 0.108 & \textcolor{green}{0.103} & 0.360 & 0.180 & \textcolor{green}{0.125} \\ 
 & 90 & 1.084 & 0.434 & \textcolor{green}{0.320} & 0.307 & 0.322 & \textcolor{green}{0.164} & 0.153 & 0.164 & \textcolor{green}{0.135} & 0.794 & 0.211 & \textcolor{green}{0.200} & 1.187 & 0.430 & \textcolor{green}{0.333} \\ 
 & 120 & 3.662 & 1.322 & \textcolor{green}{0.963} & 0.613 & 0.681 & \textcolor{green}{0.289} & 0.275 & 0.302 & \textcolor{green}{0.225} & 2.415 & 0.553 & \textcolor{green}{0.470} & 4.108 & 1.268 & \textcolor{green}{0.947} \\ 
\hline
\multirow{3}{*}{\makecell{Russell 2000 \\ Excerpt}} 
 & 60 & 0.305 & 0.192 & \textcolor{green}{0.118} & 0.123 & 0.138 & \textcolor{green}{0.075} & 0.068 & 0.081 & \textcolor{green}{0.065} & 0.237 & 0.106 & \textcolor{green}{0.084} & 0.310 & 0.142 & \textcolor{green}{0.083} \\ 
 & 90 & 0.699 & 0.610 & \textcolor{green}{0.151} & 0.296 & 0.241 & \textcolor{green}{0.137} & 0.143 & 0.168 & \textcolor{green}{0.127} & 0.622 & 0.218 & \textcolor{green}{0.147} & 0.795 & 0.287 & \textcolor{green}{0.160} \\ 
 & 120 & 1.454 & 0.767 & \textcolor{green}{0.365} & 0.503 & 0.425 & \textcolor{green}{0.251} & 0.246 & 0.289 & \textcolor{green}{0.225} & 1.198 & 0.450 & \textcolor{green}{0.298} & 1.586 & 0.632 & \textcolor{green}{0.331} \\ 
\hline
\multirow{3}{*}{\makecell{Technology \\ Sector}} 
 & 60 & 0.422 & 0.196 & \textcolor{green}{0.099} & 0.162 & 0.214 & \textcolor{green}{0.101} & 0.097 & 0.116 & \textcolor{green}{0.091} & 0.307 & 0.141 & \textcolor{green}{0.103} & 0.466 & 0.201 & \textcolor{green}{0.100} \\ 
 & 90 & 1.263 & 0.459 & \textcolor{green}{0.193} & 0.352 & 0.290 & \textcolor{green}{0.178} & 0.184 & 0.226 & \textcolor{green}{0.160} & 0.838 & 0.285 & \textcolor{green}{0.185} & 1.265 & 0.454 & \textcolor{green}{0.179} \\ 
 & 120 & 2.852 & 0.937 & \textcolor{green}{0.341} & 0.598 & 0.660 & \textcolor{green}{0.293} & 0.304 & 0.400 & \textcolor{green}{0.249} & 1.779 & 0.433 & \textcolor{green}{0.327} & 2.857 & 0.948 & \textcolor{green}{0.313} \\ 
\hline
\multirow{3}{*}{\makecell{Consumer \\ Sector}}
 & 60 & 0.026 & 0.020 & \textcolor{green}{0.013} & 0.014 & 0.016 & \textcolor{green}{0.011} & 0.010 & 0.012 & \textcolor{green}{0.010} & 0.021 & 0.016 & \textcolor{green}{0.012} & 0.026 & 0.021 & \textcolor{green}{0.013} \\ 
 & 90 & 0.057 & 0.045 & \textcolor{green}{0.026} & 0.028 & 0.031 & \textcolor{green}{0.021} & 0.018 & 0.022 & \textcolor{green}{0.017} & 0.044 & 0.031 & \textcolor{green}{0.022} & 0.059 & 0.043 & \textcolor{green}{0.026} \\ 
 & 120 & 0.113 & 0.087 & \textcolor{green}{0.047} & 0.046 & 0.049 & \textcolor{green}{0.033} & 0.029 & 0.035 & \textcolor{green}{0.026} & 0.081 & 0.054 & \textcolor{green}{0.038} & 0.112 & 0.078 & \textcolor{green}{0.046} \\ 
\hline
\multirow{3}{*}{\makecell{Energy \\ Sector}} 
 & 60 & 0.037 & 0.033 & \textcolor{green}{0.025} & 0.027 & 0.028 & \textcolor{green}{0.023} & \textcolor{green}{0.019} & 0.020 & 0.019 & 0.036 & 0.029 & \textcolor{green}{0.027} & 0.039 & 0.033 & \textcolor{green}{0.027} \\ 
 & 90 & 0.082 & 0.069 & \textcolor{green}{0.053} & 0.055 & 0.058 & \textcolor{green}{0.043} & \textcolor{green}{0.035} & 0.038 & 0.036 & 0.075 & 0.056 & \textcolor{green}{0.052} & 0.085 & 0.071 & \textcolor{green}{0.055} \\ 
 & 120 & 0.150 & 0.127 & \textcolor{green}{0.094} & 0.093 & 0.100 & \textcolor{green}{0.075} & \textcolor{green}{0.059} & 0.065 & 0.059 & 0.143 & 0.100 & \textcolor{green}{0.089} & 0.158 & 0.130 & \textcolor{green}{0.096} \\ 
\hline
\multirow{3}{*}{\makecell{Nasdaq Mixed \\ Excerpt}} 
 & 60 & 0.058 & 0.041 & \textcolor{green}{0.031} & 0.035 & 0.036 & \textcolor{green}{0.029} & 0.026 & 0.028 & \textcolor{green}{0.026} & 0.052 & 0.034 & \textcolor{green}{0.029} & 0.059 & 0.044 & \textcolor{green}{0.032} \\ 
 & 90 & 0.125 & 0.093 & \textcolor{green}{0.058} & 0.071 & 0.069 & \textcolor{green}{0.050} & 0.049 & 0.053 & \textcolor{green}{0.044} & 0.109 & 0.067 & \textcolor{green}{0.056} & 0.128 & 0.086 & \textcolor{green}{0.061} \\ 
 & 120 & 0.223 & 0.165 & \textcolor{green}{0.100} & 0.113 & 0.110 & \textcolor{green}{0.078} & 0.077 & 0.085 & \textcolor{green}{0.067} & 0.192 & 0.116 & \textcolor{green}{0.088} & 0.228 & 0.149 & \textcolor{green}{0.101} \\ 
\hline
\multirow{3}{*}{\makecell{Macro and \\ Market Series}}
 & 60 & 0.019 & 0.016 & \textcolor{green}{0.012} & 0.012 & 0.014 & \textcolor{green}{0.011} & \textcolor{green}{0.009} & 0.009 & 0.009 & 0.018 & 0.012 & \textcolor{green}{0.011} & 0.020 & 0.017 & \textcolor{green}{0.011} \\ 
 & 90 & 0.040 & 0.029 & \textcolor{green}{0.020} & 0.026 & 0.025 & \textcolor{green}{0.020} & \textcolor{green}{0.017} & 0.018 & 0.017 & 0.037 & 0.024 & \textcolor{green}{0.021} & 0.045 & 0.031 & \textcolor{green}{0.021} \\ 
 & 120 & 0.073 & 0.048 & \textcolor{green}{0.036} & 0.040 & 0.041 & \textcolor{green}{0.031} & \textcolor{green}{0.026} & 0.027 & 0.026 & 0.061 & 0.042 & \textcolor{green}{0.034} & 0.079 & 0.052 & \textcolor{green}{0.035} \\ 
\hline
\end{tabular}
}
\end{table}

\subsubsection{Rolling Price Trajectories}
\label{Representative Continuous Rolling Price Trajectories}

In addition to the window-level reconstructed-price MSE, we examine rolling price trajectories to show how return prediction errors accumulate across successive forecast blocks. At each forecast origin, the model predicts returns for the next $H$ trading days, after which the forecast origin is advanced by $H$ trading days to the next non-overlapping block. Unlike the primary evaluation metric, which resets the relative price to 1 at each forecast origin, this visualization initializes the relative price only once and then compounds predicted returns across consecutive blocks. This provides a complementary view of cumulative trajectory drift over the test period.

\begin{figure}[t]
\centering
\includegraphics[width=1.0\linewidth]{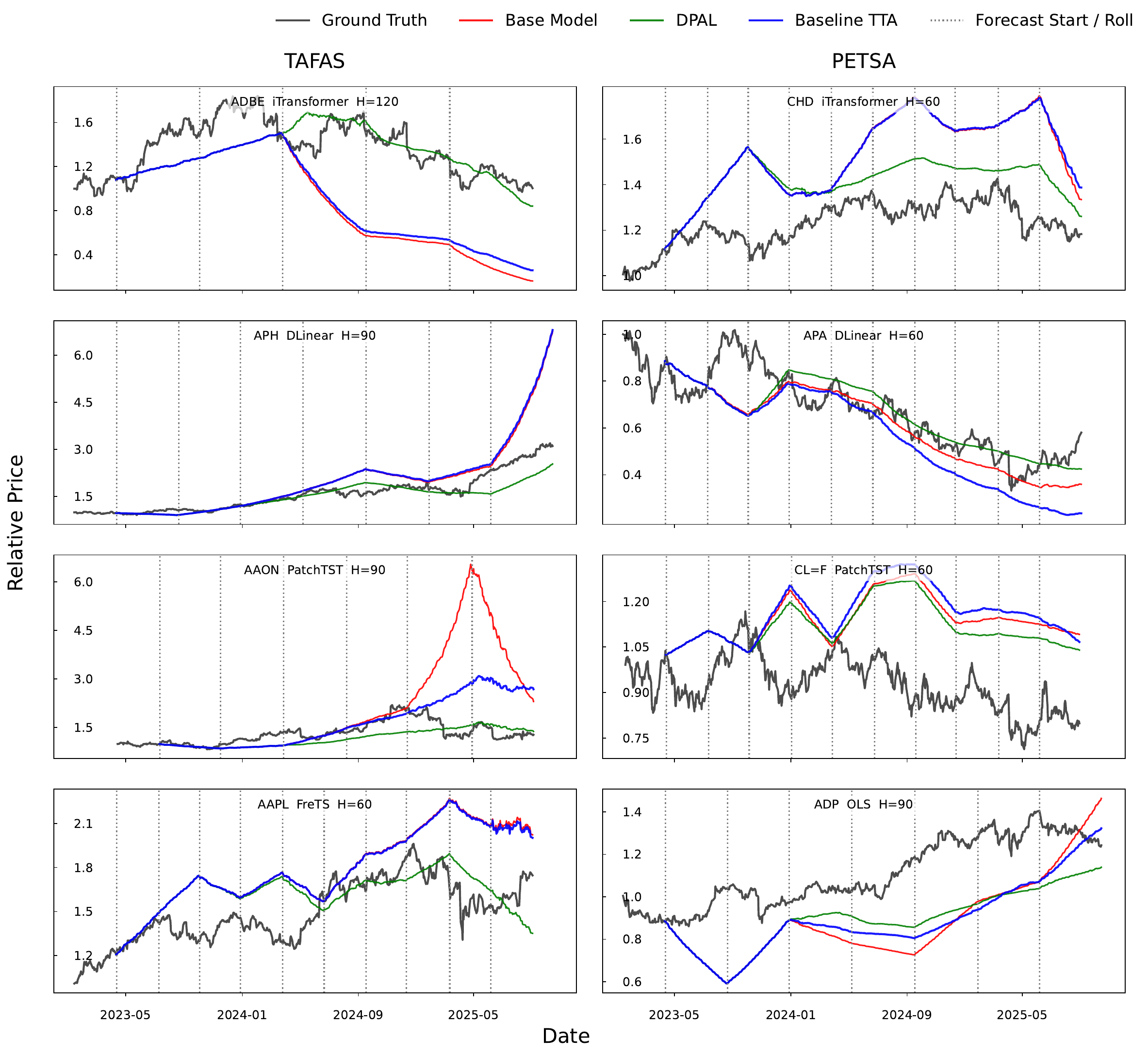}
\caption{Representative relative-price trajectories across consecutive, non-overlapping forecast blocks. The left and right columns show TAFAS and PETSA, respectively. The dark gray line denotes the ground truth, the red line denotes the frozen base model, the blue line denotes the baseline TTA method, and the green line denotes the corresponding TTA method with DPAL. Vertical dotted lines mark the forecast origins. Each price trajectory is initialized to $1$ only once at the beginning of the full sequence and is then accumulated across non-overlapping $H$-day forecast blocks.}
\vspace{-12pt}
\label{fig:representative_price_trajectories}
\end{figure}

Figure~\ref{fig:representative_price_trajectories} presents eight representative cases, with one case drawn from each dataset. The cases are selected to cover both adaptation frameworks and a range of forecasting backbones and horizons. For each dataset, we use the first series in the stored ordering as a fixed selection rule. Collectively, the eight cases cover all five forecasting backbones, all three forecasting horizons, $H\in\{60,90,120\}$, and both TTA frameworks, TAFAS \citep{TAFAS} and PETSA \citep{PETSA}.

In the displayed cases, trajectories produced with DPAL often remain closer to the observed price paths over successive forecast blocks. Compared with the base model, TAFAS and PETSA, the corresponding DPAL variants often show less cumulative drift. These qualitative patterns are consistent with the reductions in reconstructed-price MSE reported in the preceding tables, although the magnitude of the improvement varies across experimental settings.

\subsection{Robustness to Loss Coefficients}
\label{Coefficient Robustness}

To examine the robustness of DPAL to the choice of $\alpha$ and $\beta$, we evaluate $\alpha \in \{10^3, 10^4, 10^5, 10^6, 10^7\}$ and $\beta \in \{10^0, 10^1, 10^2, 10^3, 10^4\}$. Across all evaluated settings, we evaluate every $(\alpha,\beta)$ configuration while keeping all other training and evaluation settings unchanged.

Because reconstructed-price MSE can differ substantially in scale across datasets and forecasting settings, directly averaging raw MSE values would give disproportionate influence to settings with larger errors. We therefore compare each $(\alpha,\beta)$ configuration with the common configuration $(\alpha,\beta)=(10^5,10^2)$ used in the main experiments. For each setting $s$, the relative change is defined as
\begin{equation}
\Delta_s(\alpha,\beta)
=
\frac{\mathrm{MSE}_s(\alpha,\beta)}
{\mathrm{MSE}_s(10^5,10^2)}
-1.
\end{equation}

For each TTA framework, every table entry reports the median of $\Delta_s(\alpha,\beta)$ across all evaluated settings. Positive values indicate higher reconstructed-price MSE than the common configuration and therefore worse performance, whereas negative values indicate lower MSE and better performance. We use the median to reduce the influence of a small number of settings with unusually large relative changes. 

As shown in Table~\ref{tab:coefficient_robustness}, small values of $\alpha$, particularly $\alpha=10^3$, result in substantially higher relative MSE under both TAFAS \citep{TAFAS} and PETSA \citep{PETSA}. In contrast, most configurations with $\alpha\geq10^5$ perform similarly to the common configuration used in the main experiments. Although a few configurations yield slight improvements, the median improvements remain below $0.5\%$. Overall, the results indicate a broad region of stable performance around the coefficient scales used in the main experiments.

\begin{table}[!htbp]
\centering
\renewcommand{\arraystretch}{1.3} 

\begin{minipage}{0.48\textwidth}
\centering
\resizebox{\linewidth}{!}{
\begin{tabular}{c|ccccc}
\multicolumn{6}{c}{\textbf{(a) TAFAS}} \\[-0.3ex]
\multicolumn{6}{c}{Median MSE change relative to the common configuration} \\ \hline

\textbf{$\bm{\alpha \backslash \beta}$} & \textbf{$10^0$} & \textbf{$10^1$} & \textbf{$10^2$} & \textbf{$10^3$} & \textbf{$10^4$} \\ \hline

\textbf{$10^3$} & \cellcolor[HTML]{ED6A6A}20.01\% & \cellcolor[HTML]{F08484}18.43\% & \cellcolor[HTML]{E85757}23.05\% & \cellcolor[HTML]{E95A5A}22.75\% & \cellcolor[HTML]{E95B5B}22.63\% \\
\textbf{$10^4$} & \cellcolor[HTML]{FCE0E0}4.58\% & \cellcolor[HTML]{FCE6E6}3.90\% & \cellcolor[HTML]{F7AAAA}7.46\% & \cellcolor[HTML]{F37E7E}13.25\% & \cellcolor[HTML]{F07272}14.73\% \\ \hhline{~|~~|-|~~}
\textbf{$10^5$} & \cellcolor[HTML]{FFFFFF}0.66\% & \cellcolor[HTML]{FFFFFF}0.61\% & \multicolumn{1}{|c|}{\cellcolor[HTML]{FFFFFF}0.00\%} & \cellcolor[HTML]{F9E8E8}2.46\% & \cellcolor[HTML]{F7A5A5}7.80\% \\ \hhline{~|~~|-|~~}
\textbf{$10^6$} & \cellcolor[HTML]{FFFFFF}0.27\% & \cellcolor[HTML]{FFFFFF}0.22\% & \cellcolor[HTML]{FFFFFF}0.12\% & \cellcolor[HTML]{E2EFDA}-0.43\% & \cellcolor[HTML]{FCECEC}2.07\% \\
\textbf{$10^7$} & \cellcolor[HTML]{FFFFFF}0.16\% & \cellcolor[HTML]{FFFFFF}0.17\% & \cellcolor[HTML]{FFFFFF}0.17\% & \cellcolor[HTML]{FFFFFF}0.03\% & \cellcolor[HTML]{E2EFDA}-0.39\% \\ \hline
\end{tabular}
}
\end{minipage}
\hfill 
\begin{minipage}{0.48\textwidth}
\centering
\resizebox{\linewidth}{!}{
\begin{tabular}{c|ccccc}
\multicolumn{6}{c}{\textbf{(b) PETSA}} \\[-0.3ex]
\multicolumn{6}{c}{Median MSE change relative to the common configuration} \\ \hline

\textbf{$\bm{\alpha \backslash \beta}$} & \textbf{$10^0$} & \textbf{$10^1$} & \textbf{$10^2$} & \textbf{$10^3$} & \textbf{$10^4$} \\ \hline

\textbf{$10^3$} & \cellcolor[HTML]{F08484}13.06\% & \cellcolor[HTML]{F37E7E}11.93\% & \cellcolor[HTML]{F08484}13.51\% & \cellcolor[HTML]{ED6A6A}15.83\% & \cellcolor[HTML]{ED6A6A}15.50\% \\
\textbf{$10^4$} & \cellcolor[HTML]{FCE0E0}2.63\% & \cellcolor[HTML]{FCE6E6}1.99\% & \cellcolor[HTML]{FCE0E0}3.87\% & \cellcolor[HTML]{F7AAAA}7.90\% & \cellcolor[HTML]{F37E7E}8.81\% \\ \hhline{~|~~|-|~~}
\textbf{$10^5$} & \cellcolor[HTML]{FFFFFF}0.62\% & \cellcolor[HTML]{FFFFFF}0.50\% & \multicolumn{1}{|c|}{\cellcolor[HTML]{FFFFFF}0.00\%} & \cellcolor[HTML]{FCE6E6}1.90\% & \cellcolor[HTML]{F7A5A5}5.79\% \\ \hhline{~|~~|-|~~}
\textbf{$10^6$} & \cellcolor[HTML]{FFFFFF}0.30\% & \cellcolor[HTML]{FFFFFF}0.21\% & \cellcolor[HTML]{FFFFFF}0.24\% & \cellcolor[HTML]{E2EFDA}-0.18\% & \cellcolor[HTML]{FCE6E6}1.77\% \\
\textbf{$10^7$} & \cellcolor[HTML]{FFFFFF}0.19\% & \cellcolor[HTML]{FFFFFF}0.07\% & \cellcolor[HTML]{FFFFFF}0.09\% & \cellcolor[HTML]{FFFFFF}0.23\% & \cellcolor[HTML]{E2EFDA}-0.17\% \\ \hline
\end{tabular}
}
\end{minipage}
\caption{Robustness of DPAL to the loss coefficients $\alpha$ and $\beta$ under (a) TAFAS and (b) PETSA. Each cell reports the median relative change in reconstructed-price MSE, expressed as a percentage, from the common configuration $(\alpha,\beta)=(10^5,10^2)$ used in the main experiments across all evaluated settings within the corresponding TTA framework. Positive values indicate higher MSE and worse performance relative to the common configuration, whereas negative values indicate lower MSE and better performance. The outlined cells identify the common configuration, and the red and green shading denote performance degradation and improvement, respectively.}
\label{tab:coefficient_robustness}
\end{table}

To further assess the robustness of the common configuration, we compare it with the lowest MSE obtained for each evaluated setting across all coefficient configurations. Specifically, we define
\begin{equation}
G_s = \frac{\mathrm{MSE}_s(10^5,10^2)}{\min_{\alpha,\beta}\mathrm{MSE}_s(\alpha,\beta)} - 1.
\end{equation}

\begin{table}[!htbp]
\centering
\renewcommand{\arraystretch}{1.15}
\begin{tabular}{lcccc}
\toprule
\textbf{Framework}
& \textbf{Settings}
& \textbf{$G_s \leq 0.05$}
& \textbf{Share}
& \textbf{Median $G_s$} \\
\midrule
TAFAS    & 120 & 108 & 90.0\% & 0.98\% \\
PETSA    & 120 &  94 & 78.3\% & 1.54\% \\
\midrule
Combined & 240 & 202 & 84.2\% & 1.30\% \\
\bottomrule
\end{tabular}
\caption{Comparison between the common configuration $(\alpha,\beta)=(10^5,10^2)$ and the lowest-MSE configuration for each setting. The third column reports the number of
settings with $G_s \leq 0.05$, corresponding to a reconstructed-price
MSE no more than 5\% higher than the lowest MSE.}
\label{tab:reference_grid_gap}
\end{table}

We regard the common configuration as being close to the lowest-MSE configuration for setting $s$ when $G_s \leq 0.05$. For each TTA framework, the evaluation covers eight datasets, five forecasting backbones, and three prediction horizons. Table~\ref{tab:reference_grid_gap} summarizes the number and share of settings satisfying this criterion, together with the median $G_s$.

Across all evaluated settings, the common configuration is within 5\% of the lowest MSE in 84.2\% of cases, with a median $G_s$ of 1.30\%. These results indicate that DPAL generally remains robust to the choice of loss coefficients.

\section{Conclusions, Limitations, and Future Work}

\textbf{Conclusions.}
This work introduces Direction- and Price-Aware Adaptation Loss (DPAL), an adaptation objective for financial time series forecasting that combines pointwise return fitting with directional and price-calibration terms. DPAL is used as the adaptation objective in TAFAS \citep{TAFAS} and PETSA \citep{PETSA} without modifying the source forecaster or the adapter architecture. Across the evaluated settings, using DPAL reduces reconstructed-price MSE in most configurations for both TAFAS and PETSA. The rolling price trajectories further show that the corresponding DPAL variants often stay closer to the ground-truth price paths with less cumulative drift over successive forecast blocks. Overall, these results indicate that DPAL can improve test-time adaptation for financial time series forecasting under the evaluated settings.

\textbf{Limitations.}
Although the evaluation covers multiple financial datasets, it remains limited to the markets and time periods represented in the collected data. In addition, the current formulation of DPAL uses relatively simple directional and price-calibration terms.

\textbf{Future Work.}
Future work could extend the evaluation to broader market regimes and longer time periods, and explore more flexible directional and price-calibration terms. DPAL could also be studied in higher-frequency financial forecasting settings, where return dynamics and adaptation timing may differ from those considered here.

\bibliographystyle{tmlr}
\bibliography{stock} 

\clearpage
\appendix
\section{Additional Experimental Results}
\label{appendix:additional_results}

\subsection{Ablation Study}
\label{appendix:ablation}
To examine the individual contributions of the proposed direction-aware loss ($\mathcal{L}_\mathrm{dir}$) and price calibration loss ($\mathcal{L}_\mathrm{price}$), we conducted an ablation study. We evaluated all possible combinations of the three loss components. Specifically, we tested seven loss configurations: standard Mean Squared Error alone ($\mathrm{MSE}$), direction-aware loss alone ($\alpha\mathcal{L}_\mathrm{dir}$), price calibration loss alone ($\beta\mathcal{L}_\mathrm{price}$), their pairwise combinations ($\mathrm{MSE} + \alpha\mathcal{L}_\mathrm{dir}$, $\mathrm{MSE} + \beta\mathcal{L}_\mathrm{price}$, $\alpha\mathcal{L}_\mathrm{dir} + \beta\mathcal{L}_\mathrm{price}$), and the full joint formulation ($\mathrm{MSE} + \alpha\mathcal{L}_\mathrm{dir} + \beta\mathcal{L}_\mathrm{price}$). These configurations were evaluated across different datasets, backbone models, and forecast horizons under both the TAFAS \citep{TAFAS} and PETSA \citep{PETSA} frameworks. Tables \ref{TAFAS magnificent 7}--\ref{PETSA index} report the results.

\begin{table}[htbp]
\renewcommand{\thetable}{A.1}
\centering
\caption{Ablation results for different combinations of the loss components under the TAFAS and PETSA frameworks. Base denotes the frozen backbone without test-time adaptation. The check-marked columns represent the seven loss configurations. In the PETSA tables, the additional column labeled \textit{original} reports results using PETSA's original adaptation objective. Each entry reports the reconstructed price MSE for a given backbone model and forecast horizon $H$, with lower values indicating better performance. The best result (lowest MSE) in each row is highlighted in green, and the second-best result (second-lowest MSE) is marked in blue. All values are rounded to five decimal places, whereas the rankings are determined using the original unrounded results.} 
\label{ablation_study_results}

\begin{subtable}{\textwidth}
\renewcommand{\thetable}{A.1.1}
\centering
\vspace{1.5em}
\caption{TAFAS Magnificent 7}
\label{TAFAS magnificent 7}
\renewcommand{\arraystretch}{1.2} 
\resizebox{\textwidth}{!}{
\begin{tabular}{cc|ccccccccc}
\hline
\multirow{3}{*}{Model} & \multirow{3}{*}{$H$} & \multirow{3}{*}{Base} & mse & $\checkmark$ & & & $\checkmark$ & $\checkmark$ & & $\checkmark$ \\
 & & & dir & & $\checkmark$ & & $\checkmark$ & & $\checkmark$ & $\checkmark$ \\
 & & & price & & & $\checkmark$ & & $\checkmark$ & $\checkmark$ & $\checkmark$ \\
\hline
\multirow{3}{*}{iTransformer} 
 & 60 & 1.61743 & & 1.13657 & 0.69084 & 1.01203 & 0.54975 & 0.74609 & \textcolor{green}{0.54121} & \textcolor{blue}{0.54736} \\ 
 & 90 & 6.16393 & & 3.63401 & 1.86856 & 2.85430 & 1.51084 & 2.20571 & \textcolor{green}{1.47397} & \textcolor{blue}{1.48232} \\ 
 & 120 & 17.71737 & & 9.08681 & 4.47373 & 8.62511 & 2.80610 & 5.23148 & \textcolor{green}{2.77784} & \textcolor{blue}{2.77978} \\ 
\hline
\multirow{3}{*}{DLinear} 
 & 60 & 0.75416 & & 0.74219 & 0.51022 & 0.70659 & 0.47830 & 0.69959 & \textcolor{green}{0.47640} & \textcolor{blue}{0.47661} \\ 
 & 90 & 2.06209 & & 1.99329 & 1.10354 & 1.54488 & 0.98637 & 1.52346 & \textcolor{green}{0.94325} & \textcolor{blue}{0.94340} \\ 
 & 120 & 3.87860 & & 3.68547 & 1.95164 & 3.06590 & 1.52380 & 2.93949 & \textcolor{green}{1.49955} & \textcolor{blue}{1.50322} \\ 
\hline
\multirow{3}{*}{FreTS} 
 & 60 & 0.52950 & & 0.52818 & 0.48641 & 0.50994 & 0.47605 & 0.50921 & \textcolor{green}{0.47509} & \textcolor{blue}{0.47602} \\ 
 & 90 & 1.15248 & & 1.15159 & 1.01343 & 1.09132 & 0.94249 & 1.08163 & \textcolor{green}{0.93252} & \textcolor{blue}{0.93348} \\ 
 & 120 & 2.03039 & & 2.03178 & 1.82891 & 1.98825 & \textcolor{green}{1.57614} & 1.95812 & \textcolor{blue}{1.60539} & 1.61589 \\ 
\hline
\multirow{3}{*}{PatchTST} 
 & 60 & 1.50215 & & 0.76392 & 0.60709 & 0.93288 & 0.49720 & 0.54765 & \textcolor{blue}{0.49001} & \textcolor{green}{0.48863} \\ 
 & 90 & 5.26547 & & 2.29294 & 1.61467 & 2.07171 & 1.19600 & 1.25270 & \textcolor{blue}{1.14013} & \textcolor{green}{1.13853} \\ 
 & 120 & 13.14714 & & 5.50601 & 3.52361 & 6.08743 & 2.15644 & 3.06051 & \textcolor{blue}{2.07796} & \textcolor{green}{2.07467} \\ 
\hline
\multirow{3}{*}{OLS} 
 & 60 & 1.78386 & & 1.11830 & 0.61121 & 0.89851 & 0.52656 & 0.69139 & \textcolor{blue}{0.51812} & \textcolor{green}{0.51547} \\ 
 & 90 & 5.76070 & & 2.86407 & 1.59095 & 1.97215 & \textcolor{blue}{1.38104} & 1.53145 & 1.38284 & \textcolor{green}{1.37908} \\ 
 & 120 & 15.58452 & & 6.79152 & 3.84079 & 6.31496 & \textcolor{green}{2.92593} & 4.29467 & 3.01510 & \textcolor{blue}{3.00353} \\ 
\hline
\end{tabular}
}
\end{subtable}
\end{table}

\begin{table}[htbp]
\ContinuedFloat
\centering
\begin{subtable}{\textwidth}
\renewcommand{\thetable}{A.1.2}
\centering
\caption{TAFAS S\&P 500 Excerpt}
\label{TAFAS sp500}
\renewcommand{\arraystretch}{1.2} 
\resizebox{\textwidth}{!}{
\begin{tabular}{cc|ccccccccc}
\hline
\multirow{3}{*}{Model} & \multirow{3}{*}{$H$} & \multirow{3}{*}{Base} & mse & $\checkmark$ & & & $\checkmark$ & $\checkmark$ & & $\checkmark$ \\
 & & & dir & & $\checkmark$ & & $\checkmark$ & & $\checkmark$ & $\checkmark$ \\
 & & & price & & & $\checkmark$ & & $\checkmark$ & $\checkmark$ & $\checkmark$ \\
\hline
\multirow{3}{*}{iTransformer} 
 & 60 & 0.31049 & & 0.23600 & 0.13022 & 0.18126 & \textcolor{blue}{0.10350} & 0.13573 & \textcolor{green}{0.10346} & 0.10370 \\ 
 & 90 & 1.08383 & & 0.75797 & 0.32101 & 0.57364 & \textcolor{green}{0.28542} & 0.45288 & 0.28627 & \textcolor{blue}{0.28569} \\ 
 & 120 & 3.66154 & & 2.50056 & 0.92215 & 2.32753 & \textcolor{green}{0.88228} & 1.72442 & 0.89078 & \textcolor{blue}{0.88937} \\ 
\hline
\multirow{3}{*}{DLinear} 
 & 60 & 0.12715 & & 0.12572 & 0.09147 & 0.11144 & 0.08590 & 0.10870 & \textcolor{green}{0.08572} & \textcolor{blue}{0.08573} \\ 
 & 90 & 0.30746 & & 0.29315 & 0.17772 & 0.22061 & 0.16674 & 0.21476 & \textcolor{blue}{0.16449} & \textcolor{green}{0.16436} \\ 
 & 120 & 0.61329 & & 0.59441 & 0.31469 & 0.45443 & 0.29504 & 0.43984 & \textcolor{blue}{0.28896} & \textcolor{green}{0.28857} \\ 
\hline
\multirow{3}{*}{FreTS} 
 & 60 & 0.07956 & & 0.07926 & 0.07572 & 0.07905 & \textcolor{green}{0.07489} & 0.07914 & \textcolor{blue}{0.07518} & 0.07525 \\ 
 & 90 & 0.15343 & & 0.15248 & 0.14383 & 0.15360 & 0.13732 & 0.15337 & \textcolor{blue}{0.13696} & \textcolor{green}{0.13695} \\ 
 & 120 & 0.27454 & & 0.27336 & 0.25104 & 0.27378 & 0.23120 & 0.27318 & \textcolor{green}{0.23000} & \textcolor{blue}{0.23053} \\ 
\hline
\multirow{3}{*}{PatchTST} 
 & 60 & 0.28742 & & 0.13806 & 0.09950 & 0.16196 & \textcolor{blue}{0.09453} & 0.09919 & 0.09454 & \textcolor{green}{0.09428} \\ 
 & 90 & 0.79353 & & 0.31934 & 0.19844 & 0.48195 & \textcolor{green}{0.19618} & 0.27224 & 0.19784 & \textcolor{blue}{0.19696} \\ 
 & 120 & 2.41472 & & 0.73501 & \textcolor{green}{0.44984} & 1.54950 & \textcolor{blue}{0.48298} & 0.70732 & 0.50250 & 0.49946 \\ 
\hline
\multirow{3}{*}{OLS} 
 & 60 & 0.36039 & & 0.23918 & 0.13081 & 0.20259 & 0.11541 & 0.15304 & \textcolor{blue}{0.11533} & \textcolor{green}{0.11527} \\ 
 & 90 & 1.18749 & & 0.71727 & 0.34031 & 0.69211 & \textcolor{green}{0.31792} & 0.52828 & 0.31934 & \textcolor{blue}{0.31922} \\ 
 & 120 & 4.10791 & & 2.48495 & 0.98300 & 3.34456 & \textcolor{green}{0.96449} & 2.53727 & \textcolor{blue}{0.97768} & 0.97911 \\ 
\hline
\end{tabular}
}
\end{subtable}
\vspace{1.5em}

\begin{subtable}{\textwidth}
\renewcommand{\thetable}{A.1.3}
\centering
\caption{TAFAS Russell 2000 Excerpt}
\label{TAFAS russell2000}
\renewcommand{\arraystretch}{1.2} 
\resizebox{\textwidth}{!}{
\begin{tabular}{cc|ccccccccc}
\hline
\multirow{3}{*}{Model} & \multirow{3}{*}{$H$} & \multirow{3}{*}{Base} & mse & $\checkmark$ & & & $\checkmark$ & $\checkmark$ & & $\checkmark$ \\
 & & & dir & & $\checkmark$ & & $\checkmark$ & & $\checkmark$ & $\checkmark$ \\
 & & & price & & & $\checkmark$ & & $\checkmark$ & $\checkmark$ & $\checkmark$ \\
\hline
\multirow{3}{*}{iTransformer} 
 & 60 & 0.30491 & & 0.22266 & 0.10367 & 0.22091 & \textcolor{blue}{0.09676} & 0.14446 & \textcolor{green}{0.09657} & 0.10505 \\ 
 & 90 & 0.69923 & & 0.40823 & 0.21734 & 0.40426 & 0.16352 & 0.27378 & \textcolor{green}{0.15242} & \textcolor{blue}{0.15310} \\ 
 & 120 & 1.45384 & & 0.91903 & 0.50233 & 1.02163 & 0.35110 & 0.73352 & \textcolor{green}{0.33515} & \textcolor{blue}{0.33906} \\ 
\hline
\multirow{3}{*}{DLinear} 
 & 60 & 0.12283 & & 0.12568 & 0.08020 & 0.11390 & 0.07334 & 0.11505 & \textcolor{green}{0.07317} & \textcolor{blue}{0.07326} \\ 
 & 90 & 0.29572 & & 0.28654 & 0.17123 & 0.23883 & 0.14525 & 0.23303 & \textcolor{green}{0.14292} & \textcolor{blue}{0.14326} \\ 
 & 120 & 0.50299 & & 0.49319 & 0.30670 & 0.42453 & 0.26384 & 0.41608 & \textcolor{green}{0.26156} & \textcolor{blue}{0.26247} \\ 
\hline
\multirow{3}{*}{FreTS} 
 & 60 & 0.06752 & & 0.06948 & \textcolor{green}{0.06519} & 0.06822 & \textcolor{blue}{0.06544} & 0.06997 & 0.06581 & 0.06588 \\ 
 & 90 & 0.14269 & & 0.14350 & 0.13450 & 0.14204 & 0.12784 & 0.14292 & \textcolor{green}{0.12694} & \textcolor{blue}{0.12726} \\ 
 & 120 & 0.24604 & & 0.24637 & 0.23695 & 0.24966 & 0.22598 & 0.25160 & \textcolor{green}{0.22461} & \textcolor{blue}{0.22559} \\ 
\hline
\multirow{3}{*}{PatchTST} 
 & 60 & 0.23714 & & 0.10034 & 0.08920 & 0.16727 & 0.07233 & 0.08412 & \textcolor{blue}{0.07150} & \textcolor{green}{0.07135} \\ 
 & 90 & 0.62163 & & 0.23122 & 0.20489 & 0.36289 & 0.14232 & 0.17970 & \textcolor{blue}{0.13505} & \textcolor{green}{0.13505} \\ 
 & 120 & 1.19838 & & 0.53473 & 0.45986 & 0.76042 & 0.30306 & 0.37024 & \textcolor{blue}{0.27931} & \textcolor{green}{0.27867} \\ 
\hline
\multirow{3}{*}{OLS} 
 & 60 & 0.30977 & & 0.14949 & 0.09819 & 0.14626 & \textcolor{green}{0.07953} & 0.11407 & \textcolor{blue}{0.08022} & 0.08062 \\ 
 & 90 & 0.79484 & & 0.35215 & 0.22567 & 0.21361 & 0.15854 & 0.19225 & \textcolor{green}{0.14915} & \textcolor{blue}{0.14956} \\ 
 & 120 & 1.58614 & & 0.83380 & 0.50402 & 0.50750 & 0.34201 & 0.44580 & \textcolor{green}{0.33023} & \textcolor{blue}{0.33076} \\ 
\hline
\end{tabular}
}
\end{subtable}
\end{table}

\begin{table}[htbp]
\ContinuedFloat
\centering
\begin{subtable}{\textwidth}
\renewcommand{\thetable}{A.1.4}
\centering
\caption{TAFAS Technology Sector}
\label{TAFAS tech stock}
\renewcommand{\arraystretch}{1.2} 
\resizebox{\textwidth}{!}{
\begin{tabular}{cc|ccccccccc}
\hline
\multirow{3}{*}{Model} & \multirow{3}{*}{$H$} & \multirow{3}{*}{Base} & mse & $\checkmark$ & & & $\checkmark$ & $\checkmark$ & & $\checkmark$ \\
 & & & dir & & $\checkmark$ & & $\checkmark$ & & $\checkmark$ & $\checkmark$ \\
 & & & price & & & $\checkmark$ & & $\checkmark$ & $\checkmark$ & $\checkmark$ \\
\hline
\multirow{3}{*}{iTransformer} 
 & 60 & 0.42161 & & 0.21446 & 0.10671 & 0.19210 & 0.10269 & 0.15871 & \textcolor{green}{0.10198} & \textcolor{blue}{0.10243} \\ 
 & 90 & 1.26285 & & 0.51497 & 0.22179 & 0.37666 & 0.19041 & 0.29334 & \textcolor{blue}{0.18713} & \textcolor{green}{0.18650} \\ 
 & 120 & 2.85160 & & 1.31550 & 0.46201 & 0.94775 & 0.32796 & 0.67012 & \textcolor{green}{0.31184} & \textcolor{blue}{0.31325} \\ 
\hline
\multirow{3}{*}{DLinear} 
 & 60 & 0.16194 & & 0.15340 & 0.10161 & 0.14902 & 0.09750 & 0.14047 & \textcolor{blue}{0.09675} & \textcolor{green}{0.09652} \\ 
 & 90 & 0.35196 & & 0.33752 & 0.19132 & 0.24699 & 0.17665 & 0.24529 & \textcolor{green}{0.17148} & \textcolor{blue}{0.17167} \\ 
 & 120 & 0.59787 & & 0.58418 & 0.32343 & 0.43689 & 0.27900 & 0.44063 & \textcolor{green}{0.27600} & \textcolor{blue}{0.27659} \\ 
\hline
\multirow{3}{*}{FreTS} 
 & 60 & 0.09694 & & 0.09827 & 0.08959 & 0.09614 & \textcolor{blue}{0.08929} & 0.09680 & \textcolor{green}{0.08925} & 0.08936 \\ 
 & 90 & 0.18422 & & 0.18591 & 0.16319 & 0.18401 & 0.15936 & 0.18556 & \textcolor{green}{0.15860} & \textcolor{blue}{0.15911} \\ 
 & 120 & 0.30389 & & 0.30402 & 0.26617 & 0.30341 & \textcolor{blue}{0.25491} & 0.30671 & \textcolor{green}{0.25461} & 0.25510 \\ 
\hline
\multirow{3}{*}{PatchTST} 
 & 60 & 0.30671 & & 0.11481 & 0.10238 & 0.15836 & 0.09522 & 0.10280 & \textcolor{blue}{0.09508} & \textcolor{green}{0.09506} \\ 
 & 90 & 0.83752 & & 0.25486 & 0.19470 & 0.41181 & \textcolor{green}{0.17171} & 0.21413 & \textcolor{blue}{0.17211} & 0.17237 \\ 
 & 120 & 1.77862 & & 0.54692 & 0.37739 & 0.72563 & 0.29102 & 0.31431 & \textcolor{green}{0.27723} & \textcolor{blue}{0.27726} \\ 
\hline
\multirow{3}{*}{OLS} 
 & 60 & 0.46581 & & 0.19646 & 0.10178 & 0.19697 & 0.10206 & 0.15371 & \textcolor{green}{0.10080} & \textcolor{blue}{0.10170} \\ 
 & 90 & 1.26528 & & 0.39689 & 0.21062 & 0.33343 & \textcolor{green}{0.18941} & 0.29275 & \textcolor{blue}{0.19556} & 0.19723 \\ 
 & 120 & 2.85730 & & 1.02155 & 0.43815 & 0.77485 & 0.33635 & 0.64332 & \textcolor{green}{0.32268} & \textcolor{blue}{0.32373} \\ 
\hline
\end{tabular}
}
\end{subtable}
\vspace{1.5em}

\begin{subtable}{\textwidth}
\renewcommand{\thetable}{A.1.5}
\centering
\caption{TAFAS Consumer Sector}
\label{TAFAS consumption}
\renewcommand{\arraystretch}{1.2} 
\resizebox{\textwidth}{!}{
\begin{tabular}{cc|ccccccccc}
\hline
\multirow{3}{*}{Model} & \multirow{3}{*}{$H$} & \multirow{3}{*}{Base} & mse & $\checkmark$ & & & $\checkmark$ & $\checkmark$ & & $\checkmark$ \\
 & & & dir & & $\checkmark$ & & $\checkmark$ & & $\checkmark$ & $\checkmark$ \\
 & & & price & & & $\checkmark$ & & $\checkmark$ & $\checkmark$ & $\checkmark$ \\
\hline
\multirow{3}{*}{iTransformer} 
 & 60 & 0.02593 & & 0.02321 & 0.01470 & 0.02279 & 0.01293 & 0.01951 & \textcolor{green}{0.01285} & \textcolor{blue}{0.01287} \\ 
 & 90 & 0.05730 & & 0.05107 & 0.03065 & 0.04711 & \textcolor{blue}{0.02576} & 0.04303 & \textcolor{green}{0.02558} & 0.02580 \\ 
 & 120 & 0.11261 & & 0.10216 & 0.05875 & 0.10520 & \textcolor{green}{0.04693} & 0.09472 & \textcolor{blue}{0.04752} & 0.04851 \\ 
\hline
\multirow{3}{*}{DLinear} 
 & 60 & 0.01388 & & 0.01410 & 0.01197 & 0.01372 & 0.01145 & 0.01391 & \textcolor{blue}{0.01142} & \textcolor{green}{0.01139} \\ 
 & 90 & 0.02799 & & 0.02798 & 0.02277 & 0.02606 & 0.02125 & 0.02608 & \textcolor{green}{0.02086} & \textcolor{blue}{0.02086} \\ 
 & 120 & 0.04618 & & 0.04645 & 0.03707 & 0.04391 & 0.03451 & 0.04336 & \textcolor{blue}{0.03395} & \textcolor{green}{0.03391} \\ 
\hline
\multirow{3}{*}{FreTS} 
 & 60 & 0.00996 & & 0.01002 & \textcolor{green}{0.00966} & 0.00991 & \textcolor{blue}{0.00982} & 0.01009 & 0.00983 & 0.00982 \\ 
 & 90 & 0.01801 & & 0.01819 & \textcolor{green}{0.01721} & 0.01820 & 0.01735 & 0.01853 & \textcolor{blue}{0.01727} & 0.01727 \\ 
 & 120 & 0.02909 & & 0.02958 & 0.02760 & 0.02915 & 0.02756 & 0.02936 & \textcolor{green}{0.02745} & \textcolor{blue}{0.02751} \\ 
\hline
\multirow{3}{*}{PatchTST} 
 & 60 & 0.02102 & & 0.01554 & 0.01329 & 0.01831 & 0.01176 & 0.01409 & \textcolor{green}{0.01171} & \textcolor{blue}{0.01175} \\ 
 & 90 & 0.04354 & & 0.03153 & 0.02697 & 0.03689 & \textcolor{green}{0.02340} & 0.03051 & \textcolor{blue}{0.02380} & 0.02394 \\ 
 & 120 & 0.08100 & & 0.05787 & 0.04912 & 0.07020 & \textcolor{green}{0.04199} & 0.05679 & \textcolor{blue}{0.04261} & 0.04290 \\ 
\hline
\multirow{3}{*}{OLS} 
 & 60 & 0.02602 & & 0.02171 & 0.01512 & 0.02208 & 0.01325 & 0.01864 & \textcolor{green}{0.01322} & \textcolor{blue}{0.01325} \\ 
 & 90 & 0.05850 & & 0.04898 & 0.03197 & 0.04941 & 0.02890 & 0.04329 & \textcolor{blue}{0.02733} & \textcolor{green}{0.02732} \\ 
 & 120 & 0.11241 & & 0.09377 & 0.06099 & 0.14025 & 0.05283 & 0.10597 & \textcolor{blue}{0.05183} & \textcolor{green}{0.05142} \\ 
\hline
\end{tabular}
}
\end{subtable}
\end{table}

\begin{table}[htbp]
\ContinuedFloat
\centering
\begin{subtable}{\textwidth}
\renewcommand{\thetable}{A.1.6}
\centering
\caption{TAFAS Energy Sector}
\label{TAFAS energy}
\renewcommand{\arraystretch}{1.2} 
\resizebox{\textwidth}{!}{
\begin{tabular}{cc|ccccccccc}
\hline
\multirow{3}{*}{Model} & \multirow{3}{*}{$H$} & \multirow{3}{*}{Base} & mse & $\checkmark$ & & & $\checkmark$ & $\checkmark$ & & $\checkmark$ \\
 & & & dir & & $\checkmark$ & & $\checkmark$ & & $\checkmark$ & $\checkmark$ \\
 & & & price & & & $\checkmark$ & & $\checkmark$ & $\checkmark$ & $\checkmark$ \\
\hline
\multirow{3}{*}{iTransformer} 
 & 60 & 0.03685 & & 0.03582 & 0.02889 & 0.03232 & 0.02420 & 0.03052 & \textcolor{green}{0.02388} & \textcolor{blue}{0.02394} \\ 
 & 90 & 0.08183 & & 0.07929 & 0.05959 & 0.06565 & 0.04919 & 0.06332 & \textcolor{green}{0.04783} & \textcolor{blue}{0.04792} \\ 
 & 120 & 0.14979 & & 0.14584 & 0.10643 & 0.12253 & 0.08851 & 0.11775 & \textcolor{green}{0.08626} & \textcolor{blue}{0.08630} \\ 
\hline
\multirow{3}{*}{DLinear} 
 & 60 & 0.02703 & & 0.02706 & 0.02389 & 0.02606 & 0.02270 & 0.02614 & \textcolor{blue}{0.02259} & \textcolor{green}{0.02259} \\ 
 & 90 & 0.05452 & & 0.05439 & 0.04653 & 0.05057 & 0.04460 & 0.05084 & \textcolor{green}{0.04376} & \textcolor{blue}{0.04377} \\ 
 & 120 & 0.09314 & & 0.09334 & 0.07890 & 0.08959 & 0.07359 & 0.08954 & \textcolor{green}{0.07240} & \textcolor{blue}{0.07256} \\ 
\hline
\multirow{3}{*}{FreTS} 
 & 60 & 0.01859 & & 0.01862 & 0.01876 & \textcolor{green}{0.01842} & 0.01905 & \textcolor{blue}{0.01855} & 0.01907 & 0.01907 \\ 
 & 90 & \textcolor{blue}{0.03513} & & 0.03515 & 0.03553 & \textcolor{green}{0.03509} & 0.03608 & 0.03520 & 0.03617 & 0.03619 \\ 
 & 120 & \textcolor{blue}{0.05924} & & 0.05927 & 0.05937 & \textcolor{green}{0.05910} & 0.05963 & 0.05928 & 0.05958 & 0.05960 \\ 
\hline
\multirow{3}{*}{PatchTST} 
 & 60 & 0.03588 & & 0.02961 & 0.02611 & 0.03071 & 0.02307 & 0.02603 & \textcolor{green}{0.02298} & \textcolor{blue}{0.02299} \\ 
 & 90 & 0.07545 & & 0.05974 & 0.05193 & 0.06025 & 0.04637 & 0.05237 & \textcolor{blue}{0.04585} & \textcolor{green}{0.04581} \\ 
 & 120 & 0.14276 & & 0.10966 & 0.08976 & 0.11428 & 0.08681 & 0.09559 & \textcolor{blue}{0.08566} & \textcolor{green}{0.08548} \\ 
\hline
\multirow{3}{*}{OLS} 
 & 60 & 0.03871 & & 0.03468 & 0.02814 & 0.03288 & 0.02420 & 0.03084 & \textcolor{green}{0.02402} & \textcolor{blue}{0.02410} \\ 
 & 90 & 0.08505 & & 0.07633 & 0.05805 & 0.06563 & 0.04952 & 0.06329 & \textcolor{green}{0.04897} & \textcolor{blue}{0.04900} \\ 
 & 120 & 0.15842 & & 0.14309 & 0.10604 & 0.12391 & 0.09186 & 0.12003 & \textcolor{green}{0.09043} & \textcolor{blue}{0.09056} \\ 
\hline
\end{tabular}
}
\end{subtable}
\vspace{1.5em}

\begin{subtable}{\textwidth}
\renewcommand{\thetable}{A.1.7}
\centering
\caption{TAFAS Nasdaq Mixed Excerpt}
\label{TAFAS mix}
\renewcommand{\arraystretch}{1.2} 
\resizebox{\textwidth}{!}{
\begin{tabular}{cc|ccccccccc}
\hline
\multirow{3}{*}{Model} & \multirow{3}{*}{$H$} & \multirow{3}{*}{Base} & mse & $\checkmark$ & & & $\checkmark$ & $\checkmark$ & & $\checkmark$ \\
 & & & dir & & $\checkmark$ & & $\checkmark$ & & $\checkmark$ & $\checkmark$ \\
 & & & price & & & $\checkmark$ & & $\checkmark$ & $\checkmark$ & $\checkmark$ \\
\hline
\multirow{3}{*}{iTransformer} 
 & 60 & 0.05752 & & 0.05120 & 0.03579 & 0.04748 & 0.03150 & 0.04126 & \textcolor{green}{0.03117} & \textcolor{blue}{0.03120} \\ 
 & 90 & 0.12469 & & 0.11041 & 0.07129 & 0.09201 & 0.06084 & 0.08366 & \textcolor{green}{0.05923} & \textcolor{blue}{0.05948} \\ 
 & 120 & 0.22275 & & 0.19713 & 0.12478 & 0.17644 & 0.10536 & 0.15893 & \textcolor{green}{0.10383} & \textcolor{blue}{0.10428} \\ 
\hline
\multirow{3}{*}{DLinear} 
 & 60 & 0.03478 & & 0.03487 & 0.03001 & 0.03394 & 0.02854 & 0.03369 & \textcolor{blue}{0.02839} & \textcolor{green}{0.02835} \\ 
 & 90 & 0.07097 & & 0.06953 & 0.05640 & 0.06453 & 0.05199 & 0.06423 & \textcolor{blue}{0.05073} & \textcolor{green}{0.05062} \\ 
 & 120 & 0.11325 & & 0.11182 & 0.08922 & 0.10578 & 0.08064 & 0.10431 & \textcolor{blue}{0.07868} & \textcolor{green}{0.07857} \\ 
\hline
\multirow{3}{*}{FreTS} 
 & 60 & 0.02635 & & 0.02644 & \textcolor{green}{0.02542} & 0.02614 & 0.02572 & 0.02623 & 0.02572 & \textcolor{blue}{0.02569} \\ 
 & 90 & 0.04852 & & 0.04852 & 0.04512 & 0.04807 & 0.04473 & 0.04835 & \textcolor{green}{0.04445} & \textcolor{blue}{0.04447} \\ 
 & 120 & 0.07669 & & 0.07689 & 0.07040 & 0.07618 & 0.06878 & 0.07634 & \textcolor{green}{0.06835} & \textcolor{blue}{0.06843} \\ 
\hline
\multirow{3}{*}{PatchTST} 
 & 60 & 0.05240 & & 0.03595 & 0.03423 & 0.04295 & 0.02906 & 0.03036 & \textcolor{blue}{0.02885} & \textcolor{green}{0.02879} \\ 
 & 90 & 0.10923 & & 0.07137 & 0.06663 & 0.07794 & 0.05439 & 0.05619 & \textcolor{green}{0.05309} & \textcolor{blue}{0.05309} \\ 
 & 120 & 0.19164 & & 0.12389 & 0.11390 & 0.14208 & 0.09288 & 0.09489 & \textcolor{blue}{0.08997} & \textcolor{green}{0.08985} \\ 
\hline
\multirow{3}{*}{OLS} 
 & 60 & 0.05930 & & 0.04541 & 0.03629 & 0.04705 & 0.03213 & 0.03849 & \textcolor{green}{0.03196} & \textcolor{blue}{0.03198} \\ 
 & 90 & 0.12777 & & 0.09432 & 0.07364 & 0.08609 & 0.06378 & 0.07969 & \textcolor{green}{0.06323} & \textcolor{blue}{0.06332} \\ 
 & 120 & 0.22788 & & 0.16949 & 0.12910 & 0.15975 & 0.10747 & 0.14324 & \textcolor{green}{0.10574} & \textcolor{blue}{0.10652} \\ 
\hline
\end{tabular}
}
\end{subtable}
\end{table}

\begin{table}[htbp]
\ContinuedFloat
\centering
\begin{subtable}{\textwidth}
\renewcommand{\thetable}{A.1.8}
\centering
\caption{TAFAS Macro and Market Series}
\label{TAFAS index}
\renewcommand{\arraystretch}{1.2} 
\resizebox{\textwidth}{!}{
\begin{tabular}{cc|ccccccccc}
\hline
\multirow{3}{*}{Model} & \multirow{3}{*}{$H$} & \multirow{3}{*}{Base} & mse & $\checkmark$ & & & $\checkmark$ & $\checkmark$ & & $\checkmark$ \\
 & & & dir & & $\checkmark$ & & $\checkmark$ & & $\checkmark$ & $\checkmark$ \\
 & & & price & & & $\checkmark$ & & $\checkmark$ & $\checkmark$ & $\checkmark$ \\
\hline
\multirow{3}{*}{iTransformer} 
 & 60 & 0.01886 & & 0.01606 & 0.01207 & 0.01567 & 0.01099 & 0.01315 & \textcolor{blue}{0.01097} & \textcolor{green}{0.01092} \\ 
 & 90 & 0.04027 & & 0.03476 & 0.02344 & 0.02956 & 0.02199 & 0.02661 & \textcolor{blue}{0.02124} & \textcolor{green}{0.02116} \\ 
 & 120 & 0.07258 & & 0.06056 & 0.03946 & 0.05605 & 0.03702 & 0.05049 & \textcolor{blue}{0.03600} & \textcolor{green}{0.03579} \\ 
\hline
\multirow{3}{*}{DLinear} 
 & 60 & 0.01248 & & 0.01228 & 0.01101 & 0.01226 & 0.01062 & 0.01190 & \textcolor{blue}{0.01059} & \textcolor{green}{0.01057} \\ 
 & 90 & 0.02567 & & 0.02484 & 0.02159 & 0.02317 & 0.02052 & 0.02230 & \textcolor{blue}{0.02014} & \textcolor{green}{0.02012} \\ 
 & 120 & 0.04046 & & 0.03971 & 0.03356 & 0.03792 & 0.03147 & 0.03694 & \textcolor{blue}{0.03116} & \textcolor{green}{0.03101} \\ 
\hline
\multirow{3}{*}{FreTS} 
 & 60 & \textcolor{blue}{0.00882} & & 0.00883 & 0.00910 & \textcolor{green}{0.00880} & 0.00938 & 0.00897 & 0.00940 & 0.00936 \\ 
 & 90 & 0.01670 & & 0.01673 & 0.01704 & \textcolor{green}{0.01665} & 0.01743 & \textcolor{blue}{0.01667} & 0.01726 & 0.01722 \\ 
 & 120 & \textcolor{green}{0.02586} & & 0.02594 & 0.02614 & 0.02593 & 0.02659 & \textcolor{blue}{0.02589} & 0.02645 & 0.02642 \\ 
\hline
\multirow{3}{*}{PatchTST} 
 & 60 & 0.01793 & & 0.01208 & 0.01152 & 0.01388 & 0.01052 & \textcolor{green}{0.01029} & 0.01045 & \textcolor{blue}{0.01038} \\ 
 & 90 & 0.03668 & & 0.02444 & 0.02201 & 0.02495 & 0.01957 & 0.02000 & \textcolor{blue}{0.01879} & \textcolor{green}{0.01869} \\ 
 & 120 & 0.06088 & & 0.04017 & 0.03578 & 0.04431 & 0.03253 & 0.03414 & \textcolor{blue}{0.03106} & \textcolor{green}{0.03086} \\ 
\hline
\multirow{3}{*}{OLS} 
 & 60 & 0.02018 & & 0.01473 & 0.01207 & 0.01430 & 0.01097 & 0.01205 & \textcolor{blue}{0.01097} & \textcolor{green}{0.01087} \\ 
 & 90 & 0.04483 & & 0.03259 & 0.02399 & 0.02716 & 0.02192 & 0.02361 & \textcolor{blue}{0.02155} & \textcolor{green}{0.02150} \\ 
 & 120 & 0.07908 & & 0.05686 & 0.03924 & 0.04825 & 0.03570 & 0.04205 & \textcolor{green}{0.03484} & \textcolor{blue}{0.03522} \\ 
\hline
\end{tabular}
}
\end{subtable}
\vspace{1.5em}

\begin{subtable}{\textwidth}
\renewcommand{\thetable}{A.1.9} 
\centering
\caption{PETSA Magnificent 7}
\label{PETSA magnificent 7}
\renewcommand{\arraystretch}{1.2}
\resizebox{\textwidth}{!}{
\begin{tabular}{cc|cccccccccc}
\hline
\multirow{3}{*}{Model} & \multirow{3}{*}{$H$} & \multirow{3}{*}{Base} & \multirow{3}{*}{original} & mse & \checkmark & & & \checkmark & \checkmark & & \checkmark \\
 & & & & dir & & \checkmark & & \checkmark & & \checkmark & \checkmark \\
 & & & & price & & & \checkmark & & \checkmark & \checkmark & \checkmark \\
\hline
\multirow{3}{*}{iTransformer} 
 & 60 & 1.61743 & 1.07880 & & 1.00403 & 0.68592 & 1.06174 & \textcolor{blue}{0.68473} & 0.81383 & \textcolor{green}{0.66979} & 0.78249 \\
 & 90 & 6.16393 & 3.45883 & & 2.94524 & 1.66862 & 3.15973 & \textcolor{blue}{1.66476} & 2.28144 & \textcolor{green}{1.56710} & 1.84711 \\
 & 120 & 17.71737 & 7.58773 & & 7.77118 & \textcolor{blue}{3.60424} & 10.26046 & 3.63848 & 6.76601 & \textcolor{green}{3.42331} & 3.89974 \\
\hline
\multirow{3}{*}{DLinear} 
 & 60 & 0.75416 & 0.76475 & & 0.76093 & \textcolor{green}{0.49869} & 0.66956 & 0.49986 & 0.68791 & \textcolor{blue}{0.49920} & 0.50238 \\
 & 90 & 2.06209 & 1.82557 & & 1.71048 & 1.03138 & 1.50014 & 1.03347 & 1.46148 & \textcolor{green}{0.98851} & \textcolor{blue}{1.01324} \\
 & 120 & 3.87860 & 3.60260 & & 3.49313 & 1.70534 & 3.25740 & 1.70859 & 3.06473 & \textcolor{blue}{1.69181} & \textcolor{green}{1.55223} \\
\hline
\multirow{3}{*}{FreTS} 
 & 60 & 0.52950 & 0.57772 & & 0.55204 & \textcolor{blue}{0.46512} & 0.50428 & 0.46569 & 0.52456 & \textcolor{green}{0.46467} & 0.46929 \\
 & 90 & 1.15247 & 1.19871 & & 1.18350 & 0.92256 & 1.06746 & 0.92406 & 1.06852 & \textcolor{blue}{0.92024} & \textcolor{green}{0.88769} \\
 & 120 & 2.03039 & 2.14219 & & 2.01618 & \textcolor{blue}{1.53752} & 1.96713 & 1.54297 & 1.91395 & 1.56167 & \textcolor{green}{1.46289} \\
\hline
\multirow{3}{*}{PatchTST} 
 & 60 & 1.50215 & 0.70887 & & 0.65572 & 0.51385 & 0.98248 & 0.51307 & 0.62744 & \textcolor{blue}{0.50344} & \textcolor{green}{0.48915} \\
 & 90 & 5.26546 & 1.96757 & & 2.02453 & 1.37594 & 2.43478 & 1.37078 & 1.65070 & \textcolor{blue}{1.27632} & \textcolor{green}{1.10568} \\
 & 120 & 13.14714 & 5.28275 & & 5.64375 & 2.77521 & 7.35952 & 2.79816 & 4.40622 & \textcolor{blue}{2.71818} & \textcolor{green}{2.65405} \\
\hline
\multirow{3}{*}{OLS} 
 & 60 & 1.78386 & 1.10668 & & 0.82033 & 0.59045 & 1.09650 & \textcolor{blue}{0.58908} & 0.81343 & \textcolor{green}{0.57098} & 0.64600 \\
 & 90 & 5.76070 & 2.45446 & & 1.99504 & 1.45515 & 2.61408 & 1.45487 & 2.14437 & \textcolor{blue}{1.33146} & \textcolor{green}{1.29933} \\
 & 120 & 15.58452 & 5.35221 & & 5.18654 & \textcolor{blue}{3.06482} & 7.08726 & 3.07645 & 5.58615 & 3.08803 & \textcolor{green}{2.78337} \\
\hline
\end{tabular}
}
\end{subtable}
\end{table}

\begin{table}[htbp]
\ContinuedFloat
\centering
\begin{subtable}{\textwidth}
\renewcommand{\thetable}{A.1.10} 
\centering
\caption{PETSA S\&P 500 Excerpt}
\label{PETSA sp500}
\renewcommand{\arraystretch}{1.2}
\resizebox{\textwidth}{!}{
\begin{tabular}{cc|cccccccccc}
\hline
\multirow{3}{*}{Model} & \multirow{3}{*}{$H$} & \multirow{3}{*}{Base} & \multirow{3}{*}{original} & mse & \checkmark & & & \checkmark & \checkmark & & \checkmark \\
 & & & & dir & & \checkmark & & \checkmark & & \checkmark & \checkmark \\
 & & & & price & & & \checkmark & & \checkmark & \checkmark & \checkmark \\
\hline
\multirow{3}{*}{iTransformer} 
 & 60 & 0.31049 & 0.17225 & & 0.16791 & \textcolor{green}{0.11173} & 0.19314 & \textcolor{blue}{0.11210} & 0.15084 & 0.11315 & 0.12200 \\
 & 90 & 1.08383 & 0.43358 & & 0.50458 & 0.29117 & 0.47271 & \textcolor{blue}{0.29077} & 0.37550 & \textcolor{green}{0.28899} & 0.31976 \\
 & 120 & 3.66154 & 1.32155 & & 1.56566 & 0.89284 & 1.74055 & \textcolor{blue}{0.89279} & 1.24945 & \textcolor{green}{0.88830} & 0.96333 \\
\hline
\multirow{3}{*}{DLinear} 
 & 60 & 0.12715 & 0.14383 & & 0.13382 & 0.09187 & 0.12159 & 0.09231 & 0.12473 & \textcolor{blue}{0.09114} & \textcolor{green}{0.08410} \\
 & 90 & 0.30746 & 0.32153 & & 0.29276 & 0.19141 & 0.24487 & 0.19166 & 0.23608 & \textcolor{blue}{0.18748} & \textcolor{green}{0.16397} \\
 & 120 & 0.61329 & 0.68129 & & 0.60700 & 0.32525 & 0.47748 & 0.32623 & 0.44502 & \textcolor{blue}{0.30402} & \textcolor{green}{0.28872} \\
\hline
\multirow{3}{*}{FreTS} 
 & 60 & 0.07956 & 0.08676 & & 0.08135 & \textcolor{green}{0.07416} & 0.07883 & \textcolor{blue}{0.07428} & 0.08018 & 0.07434 & 0.07487 \\
 & 90 & 0.15343 & 0.16424 & & 0.15190 & \textcolor{blue}{0.13559} & 0.15249 & 0.13590 & 0.15004 & 0.13669 & \textcolor{green}{0.13526} \\
 & 120 & 0.27454 & 0.30168 & & 0.27128 & \textcolor{blue}{0.23052} & 0.27221 & 0.23106 & 0.26629 & 0.23154 & \textcolor{green}{0.22538} \\
\hline
\multirow{3}{*}{PatchTST} 
 & 60 & 0.28742 & 0.10828 & & 0.10547 & \textcolor{blue}{0.09665} & 0.17528 & 0.09671 & 0.10784 & \textcolor{green}{0.09616} & 0.10309 \\
 & 90 & 0.79353 & 0.21113 & & 0.21860 & \textcolor{blue}{0.20131} & 0.40784 & 0.20229 & 0.24826 & 0.20430 & \textcolor{green}{0.19987} \\
 & 120 & 2.41471 & 0.55255 & & 0.56534 & 0.45407 & 1.23980 & \textcolor{blue}{0.45376} & 0.61070 & \textcolor{green}{0.45336} & 0.46982 \\
\hline
\multirow{3}{*}{OLS} 
 & 60 & 0.36039 & 0.17976 & & 0.16889 & \textcolor{green}{0.11857} & 0.20191 & \textcolor{blue}{0.11861} & 0.15114 & 0.11886 & 0.12514 \\
 & 90 & 1.18749 & 0.42979 & & 0.42494 & 0.31405 & 0.43555 & \textcolor{blue}{0.31237} & 0.37003 & \textcolor{green}{0.30813} & 0.33303 \\
 & 120 & 4.10791 & 1.26770 & & 1.22648 & 0.95783 & 1.46176 & 0.95624 & 1.18374 & \textcolor{green}{0.93823} & \textcolor{blue}{0.94652} \\
\hline
\end{tabular}
}
\end{subtable}
\vspace{1.5em}

\begin{subtable}{\textwidth}
\renewcommand{\thetable}{A.1.11} 
\centering
\caption{PETSA Russell 2000 Excerpt}
\label{PETSA russell2000}
\renewcommand{\arraystretch}{1.2}
\resizebox{\textwidth}{!}{
\begin{tabular}{cc|cccccccccc}
\hline
\multirow{3}{*}{Model} & \multirow{3}{*}{$H$} & \multirow{3}{*}{Base} & \multirow{3}{*}{original} & mse & \checkmark & & & \checkmark & \checkmark & & \checkmark \\
 & & & & dir & & \checkmark & & \checkmark & & \checkmark & \checkmark \\
 & & & & price & & & \checkmark & & \checkmark & \checkmark & \checkmark \\
\hline
\multirow{3}{*}{iTransformer} 
 & 60 & 0.30491 & 0.19194 & & 0.22250 & \textcolor{blue}{0.11596} & 0.26009 & 0.11825 & 0.17019 & \textcolor{green}{0.11497} & 0.11831 \\
 & 90 & 0.69923 & 0.61013 & & 0.25800 & 0.21674 & 0.45696 & 0.21689 & 0.29670 & \textcolor{blue}{0.20746} & \textcolor{green}{0.15115} \\
 & 120 & 1.45384 & 0.76688 & & 0.62010 & 0.49672 & 1.04817 & 0.49697 & 0.67196 & \textcolor{blue}{0.47574} & \textcolor{green}{0.36467} \\
\hline
\multirow{3}{*}{DLinear} 
 & 60 & 0.12283 & 0.13785 & & 0.15255 & \textcolor{blue}{0.07802} & 0.11757 & 0.07831 & 0.11417 & 0.07846 & \textcolor{green}{0.07524} \\
 & 90 & 0.29572 & 0.24089 & & 0.30426 & 0.16120 & 0.22626 & 0.16167 & 0.22808 & \textcolor{blue}{0.16031} & \textcolor{green}{0.13707} \\
 & 120 & 0.50299 & 0.42464 & & 0.52829 & 0.28865 & 0.44771 & 0.28960 & 0.44231 & \textcolor{blue}{0.28291} & \textcolor{green}{0.25080} \\
\hline
\multirow{3}{*}{FreTS} 
 & 60 & 0.06752 & 0.08139 & & 0.07521 & \textcolor{green}{0.06437} & 0.06948 & \textcolor{blue}{0.06453} & 0.07434 & 0.06459 & 0.06536 \\
 & 90 & 0.14269 & 0.16834 & & 0.15153 & \textcolor{blue}{0.13127} & 0.14327 & 0.13163 & 0.14855 & 0.13152 & \textcolor{green}{0.12702} \\
 & 120 & 0.24603 & 0.28902 & & 0.25990 & \textcolor{blue}{0.22726} & 0.25255 & 0.22789 & 0.26295 & 0.22891 & \textcolor{green}{0.22538} \\
\hline
\multirow{3}{*}{PatchTST} 
 & 60 & 0.23714 & 0.10576 & & 0.10338 & 0.08588 & 0.16036 & 0.08608 & 0.09266 & \textcolor{blue}{0.08544} & \textcolor{green}{0.08393} \\
 & 90 & 0.62163 & 0.21849 & & 0.22426 & 0.19151 & 0.36987 & 0.19194 & 0.22143 & \textcolor{blue}{0.18946} & \textcolor{green}{0.14741} \\
 & 120 & 1.19838 & 0.44963 & & 0.48724 & 0.41816 & 0.79983 & 0.41925 & 0.46406 & \textcolor{blue}{0.41218} & \textcolor{green}{0.29811} \\
\hline
\multirow{3}{*}{OLS} 
 & 60 & 0.30977 & 0.14224 & & 0.13248 & 0.08888 & 0.13314 & 0.08926 & 0.11117 & \textcolor{blue}{0.08863} & \textcolor{green}{0.08327} \\
 & 90 & 0.79484 & 0.28714 & & 0.23213 & 0.19452 & 0.28532 & 0.19585 & 0.23270 & \textcolor{blue}{0.19405} & \textcolor{green}{0.16022} \\
 & 120 & 1.58614 & 0.63182 & & 0.59956 & 0.44173 & 0.73436 & 0.44528 & 0.56560 & \textcolor{blue}{0.43284} & \textcolor{green}{0.33070} \\
\hline
\end{tabular}
}
\end{subtable}
\end{table}

\begin{table}[htbp]
\ContinuedFloat
\centering
\begin{subtable}{\textwidth}
\renewcommand{\thetable}{A.1.12} 
\centering
\caption{PETSA Technology Sector}
\label{PETSA Technology Sector}
\renewcommand{\arraystretch}{1.2}
\resizebox{\textwidth}{!}{
\begin{tabular}{cc|cccccccccc}
\hline
\multirow{3}{*}{Model} & \multirow{3}{*}{$H$} & \multirow{3}{*}{Base} & \multirow{3}{*}{original} & mse & \checkmark & & & \checkmark & \checkmark & & \checkmark \\
 & & & & dir & & \checkmark & & \checkmark & & \checkmark & \checkmark \\
 & & & & price & & & \checkmark & & \checkmark & \checkmark & \checkmark \\
\hline
\multirow{3}{*}{iTransformer} 
 & 60 & 0.42161 & 0.19636 & & 0.15864 & \textcolor{blue}{0.09849} & 0.23867 & 0.09885 & 0.17437 & \textcolor{green}{0.09787} & 0.09929 \\
 & 90 & 1.26285 & 0.45928 & & 0.37049 & 0.19947 & 0.49278 & 0.19969 & 0.32317 & \textcolor{blue}{0.19629} & \textcolor{green}{0.19278} \\
 & 120 & 2.85160 & 0.93674 & & 0.85556 & 0.40467 & 1.17880 & 0.40739 & 0.70321 & \textcolor{blue}{0.39307} & \textcolor{green}{0.34146} \\
\hline
\multirow{3}{*}{DLinear} 
 & 60 & 0.16194 & 0.21405 & & 0.18530 & \textcolor{green}{0.09944} & 0.14540 & 0.10025 & 0.14566 & \textcolor{blue}{0.09986} & 0.10094 \\
 & 90 & 0.35196 & 0.28986 & & 0.36762 & 0.18323 & 0.23670 & 0.18316 & 0.23959 & \textcolor{green}{0.17729} & \textcolor{blue}{0.17750} \\
 & 120 & 0.59787 & 0.65952 & & 0.68040 & \textcolor{blue}{0.30069} & 0.42510 & 0.30095 & 0.42508 & 0.30098 & \textcolor{green}{0.29276} \\
\hline
\multirow{3}{*}{FreTS} 
 & 60 & 0.09694 & 0.11603 & & 0.10637 & \textcolor{green}{0.08950} & 0.09647 & \textcolor{blue}{0.08978} & 0.10346 & 0.08984 & 0.09096 \\
 & 90 & 0.18422 & 0.22629 & & 0.19976 & \textcolor{green}{0.15830} & 0.18312 & \textcolor{blue}{0.15869} & 0.19157 & 0.15958 & 0.15964 \\
 & 120 & 0.30389 & 0.39962 & & 0.33023 & \textcolor{blue}{0.25075} & 0.30568 & 0.25152 & 0.32093 & 0.25322 & \textcolor{green}{0.24903} \\
\hline
\multirow{3}{*}{PatchTST} 
 & 60 & 0.30671 & 0.14063 & & 0.12215 & 0.09916 & 0.17770 & \textcolor{blue}{0.09908} & 0.11185 & \textcolor{green}{0.09848} & 0.10292 \\
 & 90 & 0.83752 & 0.28471 & & 0.26085 & \textcolor{green}{0.18215} & 0.39180 & \textcolor{blue}{0.18219} & 0.23400 & 0.18635 & 0.18498 \\
 & 120 & 1.77862 & 0.43336 & & 0.48765 & \textcolor{blue}{0.35324} & 0.82280 & 0.35699 & 0.38591 & 0.36388 & \textcolor{green}{0.32748} \\
\hline
\multirow{3}{*}{OLS} 
 & 60 & 0.46581 & 0.20059 & & 0.15660 & 0.10417 & 0.19130 & 0.10411 & 0.14755 & \textcolor{blue}{0.10176} & \textcolor{green}{0.09960} \\
 & 90 & 1.26527 & 0.45351 & & 0.31629 & 0.20398 & 0.30476 & 0.20029 & 0.26687 & \textcolor{blue}{0.18750} & \textcolor{green}{0.17870} \\
 & 120 & 2.85730 & 0.94798 & & 0.84752 & 0.38710 & 0.65648 & 0.38714 & 0.63713 & \textcolor{blue}{0.36614} & \textcolor{green}{0.31308} \\
\hline
\end{tabular}
}
\end{subtable}
\vspace{1.5em}

\begin{subtable}{\textwidth}
\renewcommand{\thetable}{A.1.13} 
\centering
\caption{PETSA Consumer Sector}
\label{PETSA consumption}
\renewcommand{\arraystretch}{1.2}
\resizebox{\textwidth}{!}{
\begin{tabular}{cc|cccccccccc}
\hline
\multirow{3}{*}{Model} & \multirow{3}{*}{$H$} & \multirow{3}{*}{Base} & \multirow{3}{*}{original} & mse & \checkmark & & & \checkmark & \checkmark & & \checkmark \\
 & & & & dir & & \checkmark & & \checkmark & & \checkmark & \checkmark \\
 & & & & price & & & \checkmark & & \checkmark & \checkmark & \checkmark \\
\hline
\multirow{3}{*}{iTransformer} 
 & 60 & 0.02593 & 0.02026 & & 0.02009 & 0.01464 & 0.02327 & 0.01471 & 0.01956 & \textcolor{blue}{0.01459} & \textcolor{green}{0.01308} \\
 & 90 & 0.05730 & 0.04530 & & 0.04407 & \textcolor{blue}{0.03117} & 0.04808 & 0.03144 & 0.04305 & 0.03163 & \textcolor{green}{0.02636} \\
 & 120 & 0.11261 & 0.08673 & & 0.08694 & \textcolor{blue}{0.05895} & 0.09730 & 0.05914 & 0.08430 & 0.05920 & \textcolor{green}{0.04684} \\
\hline
\multirow{3}{*}{DLinear} 
 & 60 & 0.01388 & 0.01573 & & 0.01441 & 0.01182 & 0.01373 & 0.01179 & 0.01422 & \textcolor{blue}{0.01175} & \textcolor{green}{0.01121} \\
 & 90 & 0.02799 & 0.03065 & & 0.02780 & 0.02172 & 0.02609 & 0.02167 & 0.02662 & \textcolor{blue}{0.02132} & \textcolor{green}{0.02060} \\
 & 120 & 0.04618 & 0.04893 & & 0.04699 & 0.03510 & 0.04453 & 0.03506 & 0.04531 & \textcolor{blue}{0.03463} & \textcolor{green}{0.03308} \\
\hline
\multirow{3}{*}{FreTS} 
 & 60 & 0.00996 & 0.01165 & & 0.01022 & \textcolor{green}{0.00957} & 0.00985 & \textcolor{blue}{0.00958} & 0.01025 & 0.00958 & 0.00970 \\
 & 90 & 0.01801 & 0.02189 & & 0.01842 & \textcolor{green}{0.01675} & 0.01795 & 0.01676 & 0.01876 & \textcolor{blue}{0.01675} & 0.01684 \\
 & 120 & 0.02909 & 0.03534 & & 0.03009 & \textcolor{blue}{0.02645} & 0.02917 & 0.02652 & 0.03063 & 0.02659 & \textcolor{green}{0.02641} \\
\hline
\multirow{3}{*}{PatchTST} 
 & 60 & 0.02102 & 0.01636 & & 0.01402 & 0.01274 & 0.01792 & 0.01272 & 0.01395 & \textcolor{blue}{0.01272} & \textcolor{green}{0.01158} \\
 & 90 & 0.04354 & 0.03144 & & 0.02790 & 0.02518 & 0.03430 & 0.02519 & 0.02846 & \textcolor{blue}{0.02508} & \textcolor{green}{0.02236} \\
 & 120 & 0.08100 & 0.05415 & & 0.05189 & 0.04500 & 0.06623 & 0.04500 & 0.05220 & \textcolor{blue}{0.04456} & \textcolor{green}{0.03797} \\
\hline
\multirow{3}{*}{OLS} 
 & 60 & 0.02602 & 0.02091 & & 0.01894 & 0.01392 & 0.02072 & 0.01388 & 0.01908 & \textcolor{blue}{0.01378} & \textcolor{green}{0.01276} \\
 & 90 & 0.05850 & 0.04311 & & 0.03763 & 0.02990 & 0.03942 & 0.03002 & 0.03683 & \textcolor{blue}{0.02971} & \textcolor{green}{0.02599} \\
 & 120 & 0.11241 & 0.07839 & & 0.07447 & 0.05680 & 0.08246 & 0.05731 & 0.07439 & \textcolor{blue}{0.05656} & \textcolor{green}{0.04572} \\
\hline
\end{tabular}
}
\end{subtable}
\end{table}

\begin{table}[htbp]
\ContinuedFloat
\centering
\begin{subtable}{\textwidth}
\renewcommand{\thetable}{A.1.14} 
\centering
\caption{PETSA Energy Sector}
\label{PETSA energy}
\renewcommand{\arraystretch}{1.2}
\resizebox{\textwidth}{!}{
\begin{tabular}{cc|cccccccccc}
\hline
\multirow{3}{*}{Model} & \multirow{3}{*}{$H$} & \multirow{3}{*}{Base} & \multirow{3}{*}{original} & mse & \checkmark & & & \checkmark & \checkmark & & \checkmark \\
 & & & & dir & & \checkmark & & \checkmark & & \checkmark & \checkmark \\
 & & & & price & & & \checkmark & & \checkmark & \checkmark & \checkmark \\
\hline
\multirow{3}{*}{iTransformer} 
 & 60 & 0.03685 & 0.03276 & & 0.03380 & 0.02552 & 0.03316 & 0.02573 & 0.03207 & \textcolor{blue}{0.02506} & \textcolor{green}{0.02497} \\
 & 90 & 0.08183 & 0.06904 & & 0.07473 & \textcolor{blue}{0.05257} & 0.06839 & 0.05297 & 0.06621 & \textcolor{green}{0.05097} & 0.05294 \\
 & 120 & 0.14979 & 0.12655 & & 0.14045 & 0.09470 & 0.13382 & 0.09494 & 0.13085 & \textcolor{green}{0.09220} & \textcolor{blue}{0.09400} \\
\hline
\multirow{3}{*}{DLinear} 
 & 60 & 0.02703 & 0.02834 & & 0.02827 & 0.02421 & 0.02626 & 0.02419 & 0.02687 & \textcolor{blue}{0.02403} & \textcolor{green}{0.02324} \\
 & 90 & 0.05452 & 0.05819 & & 0.05602 & 0.04829 & 0.05110 & 0.04821 & 0.05202 & \textcolor{blue}{0.04665} & \textcolor{green}{0.04255} \\
 & 120 & 0.09314 & 0.10043 & & 0.09706 & 0.08275 & 0.08971 & 0.08259 & 0.09164 & \textcolor{blue}{0.08163} & \textcolor{green}{0.07527} \\
\hline
\multirow{3}{*}{FreTS} 
 & 60 & \textcolor{blue}{0.01859}& 0.01973 & & 0.01900 & 0.01865 & \textcolor{green}{0.01823} & 0.01866 & 0.01878 & 0.01860 & 0.01886 \\
 & 90 & \textcolor{blue}{0.03513}& 0.03833 & & 0.03655 & 0.03552 & \textcolor{green}{0.03482} & 0.03556 & 0.03605 & 0.03545 & 0.03595 \\
 & 120 & 0.05924 & 0.06452 & & 0.06152 & 0.05936 & \textcolor{green}{0.05863} & 0.05939 & 0.06076 & \textcolor{blue}{0.05919} & 0.05933 \\
\hline
\multirow{3}{*}{PatchTST} 
 & 60 & 0.03588 & 0.02903 & & 0.02672 & \textcolor{blue}{0.02491} & 0.03211 & 0.02492 & 0.03132 & \textcolor{green}{0.02461} & 0.02684 \\
 & 90 & 0.07545 & 0.05649 & & 0.05616 & \textcolor{blue}{0.04993} & 0.06565 & 0.04995 & 0.06229 & \textcolor{green}{0.04937} & 0.05191 \\
 & 120 & 0.14276 & 0.09991 & & 0.10447 & \textcolor{blue}{0.08882} & 0.12496 & 0.08890 & 0.11815 & \textcolor{green}{0.08832} & 0.08923 \\
\hline
\multirow{3}{*}{OLS} 
 & 60 & 0.03871 & 0.03295 & & 0.03161 & \textcolor{blue}{0.02646} & 0.03129 & 0.02658 & 0.03033 & \textcolor{green}{0.02612} & 0.02704 \\
 & 90 & 0.08505 & 0.07124 & & 0.06834 & \textcolor{blue}{0.05419} & 0.06361 & 0.05445 & 0.06359 & \textcolor{green}{0.05397} & 0.05501 \\
 & 120 & 0.15842 & 0.13003 & & 0.13230 & \textcolor{blue}{0.09639} & 0.12342 & 0.09656 & 0.12116 & 0.09744 & \textcolor{green}{0.09603} \\
\hline
\end{tabular}
}
\end{subtable}
\vspace{1.5em}

\begin{subtable}{\textwidth}
\renewcommand{\thetable}{A.1.15} 
\centering
\caption{PETSA Nasdaq Mixed Excerpt}
\label{PETSA mix}
\renewcommand{\arraystretch}{1.2}
\resizebox{\textwidth}{!}{
\begin{tabular}{cc|cccccccccc}
\hline
\multirow{3}{*}{Model} & \multirow{3}{*}{$H$} & \multirow{3}{*}{Base} & \multirow{3}{*}{original} & mse & \checkmark & & & \checkmark & \checkmark & & \checkmark \\
 & & & & dir & & \checkmark & & \checkmark & & \checkmark & \checkmark \\
 & & & & price & & & \checkmark & & \checkmark & \checkmark & \checkmark \\
\hline
\multirow{3}{*}{iTransformer} 
 & 60 & 0.05752 & 0.04065 & & 0.04152 & 0.03367 & 0.04882 & 0.03360 & 0.04129 & \textcolor{blue}{0.03329} & \textcolor{green}{0.03114} \\
 & 90 & 0.12469 & 0.09315 & & 0.09210 & 0.06737 & 0.09673 & 0.06748 & 0.08499 & \textcolor{blue}{0.06637} & \textcolor{green}{0.05752} \\
 & 120 & 0.22275 & 0.16534 & & 0.16415 & 0.12091 & 0.18026 & 0.12111 & 0.15351 & \textcolor{blue}{0.11866} & \textcolor{green}{0.10034} \\
\hline
\multirow{3}{*}{DLinear} 
 & 60 & 0.03478 & 0.03641 & & 0.03594 & 0.02890 & 0.03388 & 0.02885 & 0.03488 & \textcolor{blue}{0.02873} & \textcolor{green}{0.02851} \\
 & 90 & 0.07097 & 0.06912 & & 0.06945 & 0.05287 & 0.06320 & 0.05272 & 0.06362 & \textcolor{blue}{0.05144} & \textcolor{green}{0.04959} \\
 & 120 & 0.11325 & 0.10974 & & 0.11371 & 0.08360 & 0.10743 & 0.08359 & 0.10638 & \textcolor{blue}{0.08270} & \textcolor{green}{0.07792} \\
\hline
\multirow{3}{*}{FreTS} 
 & 60 & 0.02635 & 0.02813 & & 0.02665 & \textcolor{green}{0.02516} & 0.02580 & \textcolor{blue}{0.02517} & 0.02647 & 0.02520 & 0.02558 \\
 & 90 & 0.04852 & 0.05346 & & 0.04894 & \textcolor{blue}{0.04432} & 0.04718 & 0.04436 & 0.04837 & 0.04446 & \textcolor{green}{0.04428} \\
 & 120 & 0.07669 & 0.08457 & & 0.07767 & \textcolor{blue}{0.06790} & 0.07542 & 0.06802 & 0.07704 & 0.06836 & \textcolor{green}{0.06736} \\
\hline
\multirow{3}{*}{PatchTST} 
 & 60 & 0.05240 & 0.03417 & & 0.03336 & 0.03156 & 0.04403 & 0.03151 & 0.03366 & \textcolor{blue}{0.03139} & \textcolor{green}{0.02936} \\
 & 90 & 0.10923 & 0.06663 & & 0.06762 & 0.06028 & 0.08248 & 0.06023 & 0.06484 & \textcolor{blue}{0.05988} & \textcolor{green}{0.05584} \\
 & 120 & 0.19164 & 0.11587 & & 0.12027 & \textcolor{blue}{0.10330} & 0.15357 & 0.10339 & 0.11577 & 0.10337 & \textcolor{green}{0.08818} \\
\hline
\multirow{3}{*}{OLS} 
 & 60 & 0.05930 & 0.04433 & & 0.03844 & 0.03375 & 0.04456 & 0.03383 & 0.03932 & \textcolor{blue}{0.03367} & \textcolor{green}{0.03216} \\
 & 90 & 0.12777 & 0.08605 & & 0.07723 & 0.06642 & 0.08503 & 0.06651 & 0.08104 & \textcolor{blue}{0.06617} & \textcolor{green}{0.06076} \\
 & 120 & 0.22788 & 0.14938 & & 0.14265 & \textcolor{blue}{0.11526} & 0.15923 & 0.11557 & 0.14362 & 0.11667 & \textcolor{green}{0.10120} \\
\hline
\end{tabular}
}
\end{subtable}
\end{table}

\begin{table}[htbp]
\ContinuedFloat
\centering
\begin{subtable}{\textwidth}
\renewcommand{\thetable}{A.1.16} 
\centering
\vspace{-15pt}
\caption{PETSA Macro and Market Series}
\label{PETSA index}
\renewcommand{\arraystretch}{1.2}
\resizebox{\textwidth}{!}{
\begin{tabular}{cc|cccccccccc}
\hline
\multirow{3}{*}{Model} & \multirow{3}{*}{$H$} & \multirow{3}{*}{Base} & \multirow{3}{*}{original} & mse & \checkmark & & & \checkmark & \checkmark & & \checkmark \\
 & & & & dir & & \checkmark & & \checkmark & & \checkmark & \checkmark \\
 & & & & price & & & \checkmark & & \checkmark & \checkmark & \checkmark \\
\hline
\multirow{3}{*}{iTransformer} 
 & 60 & 0.01886 & 0.01564 & & 0.01422 & 0.01209 & 0.01602 & 0.01211 & 0.01361 & \textcolor{blue}{0.01196} & \textcolor{green}{0.01157} \\
 & 90 & 0.04027 & 0.02911 & & 0.02804 & 0.02405 & 0.03165 & 0.02404 & 0.02737 & \textcolor{blue}{0.02279} & \textcolor{green}{0.02004} \\
 & 120 & 0.07258 & 0.04806 & & 0.04782 & 0.04021 & 0.05915 & 0.04006 & 0.04824 & \textcolor{blue}{0.03831} & \textcolor{green}{0.03643} \\
\hline
\multirow{3}{*}{DLinear} 
 & 60 & 0.01248 & 0.01361 & & 0.01309 & 0.01091 & 0.01249 & 0.01085 & 0.01268 & \textcolor{blue}{0.01082} & \textcolor{green}{0.01064} \\
 & 90 & 0.02567 & 0.02491 & & 0.02587 & 0.02114 & 0.02362 & 0.02111 & 0.02295 & \textcolor{blue}{0.02087} & \textcolor{green}{0.02017} \\
 & 120 & 0.04046 & 0.04127 & & 0.04111 & 0.03303 & 0.04040 & 0.03289 & 0.04087 & \textcolor{blue}{0.03266} & \textcolor{green}{0.03058} \\
\hline
\multirow{3}{*}{FreTS} 
 & 60 & \textcolor{blue}{0.00882} & 0.00933 & & 0.00901 & 0.00909 & \textcolor{green}{0.00879} & 0.00907 & 0.00899 & 0.00907 & 0.00928 \\
 & 90 & \textcolor{green}{0.01670} & 0.01752 & & 0.01694 & 0.01716 & \textcolor{blue}{0.01676} & 0.01711 & 0.01698 & 0.01708 & 0.01722 \\
 & 120 & \textcolor{green}{0.02586} & 0.02707 & & 0.02628 & 0.02615 & 0.02627 & \textcolor{blue}{0.02612} & 0.02649 & 0.02615 & 0.02624 \\
\hline
\multirow{3}{*}{PatchTST} 
 & 60 & 0.01793 & \textcolor{blue}{0.01152} & & 0.01171 & 0.01176 & 0.01504 & 0.01168 & 0.01155 & 0.01164 & \textcolor{green}{0.01081} \\
 & 90 & 0.03668 & 0.02402 & & 0.02521 & 0.02488 & 0.02933 & 0.02467 & 0.02417 & \textcolor{blue}{0.02389} & \textcolor{green}{0.02053} \\
 & 120 & 0.06088 & 0.04170 & & 0.04323 & 0.03983 & 0.05006 & 0.03955 & 0.03967 & \textcolor{blue}{0.03866} & \textcolor{green}{0.03350} \\
\hline
\multirow{3}{*}{OLS} 
 & 60 & 0.02018 & 0.01708 & & 0.01359 & 0.01213 & 0.01477 & \textcolor{blue}{0.01204} & 0.01307 & 0.01216 & \textcolor{green}{0.01112} \\
 & 90 & 0.04483 & 0.03102 & & 0.02809 & 0.02424 & 0.02880 & 0.02420 & 0.02505 & \textcolor{blue}{0.02363} & \textcolor{green}{0.02134} \\
 & 120 & 0.07908 & 0.05202 & & 0.04628 & 0.04014 & 0.05234 & 0.04018 & 0.04400 & \textcolor{blue}{0.03965} & \textcolor{green}{0.03511} \\
\hline
\end{tabular}
}
\end{subtable}
\end{table}
\FloatBarrier

\subsection{Additional Quantitative Results}
\label{appendix_additional_results}

\subsubsection{Relative Improvements in Reconstructed Price MSE by Dataset}
\label{appendix_relative_improvement}
To compare changes in reconstructed-price error after replacing the original adaptation objectives with DPAL, Tables \ref{tafas_relative_improvement} and \ref{petsa_relative_improvement} report the relative improvements obtained by replacing the original adaptation objectives of TAFAS \citep{TAFAS} and PETSA \citep{PETSA} with DPAL, respectively. The reported percentages are calculated from the MSE values shown in Tables \ref{main results tafas} and \ref{main results petsa}.

\begin{table}[H]
    \centering
    \footnotesize
    \setlength{\tabcolsep}{2pt}
    \begin{minipage}{0.40\textwidth}
        \renewcommand{\thetable}{A.2.1} 
        \centering
        \caption{TAFAS Relative Improvement}
        \label{tafas_relative_improvement}
        \renewcommand{\arraystretch}{1.2}
        \begin{tabular}{cc|ccccc}
        \hline
        Datasets & H & iTransformer & DLinear & FreTS & PatchTST & OLS \\
        \hline
        \multirow{3}{*}{\makecell{Magnificent 7}}
         & 60 & 51.9\% & 35.7\% & 9.8\% & 36.0\% & 53.9\% \\
         & 90 & 59.2\% & 52.7\% & 19.0\% & 50.3\% & 51.9\% \\
         & 120 & 69.4\% & 59.2\% & 20.5\% & 62.3\% & 55.8\% \\
        \hline
        \multirow{3}{*}{\makecell{S\&P 500\\Excerpt}}
         & 60 & 55.9\% & 31.7\% & 5.1\% & 31.9\% & 51.9\% \\
         & 90 & 62.3\% & 44.0\% & 9.9\% & 38.2\% & 55.5\% \\
         & 120 & 64.5\% & 51.3\% & 15.4\% & 32.1\% & 60.6\% \\
        \hline
        \multirow{3}{*}{\makecell{Russell 2000\\Excerpt}}
         & 60 & 52.9\% & 42.1\% & 4.3\% & 29.0\% & 45.6\% \\
         & 90 & 62.5\% & 50.2\% & 11.2\% & 41.6\% & 57.4\% \\
         & 120 & 63.1\% & 46.9\% & 8.1\% & 47.9\% & 60.3\% \\
        \hline
        \multirow{3}{*}{\makecell{Technology\\Sector}}
         & 60 & 52.3\% & 36.6\% & 9.2\% & 17.4\% & 48.0\% \\
         & 90 & 63.7\% & 49.1\% & 14.5\% & 32.5\% & 50.4\% \\
         & 120 & 76.2\% & 52.6\% & 16.1\% & 49.4\% & 68.3\% \\
        \hline
        \multirow{3}{*}{\makecell{Consumer\\Sector}}
         & 60 & 43.5\% & 21.4\% & 0.0\% & 25.0\% & 40.9\% \\
         & 90 & 49.0\% & 25.0\% & 5.6\% & 25.0\% & 44.9\% \\
         & 120 & 52.0\% & 26.1\% & 6.7\% & 25.9\% & 45.7\% \\
        \hline
        \multirow{3}{*}{\makecell{Energy\\Sector}}
         & 60 & 33.3\% & 14.8\% & 0.0\% & 23.3\% & 31.4\% \\
         & 90 & 39.2\% & 18.5\% & -2.9\% & 23.3\% & 35.5\% \\
         & 120 & 41.1\% & 21.5\% & -1.7\% & 22.7\% & 36.4\% \\
        \hline
        \multirow{3}{*}{\makecell{Nasdaq Mixed\\Excerpt}}
         & 60 & 39.2\% & 20.0\% & 0.0\% & 19.4\% & 28.9\% \\
         & 90 & 46.4\% & 27.1\% & 10.2\% & 25.4\% & 33.0\% \\
         & 120 & 47.2\% & 29.5\% & 11.7\% & 27.4\% & 36.7\% \\
        \hline
        \multirow{3}{*}{\makecell{Macro and\\Market Series}}
         & 60 & 31.3\% & 8.3\% & 0.0\% & 16.7\% & 26.7\% \\
         & 90 & 40.0\% & 20.0\% & 0.0\% & 20.8\% & 33.3\% \\
         & 120 & 41.0\% & 22.5\% & 0.0\% & 22.5\% & 38.6\% \\
        \hline
        \end{tabular}
    \end{minipage}
    \hfill
    \begin{minipage}{0.40\textwidth}
        \renewcommand{\thetable}{A.2.2} 
        \centering
        \caption{PETSA Relative Improvement}
        \label{petsa_relative_improvement}
        \renewcommand{\arraystretch}{1.2}
        \begin{tabular}{cc|ccccc}
        \hline
        Datasets & H & iTransformer & DLinear & FreTS & PatchTST & OLS \\
        \hline
        \multirow{3}{*}{Magnificent 7} 
         & 60 & 27.5\% & 34.4\% & 18.9\% & 31.0\% & 41.6\% \\
         & 90 & 46.6\% & 44.5\% & 25.9\% & 43.8\% & 47.1\% \\
         & 120 & 48.6\% & 56.9\% & 31.7\% & 49.8\% & 48.0\% \\
        \hline
        \multirow{3}{*}{\makecell{S\&P 500\\Excerpt}}
         & 60 & 29.1\% & 41.7\% & 13.8\% & 4.6\% & 30.6\% \\
         & 90 & 26.3\% & 49.1\% & 17.7\% & 5.2\% & 22.6\% \\
         & 120 & 27.2\% & 57.6\% & 25.5\% & 15.0\% & 25.3\% \\
        \hline
        \multirow{3}{*}{\makecell{Russell 2000\\Excerpt}}
         & 60 & 38.5\% & 45.7\% & 19.8\% & 20.8\% & 41.5\% \\
         & 90 & 75.2\% & 43.2\% & 24.4\% & 32.6\% & 44.3\% \\
         & 120 & 52.4\% & 40.9\% & 22.1\% & 33.8\% & 47.6\% \\
        \hline
        \multirow{3}{*}{\makecell{Technology\\Sector}}
         & 60 & 49.5\% & 52.8\% & 21.6\% & 27.0\% & 50.2\% \\
         & 90 & 58.0\% & 38.6\% & 29.2\% & 35.1\% & 60.6\% \\
         & 120 & 63.6\% & 55.6\% & 37.8\% & 24.5\% & 67.0\% \\
        \hline
        \multirow{3}{*}{\makecell{Consumer \\ Sector}}
         & 60 & 35.0\% & 31.3\% & 16.7\% & 25.0\% & 38.1\% \\
         & 90 & 42.2\% & 32.3\% & 22.7\% & 29.0\% & 39.5\% \\
         & 120 & 46.0\% & 32.7\% & 25.7\% & 29.6\% & 41.0\% \\
        \hline
        \multirow{3}{*}{\makecell{Energy\\Sector}}
         & 60 & 24.2\% & 17.9\% & 5.0\% & 6.9\% & 18.2\% \\
         & 90 & 23.2\% & 25.9\% & 5.3\% & 7.1\% & 22.5\% \\
         & 120 & 26.0\% & 25.0\% & 9.2\% & 11.0\% & 26.2\% \\
        \hline
        \multirow{3}{*}{\makecell{Nasdaq Mixed\\Excerpt}}
         & 60 & 24.4\% & 19.4\% & 7.1\% & 14.7\% & 27.3\% \\
         & 90 & 37.6\% & 27.5\% & 17.0\% & 16.4\% & 29.1\% \\
         & 120 & 39.4\% & 29.1\% & 21.2\% & 24.1\% & 32.2\% \\
        \hline
        \multirow{3}{*}{\makecell{Macro and\\Market Series}}
         & 60 & 25.0\% & 21.4\% & 0.0\% & 8.3\% & 35.3\% \\
         & 90 & 31.0\% & 20.0\% & 5.6\% & 12.5\% & 32.3\% \\
         & 120 & 25.0\% & 24.4\% & 3.7\% & 19.0\% & 32.7\% \\
        \hline
        \end{tabular}
    \end{minipage}
    \vspace{-15pt}
\end{table}

For an adaptation framework ${f}\in\{\mathrm{TAFAS},\mathrm{PETSA}\}$, the relative improvement is defined as$$I_{f} = \frac{\operatorname{MSE}_{f} - \operatorname{MSE}_{f\text{+}\mathrm{DPAL}}}{\operatorname{MSE}_{f}} \times 100\%.$$
A positive value indicates that the setting using DPAL has a lower MSE of reconstructed prices than the corresponding baseline framework, whereas a negative value indicates a higher error.

For each framework, positive relative improvements occur in most combinations of datasets, forecasting backbones, and forecast horizons. The magnitude of the improvement varies across datasets, forecasting backbones, and forecast horizons. 

Relative improvements are calculated from the MSE values rounded to three decimal places in Tables \ref{main results tafas} and \ref{main results petsa}. Percentages close to zero may be affected by rounding.

\subsubsection{Pointwise Errors in Standardized Return Space}
\label{appendix_standardized_return_errors}
The primary evaluation in the main results measures the MSE of reconstructed relative prices. For completeness, Tables~\ref{tafas mse}--\ref{petsa mae} additionally report pointwise MSE and MAE computed directly in the standardized return space. These MSE and MAE are calculated before inverse transformation and price reconstruction. Lower values indicate better pointwise return accuracy. All comparison counts in this subsection are calculated using the unrounded results, whereas the values displayed in the tables are rounded to three decimal places.

\begin{table*}[htbp]
\renewcommand{\thetable}{A.2.3} 
\centering
\caption{Forecasting performance of TAFAS measured by MSE in the standardized return space. Lower values are better. The best result within each backbone and horizon is highlighted in green.}
\label{tafas mse}
\renewcommand{\arraystretch}{1.2} 
\resizebox{\textwidth}{!}{
\begin{tabular}{cc|lcr|lcr|lcr|lcr|lcr}
\hline
\multirow{2}{*}{Datasets} & \multirow{2}{*}{$H$} & \multicolumn{3}{c|}{iTransformer} & \multicolumn{3}{c|}{DLinear} & \multicolumn{3}{c|}{FreTS} & \multicolumn{3}{c|}{PatchTST} & \multicolumn{3}{c}{OLS} \\
\cline{3-5} \cline{6-8} \cline{9-11} \cline{12-14} \cline{15-17}
 & & Base & TAFAS & \makecell{with\\DPAL} & Base & TAFAS & \makecell{with\\DPAL} & Base & TAFAS & \makecell{with\\DPAL} & Base & TAFAS & \makecell{with\\DPAL} & Base & TAFAS & \makecell{with\\DPAL} \\
\hline
\multirow{3}{*}{Magnificent 7} 
 & 60 & 1.150 & 1.139 & \textcolor{green}{1.126} & 1.114 & 1.113 & \textcolor{green}{1.110} & 1.135 & 1.134 & \textcolor{green}{1.125} & 1.145 & 1.129 & \textcolor{green}{1.128} & 1.139 & 1.127 & \textcolor{green}{1.120} \\ 
 & 90 & 1.152 & 1.141 & \textcolor{green}{1.135} & 1.123 & 1.122 & \textcolor{green}{1.119} & 1.137 & 1.136 & \textcolor{green}{1.129} & 1.154 & \textcolor{green}{1.136} & 1.139 & 1.146 & 1.133 & \textcolor{green}{1.130} \\ 
 & 120 & 1.162 & 1.152 & \textcolor{green}{1.144} & 1.133 & 1.132 & \textcolor{green}{1.130} & 1.144 & 1.143 & \textcolor{green}{1.138} & 1.160 & 1.144 & \textcolor{green}{1.144} & 1.155 & 1.144 & \textcolor{green}{1.140} \\ 
\hline
\multirow{3}{*}{\makecell{S\&P 500\\Excerpt}} 
 & 60 & 1.513 & 1.508 & \textcolor{green}{1.488} & 1.476 & 1.476 & \textcolor{green}{1.472} & 1.502 & 1.501 & \textcolor{green}{1.489} & 1.507 & 1.494 & \textcolor{green}{1.486} & 1.500 & 1.494 & \textcolor{green}{1.482} \\ 
 & 90 & 1.497 & 1.493 & \textcolor{green}{1.476} & 1.471 & 1.470 & \textcolor{green}{1.466} & 1.485 & 1.485 & \textcolor{green}{1.477} & 1.502 & 1.491 & \textcolor{green}{1.483} & 1.491 & 1.486 & \textcolor{green}{1.475} \\ 
 & 120 & 1.496 & 1.493 & \textcolor{green}{1.476} & 1.469 & 1.469 & \textcolor{green}{1.464} & 1.480 & 1.480 & \textcolor{green}{1.474} & 1.495 & 1.485 & \textcolor{green}{1.479} & 1.489 & 1.484 & \textcolor{green}{1.475} \\ 
\hline
\multirow{3}{*}{\makecell{Russell 2000\\Excerpt}}
 & 60 & 1.116 & 1.110 & \textcolor{green}{1.092} & 1.081 & 1.080 & \textcolor{green}{1.075} & 1.100 & 1.099 & \textcolor{green}{1.088} & 1.102 & 1.093 & \textcolor{green}{1.086} & 1.103 & 1.096 & \textcolor{green}{1.088} \\ 
 & 90 & 1.118 & 1.112 & \textcolor{green}{1.097} & 1.092 & 1.091 & \textcolor{green}{1.085} & 1.103 & 1.103 & \textcolor{green}{1.095} & 1.110 & 1.101 & \textcolor{green}{1.096} & 1.110 & 1.103 & \textcolor{green}{1.096} \\ 
 & 120 & 1.120 & 1.116 & \textcolor{green}{1.103} & 1.098 & 1.098 & \textcolor{green}{1.093} & 1.105 & 1.105 & \textcolor{green}{1.099} & 1.116 & 1.107 & \textcolor{green}{1.103} & 1.116 & 1.110 & \textcolor{green}{1.103} \\ 
\hline
\multirow{3}{*}{\makecell{Technology\\Sector}}
 & 60 & 1.465 & 1.457 & \textcolor{green}{1.440} & 1.433 & 1.432 & \textcolor{green}{1.429} & 1.451 & 1.450 & \textcolor{green}{1.441} & 1.459 & 1.444 & \textcolor{green}{1.440} & 1.452 & 1.444 & \textcolor{green}{1.437} \\ 
 & 90 & 1.458 & 1.450 & \textcolor{green}{1.439} & 1.436 & 1.435 & \textcolor{green}{1.431} & 1.444 & 1.444 & \textcolor{green}{1.438} & 1.457 & 1.446 & \textcolor{green}{1.442} & 1.452 & 1.443 & \textcolor{green}{1.438} \\ 
 & 120 & 1.458 & 1.453 & \textcolor{green}{1.441} & 1.436 & 1.435 & \textcolor{green}{1.432} & 1.442 & 1.442 & \textcolor{green}{1.437} & 1.453 & 1.444 & \textcolor{green}{1.439} & 1.451 & 1.445 & \textcolor{green}{1.439} \\ 
\hline
\multirow{3}{*}{\makecell{Consumer\\Sector}} 
 & 60 & 1.219 & 1.214 & \textcolor{green}{1.196} & 1.192 & 1.191 & \textcolor{green}{1.186} & 1.205 & 1.205 & \textcolor{green}{1.195} & 1.213 & 1.205 & \textcolor{green}{1.197} & 1.208 & 1.203 & \textcolor{green}{1.192} \\ 
 & 90 & 1.223 & 1.219 & \textcolor{green}{1.207} & 1.203 & 1.202 & \textcolor{green}{1.197} & 1.210 & 1.210 & \textcolor{green}{1.203} & 1.222 & 1.215 & \textcolor{green}{1.211} & 1.218 & 1.213 & \textcolor{green}{1.206} \\ 
 & 120 & 1.241 & 1.238 & \textcolor{green}{1.228} & 1.220 & 1.219 & \textcolor{green}{1.215} & 1.226 & 1.226 & \textcolor{green}{1.219} & 1.237 & 1.231 & \textcolor{green}{1.228} & 1.236 & 1.231 & \textcolor{green}{1.226} \\ 
\hline
\multirow{3}{*}{\makecell{Energy\\Sector}} 
 & 60 & 0.576 & 0.574 & \textcolor{green}{0.562} & 0.557 & 0.555 & \textcolor{green}{0.550} & 0.567& 0.566 & \textcolor{green}{0.560} & 0.571 & 0.565 & \textcolor{green}{0.560} & 0.565 & 0.563 & \textcolor{green}{0.559} \\ 
 & 90 & 0.573 & 0.571 & \textcolor{green}{0.562} & 0.558 & 0.557 & \textcolor{green}{0.551} & 0.566 & 0.566 & \textcolor{green}{0.561} & 0.572 & 0.566 & \textcolor{green}{0.562} & 0.567 & 0.565 & \textcolor{green}{0.561} \\ 
 & 120 & 0.578 & 0.577 & \textcolor{green}{0.570} & 0.562 & 0.562 & \textcolor{green}{0.557} & 0.569 & 0.569 & \textcolor{green}{0.565} & 0.574 & 0.570 & \textcolor{green}{0.567} & 0.572 & 0.571 & \textcolor{green}{0.567} \\ 
\hline
\multirow{3}{*}{\makecell{Nasdaq Mixed\\Excerpt}} 
 & 60 & 0.969 & 0.964 & \textcolor{green}{0.947} & 0.947 & 0.946 & \textcolor{green}{0.940} & 0.958 & 0.957 & \textcolor{green}{0.947} & 0.963 & 0.956 & \textcolor{green}{0.948} & 0.961 & 0.957 & \textcolor{green}{0.947} \\ 
 & 90 & 0.963 & 0.960 & \textcolor{green}{0.947} & 0.948 & 0.947 & \textcolor{green}{0.942} & 0.953 & 0.953 & \textcolor{green}{0.946} & 0.962 & 0.956 & \textcolor{green}{0.949} & 0.960 & 0.956 & \textcolor{green}{0.949} \\ 
 & 120 & 0.965 & 0.962 & \textcolor{green}{0.952} & 0.950 & 0.949 & \textcolor{green}{0.945} & 0.953 & 0.953 & \textcolor{green}{0.948} & 0.962 & 0.957 & \textcolor{green}{0.952} & 0.962 & 0.959 & \textcolor{green}{0.953} \\ 
\hline
\multirow{3}{*}{\makecell{Macro and\\Market Series}}
 & 60 & 0.792 & 0.786 & \textcolor{green}{0.774} & 0.769 & 0.767 & \textcolor{green}{0.765} & 0.787 & 0.786 & \textcolor{green}{0.776} & 0.788 & 0.780 & \textcolor{green}{0.774} & 0.786 & 0.779 & \textcolor{green}{0.774} \\ 
 & 90 & 0.793 & 0.788 & \textcolor{green}{0.778} & 0.777 & 0.774 & \textcolor{green}{0.772} & 0.788 & 0.787 & \textcolor{green}{0.780} & 0.796 & 0.789 & \textcolor{green}{0.784} & 0.790 & 0.784 & \textcolor{green}{0.780} \\ 
 & 120 & 0.803 & 0.799 & \textcolor{green}{0.791} & 0.788 & 0.786 & \textcolor{green}{0.784} & 0.797 & 0.791 & \textcolor{green}{0.791} & 0.805 & 0.799 & \textcolor{green}{0.796} & 0.800 & 0.794 & \textcolor{green}{0.791} \\ 
\hline
\end{tabular}
}
\end{table*}

\begin{table*}[htbp]
\renewcommand{\thetable}{A.2.4} 
\centering
\caption{Forecasting performance of TAFAS measured by MAE in the standardized return space. Lower values are better.  The best result within each backbone and horizon is highlighted in green.}
\label{tafas mae}
\renewcommand{\arraystretch}{1.2} 
\resizebox{\textwidth}{!}{
\begin{tabular}{cc|lcr|lcr|lcr|lcr|lcr}
\hline
\multirow{2}{*}{Datasets} & \multirow{2}{*}{$H$} & \multicolumn{3}{c|}{iTransformer} & \multicolumn{3}{c|}{DLinear} & \multicolumn{3}{c|}{FreTS} & \multicolumn{3}{c|}{PatchTST} & \multicolumn{3}{c}{OLS} \\
\cline{3-5} \cline{6-8} \cline{9-11} \cline{12-14} \cline{15-17}
 & & Base & TAFAS & \makecell{with\\DPAL} & Base & TAFAS & \makecell{with\\DPAL} & Base & TAFAS & \makecell{with\\DPAL} & Base & TAFAS & \makecell{with\\DPAL} & Base & TAFAS & \makecell{with\\DPAL} \\
\hline
\multirow{3}{*}{Magnificent 7} 
 & 60 & 0.761 & 0.755 & \textcolor{green}{0.750} & 0.742 & 0.742 & \textcolor{green}{0.741} & 0.754 & 0.753 & \textcolor{green}{0.748} & 0.758 & 0.751 & \textcolor{green}{0.750} & 0.756 & 0.750 & \textcolor{green}{0.747} \\ 
 & 90 & 0.759 & 0.754 & \textcolor{green}{0.752} & 0.745 & 0.744 & \textcolor{green}{0.742} & 0.752 & 0.752 & \textcolor{green}{0.748} & 0.760 & \textcolor{green}{0.752} & 0.753 & 0.756 & 0.750 & \textcolor{green}{0.749} \\ 
 & 120 & 0.761 & 0.757 & \textcolor{green}{0.754} & 0.747 & 0.746 & \textcolor{green}{0.746} & 0.753 & 0.753 & \textcolor{green}{0.750} & 0.760 & \textcolor{green}{0.753} & 0.753 & 0.757 & 0.753 & \textcolor{green}{0.751} \\ 
\hline
\multirow{3}{*}{\makecell{S\&P 500\\Excerpt}} 
 & 60 & 0.822 & 0.820 & \textcolor{green}{0.811} & 0.806 & 0.806 & \textcolor{green}{0.805} & 0.817 & 0.817 & \textcolor{green}{0.812} & 0.820 & 0.814 & \textcolor{green}{0.811} & 0.817 & 0.814 & \textcolor{green}{0.809} \\ 
 & 90 & 0.816 & 0.814 & \textcolor{green}{0.807} & 0.805 & 0.805 & \textcolor{green}{0.803} & 0.811 & 0.811 & \textcolor{green}{0.808} & 0.818 & 0.813 & \textcolor{green}{0.810} & 0.813 & 0.811 & \textcolor{green}{0.807} \\ 
 & 120 & 0.816 & 0.816 & \textcolor{green}{0.809} & 0.806 & 0.806 & \textcolor{green}{0.804} & 0.811 & 0.810 & \textcolor{green}{0.808} & 0.816 & 0.812 & \textcolor{green}{0.810} & 0.814 & 0.812 & \textcolor{green}{0.809} \\ 
\hline
\multirow{3}{*}{\makecell{Russell 2000\\Excerpt}} 
 & 60 & 0.714 & 0.712 & \textcolor{green}{0.701} & 0.696 & 0.695 & \textcolor{green}{0.692} & 0.705 & 0.705 & \textcolor{green}{0.699} & 0.706 & 0.701 & \textcolor{green}{0.698} & 0.707 & 0.704 & \textcolor{green}{0.699} \\ 
 & 90 & 0.711 & 0.709 & \textcolor{green}{0.701} & 0.698 & 0.697 & \textcolor{green}{0.694} & 0.704 & 0.704 & \textcolor{green}{0.700} & 0.707 & 0.703 & \textcolor{green}{0.700} & 0.707 & 0.704 & \textcolor{green}{0.700} \\ 
 & 120 & 0.710 & 0.708 & \textcolor{green}{0.702} & 0.699 & 0.699 & \textcolor{green}{0.696} & 0.703 & 0.703 & \textcolor{green}{0.700} & 0.707 & 0.704 & \textcolor{green}{0.702} & 0.707 & 0.705 & \textcolor{green}{0.702} \\ 
\hline
\multirow{3}{*}{\makecell{Technology\\Sector}} 
 & 60 & 0.792 & 0.788 & \textcolor{green}{0.780} & 0.777 & 0.777 & \textcolor{green}{0.775} & 0.785 & 0.785 & \textcolor{green}{0.781} & 0.790 & 0.782 & \textcolor{green}{0.780} & 0.786 & 0.782 & \textcolor{green}{0.779} \\ 
 & 90 & 0.787 & 0.783 & \textcolor{green}{0.778} & 0.777 & 0.776 & \textcolor{green}{0.774} & 0.781 & 0.781 & \textcolor{green}{0.778} & 0.786 & 0.782 & \textcolor{green}{0.779} & 0.784 & 0.780 & \textcolor{green}{0.778} \\ 
 & 120 & 0.787 & 0.785 & \textcolor{green}{0.780} & 0.777 & 0.777 & \textcolor{green}{0.775} & 0.780 & 0.780 & \textcolor{green}{0.778} & 0.785 & 0.781 & \textcolor{green}{0.779} & 0.784 & 0.781 & \textcolor{green}{0.779} \\ 
\hline
\multirow{3}{*}{\makecell{Consumer\\Sector}} 
 & 60 & 0.759 & 0.757 & \textcolor{green}{0.749} & 0.748 & 0.748 & \textcolor{green}{0.746} & 0.754 & 0.753 & \textcolor{green}{0.749} & 0.757 & 0.754 & \textcolor{green}{0.750} & 0.755 & 0.753 & \textcolor{green}{0.748} \\ 
 & 90 & 0.758 & 0.756 & \textcolor{green}{0.751} & 0.751 & 0.750 & \textcolor{green}{0.748} & 0.753 & 0.753 & \textcolor{green}{0.750} & 0.758 & 0.755 & \textcolor{green}{0.753} & 0.756 & 0.755 & \textcolor{green}{0.751} \\ 
 & 120 & 0.761 & 0.760 & \textcolor{green}{0.756} & 0.753 & 0.753 & \textcolor{green}{0.751} & 0.756 & 0.756 & \textcolor{green}{0.753} & 0.760 & 0.758 & \textcolor{green}{0.757} & 0.759 & 0.758 & \textcolor{green}{0.756} \\ 
\hline
\multirow{3}{*}{\makecell{Energy\\Sector}} 
 & 60 & 0.541 & 0.540 & \textcolor{green}{0.532} & 0.528 & 0.527 & \textcolor{green}{0.523} & 0.535 & 0.535 & \textcolor{green}{0.531} & 0.538 & 0.534 & \textcolor{green}{0.532} & 0.534 & 0.533 & \textcolor{green}{0.530} \\ 
 & 90 & 0.537 & 0.536 & \textcolor{green}{0.530} & 0.526 & 0.525 & \textcolor{green}{0.522} & 0.532 & 0.532 & \textcolor{green}{0.529} & 0.537 & 0.533 & \textcolor{green}{0.530} & 0.532 & 0.532 & \textcolor{green}{0.529} \\ 
 & 120 & 0.537 & 0.537 & \textcolor{green}{0.532} & 0.527 & 0.526 & \textcolor{green}{0.523} & 0.531 & 0.531 & \textcolor{green}{0.529} & 0.535 & 0.532 & \textcolor{green}{0.530} & 0.533 & 0.532 & \textcolor{green}{0.530} \\ 
\hline
\multirow{3}{*}{\makecell{Nasdaq Mixed\\Excerpt}} 
 & 60 & 0.682 & 0.679 & \textcolor{green}{0.671} & 0.671 & 0.670 & \textcolor{green}{0.668} & 0.677 & 0.676 & \textcolor{green}{0.672} & 0.679 & 0.675 & \textcolor{green}{0.672} & 0.678 & 0.675 & \textcolor{green}{0.672} \\ 
 & 90 & 0.678 & 0.676 & \textcolor{green}{0.671} & 0.671 & 0.670 & \textcolor{green}{0.668} & 0.674 & 0.673 & \textcolor{green}{0.671} & 0.678 & 0.675 & \textcolor{green}{0.672} & 0.677 & 0.675 & \textcolor{green}{0.672} \\ 
 & 120 & 0.678 & 0.677 & \textcolor{green}{0.673} & 0.671 & 0.671 & \textcolor{green}{0.669} & 0.673 & 0.673 & \textcolor{green}{0.671} & 0.677 & 0.675 & \textcolor{green}{0.673} & 0.677 & 0.675 & \textcolor{green}{0.673} \\ 
\hline
\multirow{3}{*}{\makecell{Macro and\\Market Series}} 
 & 60 & 0.616 & 0.613 & \textcolor{green}{0.608} & 0.605 & 0.604 & \textcolor{green}{0.603} & 0.616 & 0.615 & \textcolor{green}{0.609} & 0.614 & 0.611 & \textcolor{green}{0.609} & 0.612 & 0.610 & \textcolor{green}{0.608} \\ 
 & 90 & 0.614 & 0.612 & \textcolor{green}{0.608} & 0.607 & 0.605 & \textcolor{green}{0.604} & 0.613 & 0.613 & \textcolor{green}{0.609} & 0.616 & 0.613 & \textcolor{green}{0.612} & 0.612 & 0.610 & \textcolor{green}{0.609} \\ 
 & 120 & 0.617 & 0.615 & \textcolor{green}{0.613} & 0.610 & 0.609 & \textcolor{green}{0.608} & 0.615 & 0.614 & \textcolor{green}{0.612} & 0.619 & 0.616 & \textcolor{green}{0.615} & 0.613 & 0.613 & \textcolor{green}{0.613} \\ 
\hline
\end{tabular}
}
\end{table*}

\begin{table*}[htbp]
\renewcommand{\thetable}{A.2.5}
\centering
\caption{Forecasting performance of PETSA measured by MSE in the standardized return space. Lower values are better. The best result within each backbone and horizon is highlighted in green.}
\label{petsa mse}
\renewcommand{\arraystretch}{1.2} 
\resizebox{\textwidth}{!}{
\begin{tabular}{cc|lcr|lcr|lcr|lcr|lcr}
\hline
\multirow{2}{*}{Datasets} & \multirow{2}{*}{$H$} & \multicolumn{3}{c|}{iTransformer} & \multicolumn{3}{c|}{DLinear} & \multicolumn{3}{c|}{FreTS} & \multicolumn{3}{c|}{PatchTST} & \multicolumn{3}{c}{OLS} \\
\cline{3-5} \cline{6-8} \cline{9-11} \cline{12-14} \cline{15-17}
 & & Base & PETSA & \makecell{with\\DPAL} & Base & PETSA & \makecell{with\\DPAL} & Base & PETSA & \makecell{with\\DPAL} & Base & PETSA & \makecell{with\\DPAL} & Base & PETSA & \makecell{with\\DPAL} \\
\hline
\multirow{3}{*}{Magnificent 7} 
 & 60 & 1.150 & 1.144 & \textcolor{green}{1.143} & 1.114 & 1.123 & \textcolor{green}{1.112} & 1.135 & 1.136 & \textcolor{green}{1.132} & 1.145 & \textcolor{green}{1.136} & 1.138 & 1.139 & 1.137 & \textcolor{green}{1.129} \\ 
 & 90 & 1.152 & 1.144 & \textcolor{green}{1.138} & 1.123 & 1.123 & \textcolor{green}{1.121} & 1.137 & 1.137 & \textcolor{green}{1.135} & 1.154 & 1.142 & \textcolor{green}{1.141} & 1.146 & 1.137 & \textcolor{green}{1.135} \\ 
 & 120 & 1.162 & 1.153 & \textcolor{green}{1.145} & 1.133 & 1.133 & \textcolor{green}{1.131} & 1.144 & 1.145 & \textcolor{green}{1.142} & 1.160 & 1.150 & \textcolor{green}{1.145} & 1.155 & 1.147 & \textcolor{green}{1.144} \\ 
\hline
\multirow{3}{*}{\makecell{S\&P 500\\Excerpt}}
 & 60 & 1.513 & 1.515 & \textcolor{green}{1.490} & 1.476 & 1.477 & \textcolor{green}{1.472} & 1.502 & 1.502 & \textcolor{green}{1.494} & 1.507 & 1.503 & \textcolor{green}{1.489} & 1.500 & 1.502 & \textcolor{green}{1.485} \\ 
 & 90 & 1.497 & 1.492 & \textcolor{green}{1.478} & 1.471 & 1.472 & \textcolor{green}{1.466} & 1.485 & 1.485 & \textcolor{green}{1.481} & 1.502 & 1.493 & \textcolor{green}{1.488} & 1.491 & 1.486 & \textcolor{green}{1.477} \\ 
 & 120 & 1.496 & 1.490 & \textcolor{green}{1.479} & 1.469 & 1.470 & \textcolor{green}{1.465} & 1.480 & 1.480 & \textcolor{green}{1.477} & 1.495 & 1.485 & \textcolor{green}{1.482} & 1.489 & 1.483 & \textcolor{green}{1.476} \\ 
\hline
\multirow{3}{*}{\makecell{Russell 2000\\Excerpt}} 
 & 60 & 1.116 & 1.113 & \textcolor{green}{1.100} & 1.081 & 1.083 & \textcolor{green}{1.076} & 1.100 & 1.100 & \textcolor{green}{1.093} & 1.102 & \textcolor{green}{1.095} & 1.095 & 1.103 & 1.102 & \textcolor{green}{1.093} \\ 
 & 90 & 1.118 & 1.114 & \textcolor{green}{1.100} & 1.092 & 1.092 & \textcolor{green}{1.086} & 1.103 & 1.104 & \textcolor{green}{1.098} & 1.110 & 1.103 & \textcolor{green}{1.102} & 1.110 & 1.111 & \textcolor{green}{1.100} \\ 
 & 120 & 1.120 & 1.118 & \textcolor{green}{1.104} & 1.098 & 1.098 & \textcolor{green}{1.093} & 1.105 & 1.106 & \textcolor{green}{1.102} & 1.116 & 1.110 & \textcolor{green}{1.106} & 1.116 & 1.113 & \textcolor{green}{1.105} \\ 
\hline
\multirow{3}{*}{\makecell{Technology\\Sector}} 
 & 60 & 1.465 & 1.461 & \textcolor{green}{1.445} & 1.433 & 1.436 & \textcolor{green}{1.430} & 1.451 & 1.453 & \textcolor{green}{1.446} & 1.459 & 1.453 & \textcolor{green}{1.442} & 1.452 & 1.454 & \textcolor{green}{1.440} \\ 
 & 90 & 1.458 & 1.454 & \textcolor{green}{1.444} & 1.436 & 1.438 & \textcolor{green}{1.432} & 1.444 & 1.445 & \textcolor{green}{1.442} & 1.457 & 1.448 & \textcolor{green}{1.446} & 1.452 & 1.451 & \textcolor{green}{1.441} \\ 
 & 120 & 1.458 & 1.456 & \textcolor{green}{1.444} & 1.436 & 1.438 & \textcolor{green}{1.432} & 1.442 & 1.443 & \textcolor{green}{1.440} & 1.453 & 1.447 & \textcolor{green}{1.443} & 1.451 & 1.448 & \textcolor{green}{1.441} \\ 
\hline
\multirow{3}{*}{\makecell{Consumer\\Sector}} 
 & 60 & 1.219 & 1.216 & \textcolor{green}{1.199} & 1.192 & 1.193 & \textcolor{green}{1.186} & 1.205 & 1.206 & \textcolor{green}{1.199} & 1.213 & 1.208 & \textcolor{green}{1.201} & 1.208 & 1.208 & \textcolor{green}{1.195} \\ 
 & 90 & 1.223 & 1.220 & \textcolor{green}{1.210} & 1.203 & 1.204 & \textcolor{green}{1.198} & 1.210 & 1.211 & \textcolor{green}{1.206} & 1.222 & 1.217 & \textcolor{green}{1.213} & 1.218 & 1.218 & \textcolor{green}{1.210} \\ 
 & 120 & 1.241 & 1.241 & \textcolor{green}{1.230} & 1.220 & 1.220 & \textcolor{green}{1.216} & 1.226 & 1.226 & \textcolor{green}{1.222} & 1.237 & 1.233 & \textcolor{green}{1.229} & 1.236 & 1.235 & \textcolor{green}{1.231} \\ 
\hline
\multirow{3}{*}{\makecell{Energy\\Sector}} 
 & 60 & 0.576 & 0.575 & \textcolor{green}{0.563} & 0.557 & 0.558 & \textcolor{green}{0.551} & 0.567 & 0.568 & \textcolor{green}{0.564} & 0.571 & 0.567 & \textcolor{green}{0.564} & 0.565 & 0.565 & \textcolor{green}{0.562} \\ 
 & 90 & 0.573 & 0.571 & \textcolor{green}{0.562} & 0.558 & 0.558 & \textcolor{green}{0.551} & 0.566 & 0.567 & \textcolor{green}{0.564} & 0.572 & 0.571 & \textcolor{green}{0.564} & 0.567 & 0.567 & \textcolor{green}{0.564} \\ 
 & 120 & 0.578 & 0.577 & \textcolor{green}{0.570} & 0.562 & 0.563 & \textcolor{green}{0.557} & 0.569 & 0.569 & \textcolor{green}{0.567} & 0.574 & 0.574 & \textcolor{green}{0.568} & 0.572 & 0.572 & \textcolor{green}{0.569} \\ 
\hline
\multirow{3}{*}{\makecell{Nasdaq Mixed\\Excerpt}}
 & 60 & 0.969 & 0.967 & \textcolor{green}{0.950} & 0.947 & 0.950 & \textcolor{green}{0.942} & 0.957 & 0.959 & \textcolor{green}{0.951} & 0.963 & 0.959 & \textcolor{green}{0.951} & 0.961 & 0.966 & \textcolor{green}{0.951} \\ 
 & 90 & 0.963 & 0.961 & \textcolor{green}{0.950} & 0.948 & 0.949 & \textcolor{green}{0.943} & 0.953 & 0.954 & \textcolor{green}{0.948} & 0.962 & 0.957 & \textcolor{green}{0.953} & 0.960 & 0.965 & \textcolor{green}{0.952} \\ 
 & 120 & 0.965 & 0.963 & \textcolor{green}{0.954} & 0.950 & 0.950 & \textcolor{green}{0.945} & 0.953 & 0.954 & \textcolor{green}{0.949} & 0.962 & 0.958 & \textcolor{green}{0.954} & 0.962 & 0.961 & \textcolor{green}{0.955} \\ 
\hline
\multirow{3}{*}{\makecell{Macro and\\Market Series}}
 & 60 & 0.792 & 0.790 & \textcolor{green}{0.779} & 0.769 & 0.769 & \textcolor{green}{0.765} & 0.787 & 0.787 & \textcolor{green}{0.782} & 0.788 & 0.784 & \textcolor{green}{0.777} & 0.786 & 0.788 & \textcolor{green}{0.780} \\ 
 & 90 & 0.793 & 0.790 & \textcolor{green}{0.781} & 0.777 & 0.775 & \textcolor{green}{0.771} & 0.788 & 0.787 & \textcolor{green}{0.784} & 0.796 & 0.792 & \textcolor{green}{0.787} & 0.790 & 0.786 & \textcolor{green}{0.783} \\ 
 & 120 & 0.803 & 0.801 & \textcolor{green}{0.792} & 0.788 & 0.787 & \textcolor{green}{0.783} & 0.796 & 0.796 & \textcolor{green}{0.793} & 0.805 & 0.801 & \textcolor{green}{0.799} & 0.800 & 0.796 & \textcolor{green}{0.794} \\ 
\hline
\end{tabular}
}
\end{table*}

\begin{table*}[htbp]
\renewcommand{\thetable}{A.2.6}
\centering
\caption{Forecasting performance of PETSA measured by MAE in the standardized return space. Lower values are better. The best result within each backbone and horizon is highlighted in green.}
\label{petsa mae}
\renewcommand{\arraystretch}{1.2} 
\resizebox{\textwidth}{!}{
\begin{tabular}{cc|lcr|lcr|lcr|lcr|lcr}
\hline
\multirow{2}{*}{Datasets} & \multirow{2}{*}{$H$} & \multicolumn{3}{c|}{iTransformer} & \multicolumn{3}{c|}{DLinear} & \multicolumn{3}{c|}{FreTS} & \multicolumn{3}{c|}{PatchTST} & \multicolumn{3}{c}{OLS} \\
\cline{3-5} \cline{6-8} \cline{9-11} \cline{12-14} \cline{15-17}
 & & Base & PETSA & \makecell{with\\DPAL} & Base & PETSA & \makecell{with\\DPAL} & Base & PETSA & \makecell{with\\DPAL} & Base & PETSA & \makecell{with\\DPAL} & Base & PETSA & \makecell{with\\DPAL} \\
\hline
\multirow{3}{*}{Magnificent 7} 
 & 60 & 0.761 & 0.759 & \textcolor{green}{0.757} & 0.742 & 0.745 & \textcolor{green}{0.742} & 0.754 & 0.753 & \textcolor{green}{0.751} & 0.758 & \textcolor{green}{0.754} & 0.755 & 0.756 & 0.754 & \textcolor{green}{0.751} \\ 
 & 90 & 0.759 & 0.756 & \textcolor{green}{0.752} & 0.745 & 0.744 & \textcolor{green}{0.743} & 0.752 & 0.752 & \textcolor{green}{0.751} & 0.760 & 0.755 & \textcolor{green}{0.754} & 0.756 & \textcolor{green}{0.751} & 0.752 \\ 
 & 120 & 0.761 & 0.757 & \textcolor{green}{0.754} & 0.747 & 0.747 & \textcolor{green}{0.746} & 0.753 & 0.754 & \textcolor{green}{0.752} & 0.760 & 0.755 & \textcolor{green}{0.753} & 0.757 & 0.754 & \textcolor{green}{0.754} \\ 
\hline
\multirow{3}{*}{\makecell{S\&P 500\\Excerpt}} 
 & 60 & 0.822 & 0.822 & \textcolor{green}{0.813} & 0.806 & 0.806 & \textcolor{green}{0.805} & 0.817 & 0.817 & \textcolor{green}{0.814} & 0.820 & 0.817 & \textcolor{green}{0.812} & 0.817 & 0.817 & \textcolor{green}{0.811} \\ 
 & 90 & 0.816 & 0.814 & \textcolor{green}{0.808} & 0.805 & 0.805 & \textcolor{green}{0.803} & 0.811 & 0.811 & \textcolor{green}{0.810} & 0.818 & 0.814 & \textcolor{green}{0.813} & 0.813 & 0.811 & \textcolor{green}{0.808} \\ 
 & 120 & 0.816 & 0.815 & \textcolor{green}{0.811} & 0.806 & 0.806 & \textcolor{green}{0.805} & 0.811 & 0.810 & \textcolor{green}{0.810} & 0.816 & 0.812 & \textcolor{green}{0.812} & 0.814 & 0.812 & \textcolor{green}{0.809} \\ 
\hline
\multirow{3}{*}{\makecell{Russell 2000\\Excerpt}} 
 & 60 & 0.714 & 0.712 & \textcolor{green}{0.706} & 0.696 & 0.697 & \textcolor{green}{0.693} & 0.705 & 0.705 & \textcolor{green}{0.702} & 0.706 & \textcolor{green}{0.702} & 0.703 & 0.707 & 0.706 & \textcolor{green}{0.702} \\ 
 & 90 & 0.711 & 0.710 & \textcolor{green}{0.703} & 0.698 & 0.698 & \textcolor{green}{0.695} & 0.704 & 0.704 & \textcolor{green}{0.701} & 0.707 & 0.703 & \textcolor{green}{0.703} & 0.707 & 0.707 & \textcolor{green}{0.702} \\ 
 & 120 & 0.710 & 0.709 & \textcolor{green}{0.703} & 0.699 & 0.699 & \textcolor{green}{0.696} & 0.703 & 0.703 & \textcolor{green}{0.701} & 0.707 & 0.705 & \textcolor{green}{0.703} & 0.707 & 0.707 & \textcolor{green}{0.703} \\ 
\hline
\multirow{3}{*}{\makecell{Technology\\Sector}} 
 & 60 & 0.792 & 0.790 & \textcolor{green}{0.782} & 0.777 & 0.778 & \textcolor{green}{0.776} & 0.785 & 0.786 & \textcolor{green}{0.783} & 0.790 & 0.786 & \textcolor{green}{0.781} & 0.786 & 0.787 & \textcolor{green}{0.781} \\ 
 & 90 & 0.787 & 0.786 & \textcolor{green}{0.780} & 0.777 & 0.777 & \textcolor{green}{0.775} & 0.781 & 0.781 & \textcolor{green}{0.780} & 0.786 & 0.783 & \textcolor{green}{0.782} & 0.784 & 0.783 & \textcolor{green}{0.779} \\ 
 & 120 & 0.787 & 0.787 & \textcolor{green}{0.781} & 0.777 & 0.777 & \textcolor{green}{0.776} & 0.780 & 0.781 & \textcolor{green}{0.779} & 0.785 & 0.783 & \textcolor{green}{0.781} & 0.784 & 0.783 & \textcolor{green}{0.780} \\ 
\hline
\multirow{3}{*}{\makecell{Consumer\\Sector}} 
 & 60 & 0.759 & 0.757 & \textcolor{green}{0.751} & 0.748 & 0.748 & \textcolor{green}{0.746} & 0.754 & 0.754 & \textcolor{green}{0.751} & 0.757 & 0.755 & \textcolor{green}{0.752} & 0.755 & 0.755 & \textcolor{green}{0.750} \\ 
 & 90 & 0.758 & 0.757 & \textcolor{green}{0.753} & 0.751 & 0.750 & \textcolor{green}{0.748} & 0.753 & 0.754 & \textcolor{green}{0.751} & 0.758 & 0.756 & \textcolor{green}{0.754} & 0.756 & 0.756 & \textcolor{green}{0.753} \\ 
 & 120 & 0.761 & 0.761 & \textcolor{green}{0.757} & 0.753 & 0.753 & \textcolor{green}{0.751} & 0.756 & 0.756 & \textcolor{green}{0.754} & 0.760 & 0.759 & \textcolor{green}{0.757} & 0.759 & 0.759 & \textcolor{green}{0.757} \\ 
\hline
\multirow{3}{*}{\makecell{Energy\\Sector}} 
 & 60 & 0.541 & 0.540 & \textcolor{green}{0.533} & 0.528 & 0.528 & \textcolor{green}{0.524} & 0.535 & 0.536 & \textcolor{green}{0.533} & 0.538 & 0.535 & \textcolor{green}{0.534} & 0.534 & 0.535 & \textcolor{green}{0.532} \\ 
 & 90 & 0.537 & 0.536 & \textcolor{green}{0.530} & 0.526 & 0.526 & \textcolor{green}{0.522} & 0.532 & 0.532 & \textcolor{green}{0.531} & 0.537 & 0.535 & \textcolor{green}{0.532} & 0.532 & 0.532 & \textcolor{green}{0.531} \\ 
 & 120 & 0.537 & 0.536 & \textcolor{green}{0.533} & 0.527 & 0.527 & \textcolor{green}{0.523} & 0.531 & 0.531 & \textcolor{green}{0.530} & 0.535 & 0.534 & \textcolor{green}{0.531} & 0.533 & 0.533 & \textcolor{green}{0.531} \\ 
\hline
\multirow{3}{*}{\makecell{Nasdaq Mixed\\Excerpt}} 
 & 60 & 0.682 & 0.680 & \textcolor{green}{0.673} & 0.671 & 0.672 & \textcolor{green}{0.669} & 0.676 & 0.677 & \textcolor{green}{0.674} & 0.679 & 0.677 & \textcolor{green}{0.674} & 0.678 & 0.679 & \textcolor{green}{0.674} \\ 
 & 90 & 0.678 & 0.677 & \textcolor{green}{0.672} & 0.671 & 0.671 & \textcolor{green}{0.669} & 0.674 & 0.674 & \textcolor{green}{0.672} & 0.678 & 0.675 & \textcolor{green}{0.674} & 0.677 & 0.678 & \textcolor{green}{0.674} \\ 
 & 120 & 0.678 & 0.677 & \textcolor{green}{0.674} & 0.671 & 0.671 & \textcolor{green}{0.669} & 0.673 & 0.673 & \textcolor{green}{0.672} & 0.677 & 0.675 & \textcolor{green}{0.674} & 0.677 & 0.676 & \textcolor{green}{0.674} \\ 
\hline
\multirow{3}{*}{\makecell{Macro and\\Market Series}}
 & 60 & 0.616 & 0.616 & \textcolor{green}{0.612} & 0.605 & 0.606 & \textcolor{green}{0.603} & 0.615 & 0.615 & \textcolor{green}{0.613} & 0.614 & 0.613 & \textcolor{green}{0.610} & \textcolor{green}{0.612} & 0.615 & 0.612 \\ 
 & 90 & 0.614 & 0.614 & \textcolor{green}{0.610} & 0.607 & 0.606 & \textcolor{green}{0.604} & 0.613 & 0.613 & \textcolor{green}{0.612} & 0.616 & 0.615 & \textcolor{green}{0.613} & 0.612 & \textcolor{green}{0.612} & 0.612 \\ 
 & 120 & 0.617 & 0.617 & \textcolor{green}{0.614} & 0.610 & 0.610 & \textcolor{green}{0.608} & 0.615 & 0.614 & \textcolor{green}{0.612} & 0.619 & 0.617 & \textcolor{green}{0.617} & 0.615 & \textcolor{green}{0.614} & 0.615 \\ 
\hline
\end{tabular}
}
\end{table*}

Based on the unrounded results underlying the four tables, DPAL achieves lower standardized return errors than the corresponding original adaptation objective in 470 of the 480 comparisons. Specifically, it reduces MSE and MAE in 119 and 118 of the 120 TAFAS \citep{TAFAS} settings, respectively, and in 118 and 115 of the 120 PETSA \citep{PETSA} settings. The ten exceptions are minor, with the largest relative increase being approximately $0.24\%$. Compared with the Base model, DPAL achieves lower errors in almost all settings.

Although the reductions in standardized return MSE and MAE are generally less pronounced than those in reconstructed price MSE, they are observed across nearly all evaluated settings. This difference in magnitude is consistent with the cumulative nature of price reconstruction, as small improvements in pointwise return forecasts can become more visible when returns are compounded over the forecast horizon. Thus, the benefits of DPAL extend beyond reconstructed price accuracy to the pointwise accuracy of standardized return forecasts.

\subsubsection{Scale-consistent Evaluation}
\label{appendix_scale_consistent_evaluation}
As described in Section 4.1.4, DPAL computes its directional and price calibration losses using training-set inverse standardization, whereas the main reconstructed-price evaluation uses history-conditioned rescaling. To assess whether the gains persist when these transformations are aligned, we additionally evaluate both Original TTA and TTA + DPAL by replacing Eq. (8) with
$$\hat{r}^{sc}_{w,h,c} = \hat{r}^{norm}_{w,h,c}\sigma^{sc}_c+\mu^{sc}_c.$$
Here, $\mu^{sc}_c$ and $\sigma^{sc}_c$ are the training-set statistics defined in Section 4.1.4 and remain fixed across evaluation windows. Price reconstruction and MSE computation follow Eqs. (9)–(10), using $\hat{r}^{sc}_{w,h,c}$ as the predicted returns and retaining the observed returns on their original scale. We denote the resulting reconstructed-price error by $\mathrm{MSE}^{sc}_{price}$. For each framework, we compare the two methods across the same $S=120$ combinations of datasets, forecasting backbones, and forecast horizons. For setting $s$, we define the MSE ratio and relative improvement as
$$\rho_s = \frac{\mathrm{MSE}^{sc}_{DPAL,s}}{\mathrm{MSE}^{sc}_{Original,s}}, \qquad I_s=(1-\rho_s)\times 100\%.$$
We report the number of settings in Table~\ref{tab:scale_consistent_eval} with strictly lower MSE under DPAL ($\rho_s<1$), the median of $I_s$, and the geometric-mean improvement, defined as
$$\qquad I_{geo} =\left[ 1-\exp\left( \frac{1}{S}\sum_{s=1}^{S}\log\rho_s \right) \right]\times 100\%.$$

\begin{table}[htbp]
    \centering
    \renewcommand{\thetable}{A.2.7} 
    \caption{Scale-consistent reconstructed-price evaluation using training-set inverse standardization. Improvements are relative to Original TTA; positive values indicate lower MSE with DPAL.}
    \label{tab:scale_consistent_eval}
    \begin{tabular}{lccc}
        \toprule
        Framework & Settings improved by DPAL & Median improvement & Geometric-mean improvement \\
        \midrule
        TAFAS & 102/120 & 17.24\% & 18.24\% \\
        PETSA & 104/120 & 20.85\% & 19.66\% \\
        \bottomrule
    \end{tabular}
\end{table}

DPAL reduces reconstructed-price MSE in most settings under both frameworks, with positive median and geometric-mean improvements. These results indicate that the observed gains are not solely attributable to history-conditioned rescaling.

\subsection{Statistical Analysis}
\label{app:statistical_analysis}
\subsubsection{Exact Wilcoxon signed-rank test}
To assess whether the improvement obtained by DPAL is consistent across datasets, rather than being driven by a large number of backbone--horizon configurations from a small number of datasets, we conducted a dataset-level statistical analysis based on Tables \ref{ablation_study_results}. For each dataset ($d$), backbone ($m$), and forecasting horizon ($h$), we first calculated the natural logarithm of the ratio between the reconstructed price MSE obtained with DPAL and that obtained with the original adaptation objective:
\begin{equation}
g_{d,m,h} = \log\left(
\frac{\mathrm{MSE}_{\mathrm{DPAL},d,m,h}}
{\mathrm{MSE}_{\mathrm{Original},d,m,h}}
\right).
\end{equation}
Here, $\mathrm{MSE}_{\mathrm{Original}}$ denotes the result in the first adaptation column to the right of the frozen Base model, whereas $\mathrm{MSE}_{\mathrm{DPAL}}$ denotes the result in the last column, corresponding to the complete DPAL objective. A negative value of $g_{d,m,h}$ indicates that DPAL achieves a lower reconstructed price MSE than the original objective for the corresponding configuration.

Each table contains five forecasting backbones and three forecasting horizons, $h \in \{60,90,120\}$, resulting in 15 configurations per dataset. We therefore summarized the configuration-level log ratios within each dataset as
\begin{equation}
K_d = \frac{1}{15} \sum_{m=1}^{5} \sum_{h \in \{60,90,120\}} g_{d,m,h}.
\end{equation}
Equivalently, $\exp(K_d)$ represents the geometric mean of the DPAL-to-Original MSE ratios across the 15 configurations. Aggregation was performed before hypothesis testing because configurations evaluated on the same dataset were not treated as independent experimental units.

The resulting dataset-level scores are reported in Table~\ref{tab:dataset_level_log_ratio}. All eight scores are negative under both TAFAS \citep{TAFAS} and PETSA \citep{PETSA}.

\begin{table}
\renewcommand{\thetable}{A.3.1} 
\centering
\caption{Dataset-level mean log MSE ratios ($K_d$). Negative values indicate lower reconstructed price MSE for DPAL than for the original adaptation objective.}
\label{tab:dataset_level_log_ratio}
\begin{tabular}{c|cc}
\toprule
Dataset & TAFAS & PETSA \\
\midrule
Magnificent 7              & -0.659118 & -0.521236 \\
S\&P 500 Excerpt           & -0.571172 & -0.322853 \\
Russell 2000 Excerpt       & -0.585539 & -0.524676 \\
Technology Sector          & -0.617923 & -0.629437 \\
Consumer Sector            & -0.367755 & -0.402655 \\
Energy Sector              & -0.270304 & -0.190842 \\
Nasdaq Mixed Excerpt       & -0.327553 & -0.289190 \\
Macro and Market Series    & -0.252861 & -0.226572 \\
\bottomrule
\end{tabular}
\end{table}

We then applied an exact one-sided Wilcoxon signed-rank test separately to the eight dataset-level scores obtained under each adaptation framework. Let $\theta$ denote the location parameter of the dataset-level log ratios. The hypotheses were
\begin{equation}
H_0: \theta \geq 0, \qquad H_1: \theta < 0,
\end{equation}
with the exact null distribution evaluated at $\theta = 0$. The significance level was set to $\alpha = 0.05$.

For TAFAS \citep{TAFAS}, all eight $K_d$ values were negative, yielding a positive-rank sum of $W^{+} = 0$ and a negative-rank sum of $W^{-} = 36$. The exact one-sided test gave $p = 0.00390625$. The same result was obtained for PETSA \citep{PETSA}, for which all eight scores were also negative ($W^{+} = 0$, $W^{-} = 36$, and $p = 0.00390625$). Because there were no zero differences or tied absolute ranks, the exact probability was obtained from the $2^8 = 256$ possible sign assignments; only the assignment in which all eight scores are negative produces $W^{+} = 0$.

Therefore, the null hypothesis was rejected for both TAFAS \citep{TAFAS} and PETSA \citep{PETSA}. These results provide evidence that the location of the dataset-level mean log MSE ratios is below zero. In other words, across the evaluated datasets, DPAL produces a geometric mean reconstructed price MSE ratio below one relative to the original adaptation objective. Both conclusions also remain significant under a Bonferroni correction for the two framework-specific tests.

\subsubsection{Counts of Best and Second-Best Results}
\label{app:best_second_counts}
To further examine the consistency of each loss configuration across experimental settings, we counted the number of times each configuration achieved the best (lowest MSE) or second-best (second-lowest MSE) reconstructed price MSE in Tables \ref{ablation_study_results}. The counts were first obtained separately for each table and then aggregated across the eight datasets within each adaptation framework. Table~\ref{tab:best_second_counts} summarizes the results.

\begin{table}[h]
\renewcommand{\thetable}{A.3.2} 
\centering
\caption{Counts of best and second-best reconstructed price MSE results across the experimental settings under TAFAS and PETSA. Original denotes results using PETSA's original adaptation objective; the TAFAS baseline is represented by the MSE-only configuration.}
\label{tab:best_second_counts}
\begin{tabular}{lll|cccc}
\toprule
\multicolumn{3}{c|}{} & \multicolumn{2}{c}{TAFAS} & \multicolumn{2}{c}{PETSA} \\
\cmidrule(lr){4-5}\cmidrule(lr){6-7}
\multicolumn{3}{c|}{Configuration} & Best & Second-best & Best & Second-best \\
\midrule
\multicolumn{3}{c|}{Base} & 1 & 3 & 2 & 3 \\
\multicolumn{3}{c|}{Original} & -- & -- & 0 & 1 \\
mse & dir & price & & & & \\
\checkmark & & & 0 & 0 & 0 & 0 \\
& \checkmark & & 5 & 0 & 12 & 30 \\
& & \checkmark & 5 & 0 & 4 & 1 \\
\checkmark & \checkmark & & 14 & 10 & 0 & 19 \\
\checkmark & & \checkmark & 1 & 3 & 0 & 0 \\
& \checkmark & \checkmark & 56 & 46 & 23 & 62 \\
\checkmark & \checkmark & \checkmark & 38 & 58 & 79 & 4 \\
\bottomrule
\end{tabular}
\end{table}

Under TAFAS \citep{TAFAS}, the combination of direction-aware and price calibration losses without the pointwise MSE term ranked among the top two, achieving 56 best and 46 second-best results. The complete DPAL objective followed closely, with 38 best and 58 second-best results. Under PETSA \citep{PETSA}, complete DPAL achieved the largest number of best results (79), while the combination of direction-aware and price calibration losses appeared among the top two in 85 settings. By comparison, the MSE-only objective produced no best or second-best results under either framework, while the original PETSA \citep{PETSA} objective produced only one second-best result.

Overall, the best and second-best results are concentrated among configurations that include both direction-aware and price calibration losses. The precise pattern differs between the two adaptation frameworks: the "direction plus price" configuration is most frequently the best under TAFAS \citep{TAFAS}, whereas the complete DPAL objective is most frequently the best under PETSA \citep{PETSA}. These results indicate that the directional and price-based constraints make complementary contributions across the evaluated settings, while the most effective balance between these constraints and the pointwise MSE term may depend on the underlying adaptation framework.

\subsection{Ablation on the Price Calibration Length}
\subsubsection{Experimental Configuration}
To examine the effect of the truncation length $k$ in the price calibration loss, we conduct ablation experiments under both TAFAS \citep{TAFAS} and PETSA \citep{PETSA}. The experiments use the same eight financial datasets as the main evaluation: Magnificent 7, S\&P 500 Excerpt, Russell 2000 Excerpt, Technology Sector, Consumer Sector, Energy Sector, Nasdaq Mixed Excerpt, and Macro and Market Series. We consider five forecasting backbones, including iTransformer \citep{iTransformer}, PatchTST \citep{PatchTST}, DLinear \citep{DLinear}, FreTS \citep{FreTS}, and OLS \citep{OLS}, together with three forecast horizons $H \in \{60, 90, 120\}$. Each adaptation framework consequently contains $8 \times 5 \times 3 = 120$ combinations of dataset, forecasting backbone, and horizon. All experiments retain the data splits, pretrained models, adaptation procedure, and matured-ground-truth protocol used in the main experiments. We also fix $\alpha=10^5$, $\beta=10^2$, and all other optimization settings so that $k$ is the only factor varied in this ablation.

We compare three settings of $k$:
\begin{itemize}\item Reference setting: Follows the main experiments and uses $k = 15$ for $H = 60$, and $k = 30$ for both $H = 90$ and $H = 120$. \item Intermediate setting: Extends the calibration interval to $k = 30$, $45$, and $60$, respectively. \item Full horizon setting: Uses $k = H$, corresponding to $k = 60$, $90$, and $120$. \end{itemize}
Across these settings, only the number of forecast steps included in the price calibration loss is changed. Performance is still evaluated using reconstructed price MSE over the complete $H$-step forecast horizon.

\subsubsection{Aggregation of Results}
We first count how frequently each setting produces the lowest reconstructed price MSE. For every combination of dataset, forecasting backbone, and horizon, the three settings are compared using their unrounded MSE values, and one count is assigned to the setting with the lowest value. Because each adaptation framework contains 120 combinations, the lowest MSE counts of the three settings sum to 120.

Reconstructed price MSE can differ substantially in scale across datasets. Directly averaging the raw MSE values would consequently give greater influence to datasets with numerically larger errors. We instead use the reference setting as the baseline and aggregate the relative MSE ratios using a geometric mean.

Let $f \in \{\mathrm{TAFAS}, \mathrm{PETSA}\}$ denote the adaptation framework, let $s$ denote a setting of $k$, and let $E_{f,i}^{(s)}$ be the reconstructed price MSE of setting $s$ for experimental combination $i$. The corresponding MSE under the reference setting is denoted by $E_{f,i}^{(\mathrm{ref})}$. We first calculate the ratio:$$r_{f,i}^{(s)} = \frac{E_{f,i}^{(s)}}{E_{f,i}^{(\mathrm{ref})}}.$$

The geometric mean relative change across all $N = 120$ combinations is then defined as:$$\Delta_{f}^{(s)} = \left[ \left( \prod_{i=1}^{N} r_{f,i}^{(s)} \right)^{1/N} - 1 \right] \times 100\% = \left[ \exp\left( \frac{1}{N} \sum_{i=1}^{N} \log \frac{E_{f,i}^{(s)}}{E_{f,i}^{(\mathrm{ref})}} \right) - 1 \right] \times 100\%.$$

This calculation first normalizes each alternative setting against the reference setting within the same experimental combination and then aggregates the resulting multiplicative ratios. It therefore avoids directly averaging raw MSE values from datasets with different numerical scales. The geometric mean is also a natural summary for relative ratios: $\Delta_f^{(s)} < 0$ indicates a lower aggregate MSE than the reference setting, whereas $\Delta_f^{(s)} > 0$ indicates a higher aggregate MSE. The relative change of the reference setting is zero by definition.

\subsubsection{Results and Conclusions}

As shown in Table~\ref{test_of_k}, the reference setting achieves the largest number of lowest MSE results under both adaptation frameworks, with 51 cases under TAFAS \citep{TAFAS} and 56 under PETSA \citep{PETSA}. The intermediate setting achieves 46 and 40 lowest MSE results, respectively, while increasing the geometric mean MSE relative to the reference by only 0.48\% under TAFAS \citep{TAFAS} and 0.86\% under PETSA \citep{PETSA}. These small changes indicate that moderately extending $k$ has a limited overall effect and that the reference and intermediate settings form a relatively stable range of truncation lengths.

\begin{table}[h]
\renewcommand{\thetable}{A.4}
\centering
\caption{Ablation results for different price calibration lengths under TAFAS and PETSA. The lowest MSE count is the number of times that a setting achieves the lowest reconstructed price MSE among the 120 combinations of datasets, forecasting backbones, and horizons. Geometric mean changes are calculated relative to the reference setting.}
\label{tab}
\begin{tabular}{lccc|cc|cc}
\toprule
& \multicolumn{3}{c|}{$k$} &
\multicolumn{2}{c|}{Lowest MSE Count} &
\multicolumn{2}{c}{Geometric Mean Change} \\
\cmidrule(lr){2-4}
\cmidrule(lr){5-6}
\cmidrule(lr){7-8}
Setting & $H=60$ & $H=90$ & $H=120$
& TAFAS & PETSA & TAFAS & PETSA \\
\midrule
Reference & 15 & 30 & 30 & \textbf{51} & \textbf{56} & ----- & ----- \\
Intermediate & 30 & 45 & 60 & 46 & 40 & +0.48\% & +0.86\% \\
Full horizon & 60 & 90 & 120 & 23 & 24 & +4.95\% & +7.82\% \\
\bottomrule
\end{tabular}
\label{test_of_k}
\end{table}

When $k$ is extended to the complete forecast horizon, the number of lowest MSE results decreases to 23 under TAFAS \citep{TAFAS} and 24 under PETSA \citep{PETSA}. At the same time, the geometric mean MSE increases by 4.95\% and 7.82\%, respectively. Compared with the intermediate setting, the full horizon setting produces a substantially larger increase under both frameworks. This common pattern indicates that applying the price calibration constraint to every forecast step is generally less reliable than restricting it to an early portion of the predicted trajectory. The observation is consistent with the cumulative nature of price reconstruction: prices at more distant steps incorporate longer sequences of compounded returns, so including the complete trajectory in the adaptation loss can make the calibration objective more sensitive to accumulated deviations.

Overall, the ablation results support using a truncated interval in the price calibration loss. The reference schedule $k=15/30/30$ produces the largest number of lowest MSE results under both adaptation frameworks and is more stable overall than calibration over the full horizon. Meanwhile, the small geometric mean differences between the reference and intermediate settings suggest that DPAL does not depend on a single sharply isolated value of $k$ but remains relatively stable across a short to intermediate calibration range. The reference setting can thus be regarded as a reasonable common choice across the evaluated datasets, forecasting backbones, and adaptation frameworks.

\clearpage
\section{Additional Dataset Details}
\label{dataset_details}
\subsection{Data Source and Dataset Construction}
All raw series were obtained from Yahoo Finance through the \texttt{yfinance} package. For each dataset, all constituent tickers were downloaded jointly using \texttt{yf.download()} with \texttt{auto\_adjust=True}. Under this setting, the values returned in the \texttt{Close} field are adjusted for corporate actions and therefore differ from unadjusted closing prices. No separate \texttt{Adj Close} field was used.

The requested download interval was specified as $[2012\text{-}05\text{-}18, 2026\text{-}01\text{-}01)$, with the starting date included and the ending date excluded. The latest retained observation is therefore December 31, 2025. After return construction and missing value removal, seven datasets begin on May 21, 2012, whereas the Russell 2000 Excerpt begins on August 20, 2013 because complete observations for all of its constituent series were not available before that date.

The equity panels were manually curated. The panels were designed to cover different market segments and company types while retaining sufficiently long historical series and avoiding repeated tickers across the seven equity datasets. Table~\ref{The eight datasets} reports their complete composition.

\begin{table}[htbp]
\renewcommand{\thetable}{B.1} 
\centering
\caption{The eight datasets.}
\label{The eight datasets}
\begin{tabular}{lc p{9.5cm}}
\toprule
\textbf{Dataset} & \textbf{Number of series} & \textbf{Tickers} \\
\midrule
Magnificent 7 & 7 & AAPL, MSFT, GOOG, AMZN, NVDA, META, TSLA \\
\midrule
S\&P 500 Excerpt & 20 & CDNS, SNPS, KLAC, LRCX, MCHP, MPWR, MRVL, ON, TER, SWKS, STX, WDC, FSLR, VRSN, MSI, ROP, FICO, PTC, ZBRA, APH \\
\midrule
Russell 2000 Excerpt & 20 & MHO, WWD, FIX, AAON, ENSG, CASY, KBR, TEX, UGI, WBS, AWI, SPSC, CNO, MTG, PII, SAIA, LECO, THG, MUSA, FIBK \\
\midrule
Technology Sector & 20 & AMD, AVGO, QCOM, INTC, ASML, MU, TXN, AMAT, CRM, ADBE, TSM, ORCL, ADSK, FTNT, AKAM, NTAP, INTU, NFLX, CSCO, IBM \\
\midrule
Consumer Sector & 20 & CHD, KO, PEP, WMT, COST, TGT, CL, KMB, GIS, HSY, MKC, EL, KR, HD, LOW, TJX, NKE, MCD, SBUX, YUM \\
\midrule
Energy Sector & 20 & XOM, CVX, COP, OXY, EOG, DVN, APA, EQT, RRC, MUR, SM, NOG, SLB, HAL, NOV, VLO, MPC, PSX, KMI, WMB \\
\midrule
Nasdaq Mixed Excerpt & 20 & AMGN, GILD, ISRG, IDXX, CMCSA, TTWO, CSX, PCAR, CTAS, FAST, CPRT, ODFL, ADP, VRSK, CSGP, NXPI, CTSH, BKNG, MELI, AEP \\
\midrule
Macro and Market Series & 10 & \textasciicircum GSPC, \textasciicircum NDX, \textasciicircum MID, \textasciicircum RUT, \textasciicircum DJI, \textasciicircum TNX, DX-Y.NYB, GC=F, CL=F, \textasciicircum NYA \\
\bottomrule
\end{tabular}
\end{table}

\subsection{Return Construction and Trading Date Alignment}
Let $P_{t,c}$ denote the adjusted closing value of series $c$ on trading day $t$. The daily simple return was calculated using \texttt{pct\_change()}:
$$r_{t,c} = \frac{P_{t,c}-P_{t-1,c}}{P_{t-1,c}} = \frac{P_{t,c}}{P_{t-1,c}}-1.$$
Returns were stored as decimal values rather than percentages.

Within each dataset, the series were first aligned on the combined date index produced by the joint download. Under the default behavior of \texttt{pct\_change()} used when generating the datasets, a missing price observation was replaced by the most recently available price before the percentage change was calculated. Consequently, if a series had no new quotation on a date retained in the combined index, its return on that date was effectively recorded as zero. Rows that still contained undefined returns after this operation, including the initial row required by the percentage change calculation and rows preceding the availability of every constituent series, were removed using \texttt{dropna()}.

This procedure ensures that all series within a dataset share the same final date index, but it does not require every retained date to be a trading day for every constituent. This distinction is most relevant to the Macro and Market Series dataset because its equity indices, Treasury yield index, currency index, and commodity futures series follow different trading calendars. A date could therefore be retained when only some markets reported new values; series without a new observation on that date receive a zero return under the procedure described above. Thus, the Macro and Market Series dataset is not restricted to the strict intersection of the trading dates shared by all ten series.

Each constituent was retained as a separate target series. No aggregation across series or constituent weighting was applied.

\subsection{Chronological Data Splits}

Each dataset was arranged chronologically and divided into training, validation, and test portions using an approximate ratio of $7{:}1{:}2$. For a dataset containing $N$ return observations, the numbers assigned to the training and test portions were calculated as
\begin{equation}
N_{\mathrm{train}}=\left\lfloor 0.7N \right\rfloor,
\qquad
N_{\mathrm{test}}=\left\lfloor 0.2N \right\rfloor,
\end{equation}
and the remaining observations were assigned to validation:
\begin{equation}
N_{\mathrm{val}} = N - N_{\mathrm{train}} - N_{\mathrm{test}}.
\end{equation}
The first $N_{\mathrm{train}}$ observations formed the training portion, the following $N_{\mathrm{val}}$ observations formed the validation portion, and the final $N_{\mathrm{test}}$ observations formed the test portion. Table~\ref{tab:dataset_splits} reports the resulting date ranges, with both endpoints included.

\begin{table}[htbp]
\renewcommand{\thetable}{B.2}
\centering
\small
\setlength{\tabcolsep}{3.5pt}
\caption{Chronological data splits. The number in parentheses is the number of return observations in the corresponding portion.}
\label{tab:dataset_splits}
\begin{tabular}{lccc}
\toprule
\textbf{Dataset} & \textbf{Training} & \textbf{Validation} & \textbf{Test} \\
\midrule
Magnificent 7
& 2012-05-21--2021-11-24 (2,396)
& 2021-11-26--2023-04-11 (344)
& 2023-04-12--2025-12-31 (684) \\
S\&P 500 Excerpt
& 2012-05-21--2021-11-24 (2,396)
& 2021-11-26--2023-04-11 (344)
& 2023-04-12--2025-12-31 (684) \\
Russell 2000 Excerpt
& 2013-08-20--2022-04-11 (2,177)
& 2022-04-12--2023-07-11 (312)
& 2023-07-12--2025-12-31 (622) \\
Technology Sector
& 2012-05-21--2021-11-24 (2,396)
& 2021-11-26--2023-04-11 (344)
& 2023-04-12--2025-12-31 (684) \\
Consumer Sector
& 2012-05-21--2021-11-24 (2,396)
& 2021-11-26--2023-04-11 (344)
& 2023-04-12--2025-12-31 (684) \\
Energy Sector
& 2012-05-21--2021-11-24 (2,396)
& 2021-11-26--2023-04-11 (344)
& 2023-04-12--2025-12-31 (684) \\
Nasdaq Mixed Excerpt
& 2012-05-21--2021-11-24 (2,396)
& 2021-11-26--2023-04-11 (344)
& 2023-04-12--2025-12-31 (684) \\
Macro and Market Series
& 2012-05-21--2021-11-29 (2,399)
& 2021-11-30--2023-04-13 (344)
& 2023-04-14--2025-12-31 (685) \\
\bottomrule
\end{tabular}
\end{table}

When constructing validation and test samples, the data loader additionally included the 60 observations immediately preceding each split boundary as historical input. These observations provided the lookback context required by the forecasting models, while the corresponding prediction targets began at the validation or test boundary. The additional context therefore did not alter the target date ranges reported in Table~\ref{tab:dataset_splits}. The standardization statistics were fitted using only the training portion and were subsequently applied to the validation and test portions.

\end{document}

%% file: math_commands.tex
\usepackage{amsmath,amsfonts,bm}

\def\eqref#1{equation~\ref{#1}}
\def\Eqref#1{Equation~\ref{#1}}
\def\1{\bm{1}}

\DeclareMathAlphabet{\mathsfit}{\encodingdefault}{\sfdefault}{m}{sl}
\SetMathAlphabet{\mathsfit}{bold}{\encodingdefault}{\sfdefault}{bx}{n}